\documentclass[10pt,twocolumn,letterpaper]{article}

\usepackage{cvpr}              

\usepackage{graphicx}
\usepackage{amsmath}

\DeclareMathOperator*{\argmin}{arg\,min}

\usepackage{amssymb}
\usepackage{booktabs}
\usepackage{microtype}
\usepackage{enumitem}
\usepackage{xcolor}
\usepackage{bm}
\usepackage{algorithm2e}
\RestyleAlgo{ruled}
\SetKwComment{Comment}{/* }{ */}

\makeatletter
\robustify\@latex@warning@no@line
\makeatother
\usepackage{authblk}
\makeatletter 
\renewcommand\AB@affilsepx{~ \protect\Affilfont}

\robustify\@latex@warning@no@line
\makeatother

\usepackage{ulem}

\usepackage{comment}

\usepackage{amsmath}

\usepackage{diagbox}
\usepackage{multicol}
\usepackage{multirow}
\usepackage{float}

\usepackage{pifont}

\usepackage{lipsum}

\definecolor{cvprblue}{rgb}{0.21,0.49,0.74}
\usepackage[pagebackref,breaklinks,colorlinks,citecolor=cvprblue]{hyperref}

\usepackage[capitalize]{cleveref}
\creflabelformat{equation}{#2#1#3}
\crefname{section}{Sec.}{Secs.}
\Crefname{section}{Section}{Sections}
\Crefname{table}{Table}{Tables}
\crefname{table}{Tab.}{Tabs.}
\Crefname{algorithm}{Algorithm}{Algorithms}
\crefname{algorithm}{Algo.}{Algos.}

\def\paperID{10326} 
\def\confName{CVPR}
\def\confYear{2025}

\begin{document}
\title{Color Alignment in Diffusion}

\author[1]{Ka Chun Shum}
\author[2]{Binh-Son Hua}
\author[3]{Duc Thanh Nguyen}
\author[1]{Sai-Kit Yeung}

\affil[1]{Hong Kong University of Science and Technology}
\affil[2]{Trinity College Dublin}
\affil[3]{Deakin University}

\maketitle

\begin{abstract}

Diffusion models have shown great promise in synthesizing visually appealing images. However, it remains challenging to condition the synthesis at a fine-grained level, for instance, synthesizing image pixels following some generic color pattern. Existing image synthesis methods often produce contents that fall outside the desired pixel conditions. To address this, we introduce a novel color alignment algorithm that confines the generative process in diffusion models within a given color pattern. Specifically, we project diffusion terms, either imagery samples or latent representations, into a conditional color space to align with the input color distribution. This strategy simplifies the prediction in diffusion models within a color manifold while still allowing plausible structures in generated contents, thus enabling the generation of diverse contents that comply with the target color pattern. Experimental results demonstrate our state-of-the-art performance in conditioning and controlling of color pixels, while maintaining on-par generation quality and diversity in comparison with regular diffusion models.

\end{abstract}

\section{Introduction}
Recently, diffusion models~\cite{ho2020denoising} and their large-scale text-conditioned variants~\cite{rombach2022high,saharia2022photorealistic} have demonstrated impressive quality and diversity of generated contents in image synthesis. Subsequent works have explored leveraging this capability for possible controllability, incorporating various constraints, such as structural~\cite{zhang2023adding,mou2024t2i,ramesh2022hierarchical} and reference constraints~\cite{ye2023ip,hertz2024style}, which typically serve as additional inputs. However, due to the inherent randomness in the diffusion process, it remains challenging to control generated contents at a fine-grained level, for instance, synthesizing image pixels following some given content and generic color distribution.

In this paper, we focus on such a fine-grained control over pixel colors in synthesized images to meet with three color-conditioned generation criteria: the \textit{accuracy} in adhering to given color values; the \textit{completeness} of covering proportions of the specified colors; and the \textit{disentanglement} of the colors into various structures. Such a color-conditioned generation ability would be a great aid to numerous downstream tasks in creative artwork and graphic design~\cite{Epstein_Science_2023}. 


\begin{figure}[t]
  \begin{subfigure}{0.237\columnwidth}
    \includegraphics[width=\textwidth]{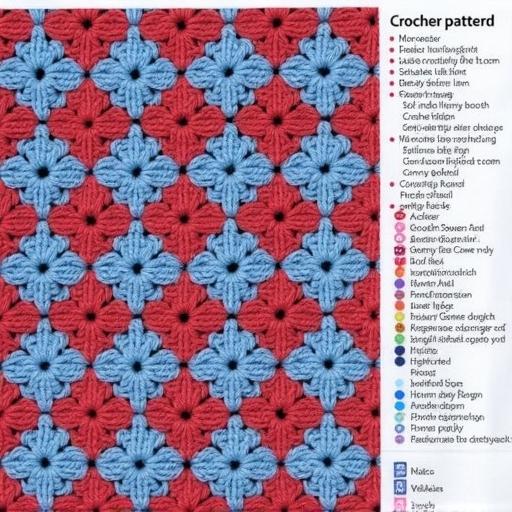}
  \end{subfigure}
  \hfill 
  \begin{subfigure}{0.237\columnwidth}
    \includegraphics[width=\textwidth]{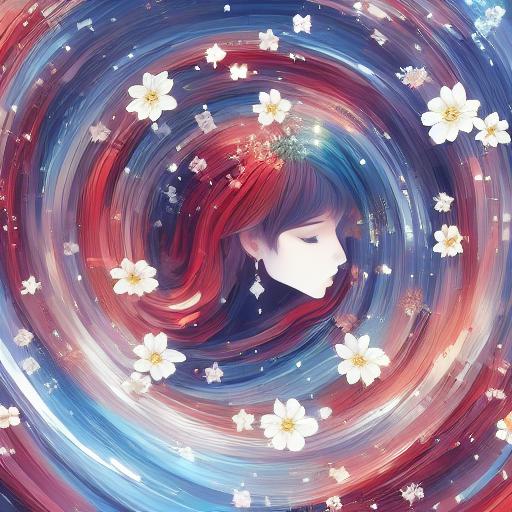}
  \end{subfigure}
  \begin{subfigure}{0.237\columnwidth}
    \includegraphics[width=\textwidth]{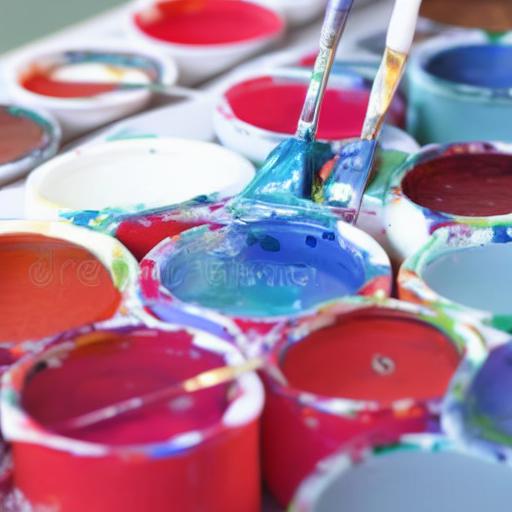}
  \end{subfigure}
  \hfill 
  \begin{subfigure}{0.237\columnwidth}
    \includegraphics[width=\textwidth]{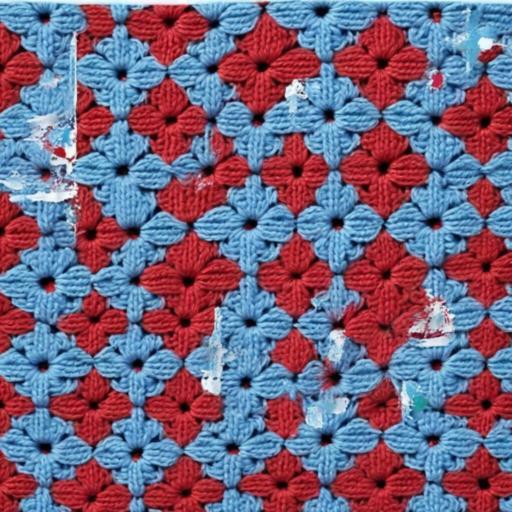}
  \end{subfigure}
  \\[0.6mm]
  \begin{subfigure}{0.237\columnwidth}
    \includegraphics[width=\textwidth]{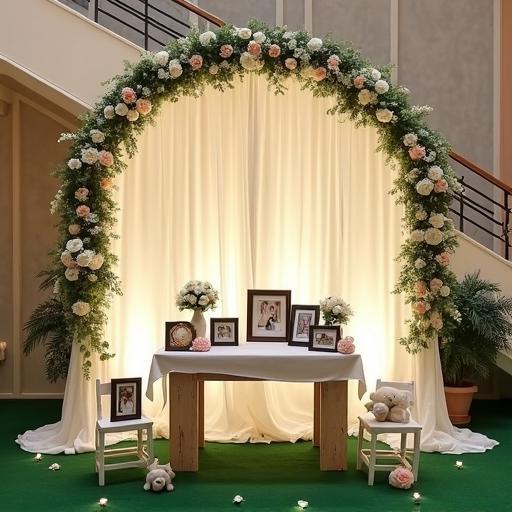}
  \end{subfigure}
  \hfill 
  \begin{subfigure}{0.237\columnwidth}
    \includegraphics[width=\textwidth]{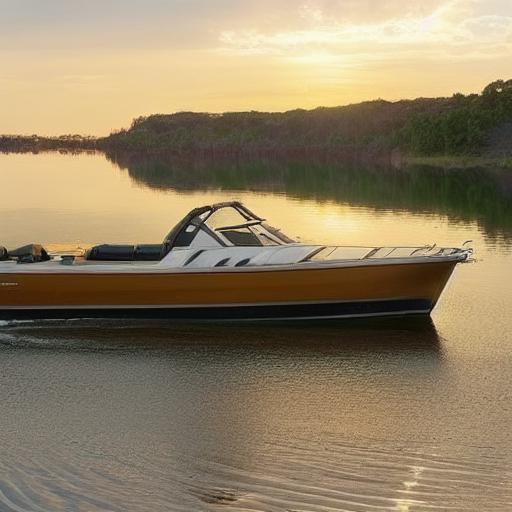}
  \end{subfigure}
  \begin{subfigure}{0.237\columnwidth}
    \includegraphics[width=\textwidth]{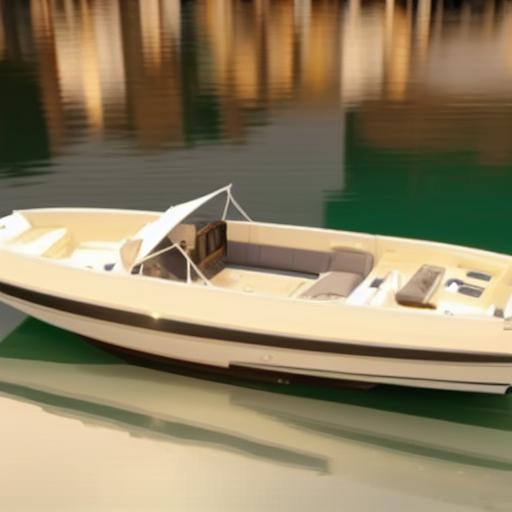}
  \end{subfigure}
  \hfill 
  \begin{subfigure}{0.237\columnwidth}
    \includegraphics[width=\textwidth]{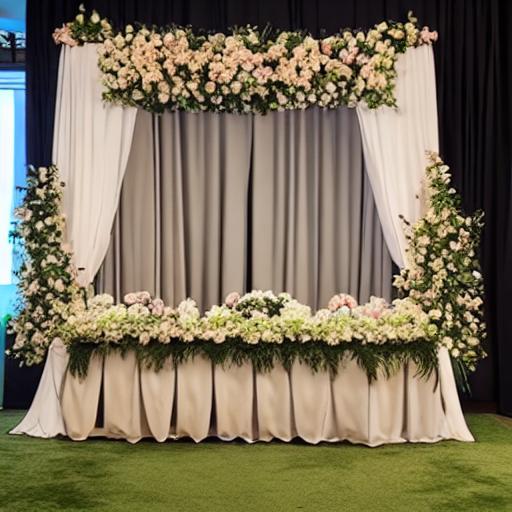}
  \end{subfigure}
  \\[2mm]
  \begin{subfigure}{0.237\columnwidth}
    \includegraphics[width=\textwidth]{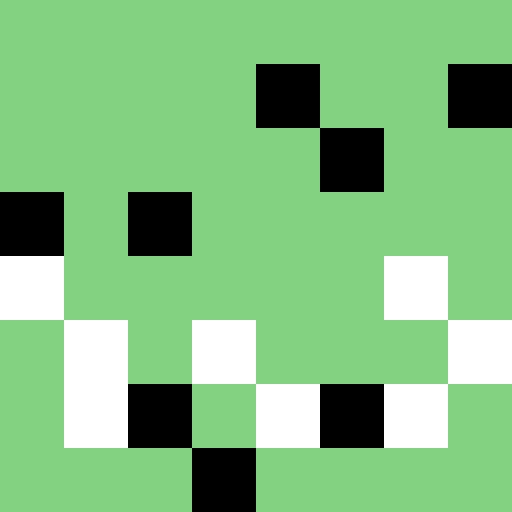}
  \end{subfigure}
  \hfill 
  \begin{subfigure}{0.237\columnwidth}
    \includegraphics[width=\textwidth]{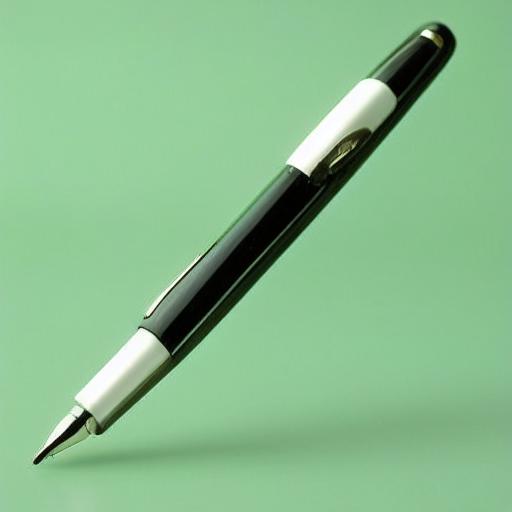}
  \end{subfigure}
  \begin{subfigure}{0.237\columnwidth}
    \includegraphics[width=\textwidth]{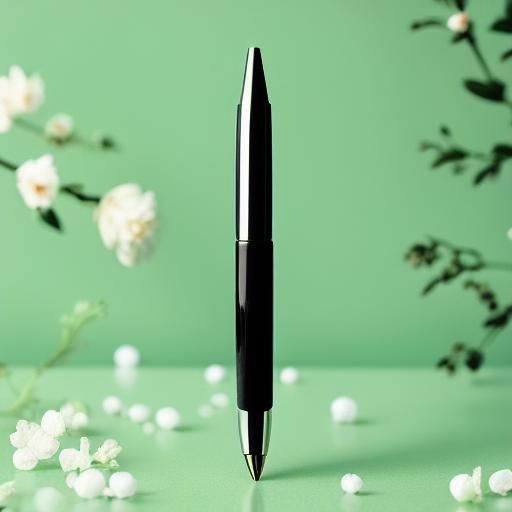}
  \end{subfigure}
  \hfill 
  \begin{subfigure}{0.237\columnwidth}
    \includegraphics[width=\textwidth]{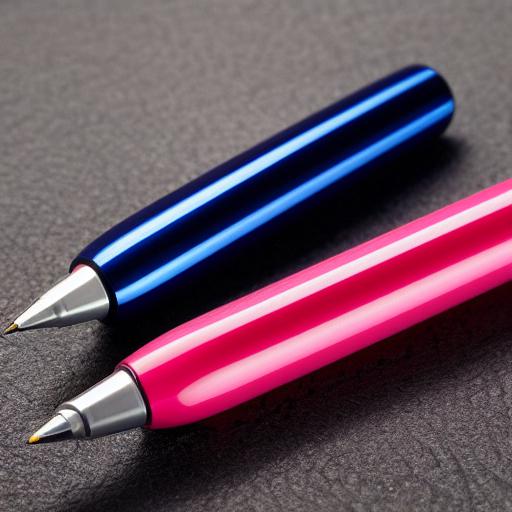}
  \end{subfigure}
  \\[0.6mm]
  \subfloat[Condition]{\includegraphics[width=0.237\columnwidth]{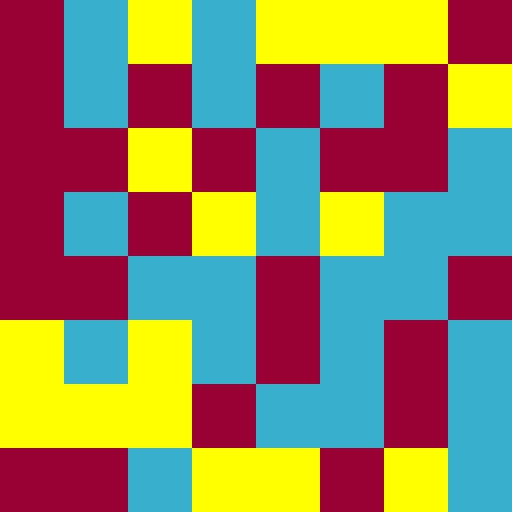}}
  \hfill 
  \subfloat[Ours]{\includegraphics[width=0.237\columnwidth]{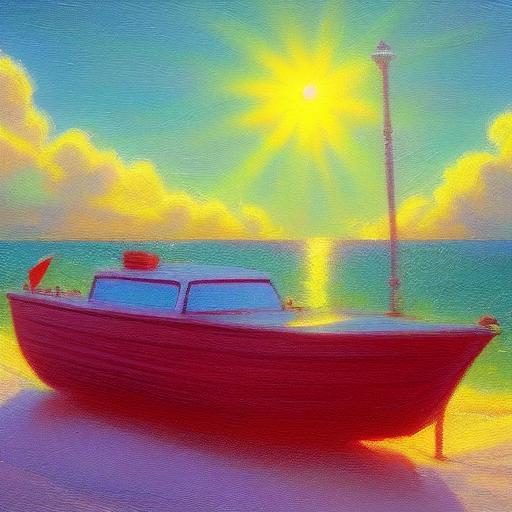} \includegraphics[width=0.237\columnwidth]{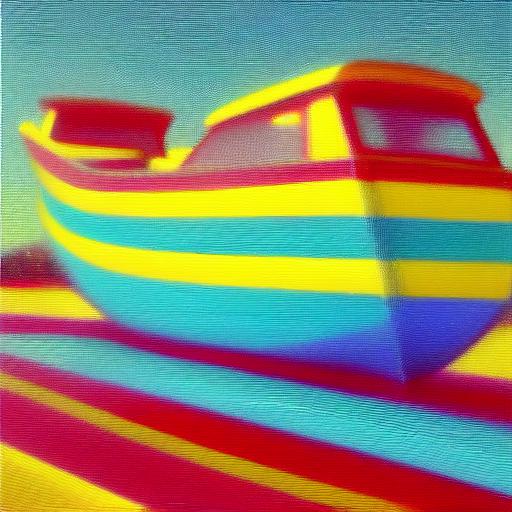}}
  \hfill 
  \subfloat[Baselines]{\includegraphics[width=0.237\columnwidth]{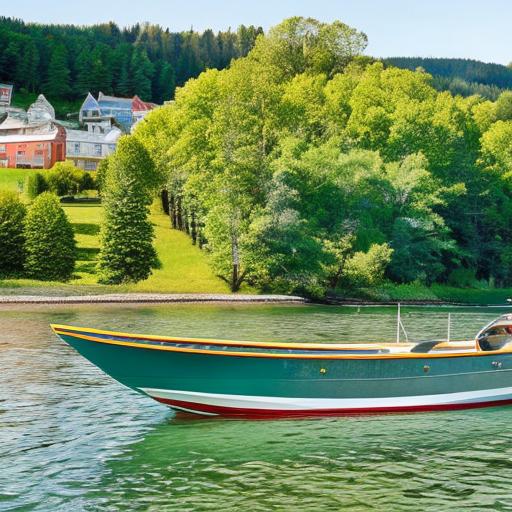}}

\caption{\textbf{Examples of color-conditioned generation} from different methods. Color conditions (a) can be derived from imagery (top two rows) or manual drawing (bottom two rows). Input text prompts are ``painting'' and ``pen'' (1st \& 3rd rows), and ``boat'' (2nd \& 4th rows). Our method (b) can generate pixels closely aligned with the color conditions and effectively disentangle the colors into different structures. Existing baselines~\cite{hertz2024style,ye2023ip,rombach2022high,zhang2023adding} (from top to bottom) (c) struggle with either the semantics (the first two rows) or colors (the last two rows) of generated contents.}
\label{fig:teaser}
\end{figure}

Different attempts of color-conditioned image generation are visualized in~\cref{fig:teaser}, where previous works often produce suboptimal results that do not align with the aforementioned criteria. In this paper, we propose an effective yet control-easy method facilitating color alignment in diffusion models for color-conditioned image synthesis. Specifically, our method accepts a color condition via a color pattern, determining color values and their proportions in a synthesized image. Different from the regular diffusion approach, we perform the color alignment on intermediate samples or latent representations during the diffusion process prior to inputting them into a denoising model for image synthesis, thereby facilitating the perception of the conditional colors in diffusion models while preserving the overall attribution of the diffusion process. Furthermore, our method can support various settings: re-training of a color-aligned diffusion model; fine-tuning of a color-aligned diffusion model from a pre-trained regular diffusion model; and zero-shot approximation (training-free) of the color alignment process with a pre-trained diffusion model. We validate our method by demonstrating its capabilities across various color conditions and compare it with existing baselines on benchmark datasets. In summary, we make the following contributions:
\begin{itemize}
    \item We address a challenging task: fine-grained color-conditioned image synthesis. We aim to generate images that closely follow a specified color pattern that defines color values and proportions without requiring an explicit structure. Our approach thus enhances the controllability and flexibility of image generation.
    
    \item We propose a novel color alignment method that operates on the diffusion process of diffusion models, allowing the diffusion to effectively perceive color conditions while maintaining the quality and diversity of generated contents. We derive various settings including re-training, fine-tuning, and a zero-shot approximation, to adapt our color alignment method into different down-stream needs.
    
    \item We conduct extensive experiments to validate our proposed method in various scenarios and controllability settings. Our results demonstrate state-of-the-art performance compared with existing baselines.
\end{itemize}

\section{Related work}
\paragraph{Neural image synthesis} has started with Variational Autoencoders (VAEs)~\cite{kingma2013auto,van2017neural,sohn2015learning} and Generative Adversarial Networks (GANs)~\cite{goodfellow2020generative,mirza2014conditional,isola2017image,karras2019style}, followed then by transformer-based methods~\cite{dosovitskiy2020image,liu2021swin} and diffusion models~\cite{ho2020denoising,song2020denoising,nichol2021improved}. Among these, Denoising Diffusion Probabilistic Models (DDPM)~\cite{ho2020denoising} have attracted considerable attention due to their generation quality and ease of extension. These models have been further developed into text-to-image (T2I) diffusion techniques~\cite{rombach2022high,saharia2022photorealistic,ramesh2022hierarchical}, which have been boosted recently to achieve greater controllability through incorporation of extra conditions~\cite{saharia2022palette,zhang2023adding,brooks2023instructpix2pix,ruiz2023dreambooth}. Our method follows the general diffusion framework with a focus on fine-grained color-conditioned image synthesis.

\paragraph{Reference-based conditional diffusion.} An intuitive condition for diffusion models is the use of an image reference. Structural conditions, such as edge maps and semantic maps, have been recently studied in~\cite{meng2021sdedit,saharia2022palette,lugmayr2022repaint,zhang2023adding,xie2023smartbrush,mou2024t2i}. These methods strictly apply spatial constraints given in the reference to align generated pixels. A more relaxed approach is image editing~\cite{hertz2022prompt,brooks2023instructpix2pix,kawar2023imagic,mokady2023null,cao2023masactrl}, where the structure and appearance in the reference loosely guide the generation of target images. Those works aim to achieve shifts in appearance while permitting a small range of spatial edits, such as changes in the size and pose of objects. Several methods~\cite{ye2023ip,ruiz2023dreambooth,hertz2024style} condition only on abstract concepts in the reference, such as image/object style, enabling more creative synthesis. In this work, we disentangle colors in the reference from their original spatial structure while preserving the color values and proportions. Our method allows a variety of user-provided color materials to construct diverse structures.

\paragraph{Sample altering in diffusion.} As indicated in~\cite{meng2021sdedit}, diffusion from the same noisy sample can result in similar outputs. Diffusion inversion~\cite{mokady2023null} is commonly used to construct samples, used directly~\cite{mokady2023null,parmar2023zero} or mapped to another diffusion process~\cite{zhang2023inversion,chung2024style,mou2024diffeditor} to diverge outputs. Several methods update intermediate latent representations with gradients calculated from external modules, e.g., other networks~\cite{yu2023freedom}, images~\cite{shi2024dragdiffusion}, features~\cite{mou2023dragondiffusion}. The Gaussian noise is defined~\cite{everaert2023diffusion} using estimated mean and variance to improve model convergence. Cold Diffusion~\cite{bansal2024cold} replaces the noising process with image degradations such as blurring and masking.

\section{Proposed method}

We first in~\cref{sec:method:diffusion_process} review the regular diffusion process. In~\cref{sec:method:align_image_space}, we introduce our color alignment technique under its simplest setting, i.e., re-training of a diffusion model with color alignment. We then in~\cref{sec:method:align_latent_space} extend our technique to latent diffusion models, where we address typical challenges in latent space diffusion and demonstrate our method's ability via fine-tuning of a pre-trained latent diffusion model to a color-aligned one. Finally, in~\cref{sec:method:zeroshot_align}, we present a zero-shot variant of our method that achieves similar generation quality without extra training. We provide an overview of our pipeline and the regular diffusion in~\cref{fig:pipeline}.

\subsection{Diffusion preliminaries}
\label{sec:method:diffusion_process}

Diffusion models~\cite{ho2020denoising} could be seen as a variant of VAEs~\cite{kingma2013auto} that map a data distribution $\mathcal{X}(\mathbf{x}_0)$ from a Gaussian distribution $\mathcal{N}(\mathbf{0}, \mathbf{I})$. In the forward process, a sample $\mathbf{x}_t$ is encoded by adding a Gaussian noise to $\mathbf{x}_0$:
\begin{equation}
    \mathbf{x}_t = \sqrt{\Bar{\alpha}_t} \mathbf{x}_0 + \sqrt{1 - \Bar{\alpha}_t} \bm{\epsilon}~,~~~ \bm{\epsilon} \sim \mathcal{N}(\mathbf{0}, \mathbf{I})
    \label{eq:regular_forward}
\end{equation}
where $t \in [1, T]$ is the time step associated with a noise strength $\Bar{\alpha}_t$ where $(1 - \Bar{\alpha}_{1}) \approx 0$ and $(1 - \Bar{\alpha}_{T}) \approx 1$.

Then, a denoising model $\bm{\epsilon}_{\theta}$ with parameters $\theta$ learns to reverse the process in~\cref{eq:regular_forward} by training with a loss $\mathcal{L}$ as,
\begin{align}
    \mathcal{L} & = \nabla_{\theta} \|~ \bm{\epsilon} - \bm{\epsilon}_{\theta}(\mathbf{x}_t, t) ~\|^2_2 \label{eq:regular_train} \\
    \hat{\mathbf{x}}_0 & = \frac{1}{\sqrt{\Bar{\alpha}_{t}}}(\mathbf{x}_t - \sqrt{1-\Bar{\alpha}_{t}} \bm{\epsilon}_{\theta}(\mathbf{x}_t, t)) \label{eq:regular_pred} \\
    \mathbf{x}_{t-1} & = \frac{\sqrt{\Bar{\alpha}_{t-1}} \beta_t}{1-\Bar{\alpha}_{t}} \hat{\mathbf{x}}_0 + \frac{\sqrt{\alpha_{t}} (1-\Bar{\alpha}_{t-1})}{1-\Bar{\alpha}_{t}} \mathbf{x}_{t} + \sigma_t \bm{\varepsilon} \label{eq:regular_step}
\end{align}
where $\hat{\mathbf{x}}_0$ is the predicted sample. $\alpha, \beta, \sigma \in \mathbb{R}$, and $\bm{\varepsilon} \sim \mathcal{N}(\mathbf{0}, \mathbf{I})$ are scheduler-related parameters~\cite{ho2020denoising}.

\subsection{Color-aligned diffusion in image space}
\label{sec:method:align_image_space}

As presented in~\cref{sec:method:diffusion_process}, the regular diffusion model learns to reconstruct $\mathbf{x}_0$ from $\mathcal{N}(\mathbf{0}, \mathbf{I})$ during training and allows sampling-free generation in testing, i.e., unconditional generation. To condition the diffusion process with color conditions, existing methods pass color text prompts~\cite{rombach2022high,ye2023ip} or referenced images~\cite{ye2023ip,hertz2024style,zhang2023adding} in parallel with $\mathbf{x}_t$ untouched. We found this approach loosely constrains the sampled data manifold, i.e., $\hat{\mathbf{x}}_0$ can be sampled with out-of-bound colors. This is because the conditions are not involved in the sampling process (in construction of $\mathbf{x}_{t}$) and the denoising model $\bm{\epsilon}_{\theta}$ hence leans towards the unconditional generation. Moreover, colors are learned in conjunction with spatial information given in referenced images, limiting the diversity of generated contents (see~\cref{fig:teaser}).

In this paper, we take a color pattern $\mathbf{c}$, determining color values and their proportions in synthesis, as a color condition. The color pattern can be an imagery or hand-drawing (see~\cref{fig:teaser}). To effectively constrain the sampling manifold with color condition, we involve the color pattern $\mathbf{c}$ into the diffusion process via a new denoising model $\bm{\epsilon}_{\theta}(\mathbf{x}_t, t, \mathbf{c})$. In addition, we relax the spatial information in $\mathbf{c}$ by using a color alignment function $f(\mathbf{x}_{t},\mathbf{c})$ that produces a color-aligned image from an image $\mathbf{x}_{t}$ and color pattern $\mathbf{c}$ as,
\begin{align}
    f(\mathbf{x}_{t},\mathbf{c})[p] = \argmin_{\mathbf{c}[q]}\| \mathbf{x}_{t}[p] - \mathbf{c}[q] \|^2_2
    \label{eq:color_alignment}
\end{align}
where $\mathbf{x}_t[p]$ and $\mathbf{c}[q]$ are colors at pixels $p$ and $q$.

\begin{figure}[t]
  \subfloat[Regular diffusion~\cite{rombach2022high}]{\includegraphics[width=0.49\columnwidth,trim={1mm 90mm 202mm 0mm},clip]{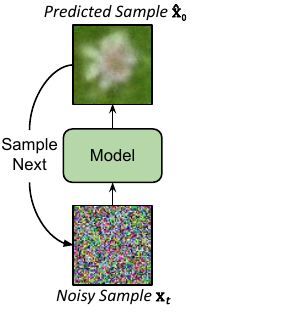}\label{fig:pipeline:a}}
  \hfill
  \subfloat[Regular diffusion with condition]{\includegraphics[width=0.49\columnwidth,trim={1mm 90mm 202mm 0mm},clip]{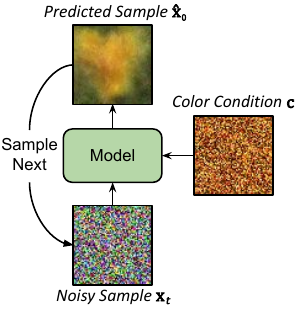}\label{fig:pipeline:b}}
  \\[0.6mm]
  \subfloat[Our color-aligned diffusion for re-training or fine-tuning]{\includegraphics[width=0.49\columnwidth,trim={1mm 72mm 202mm -3mm},clip]{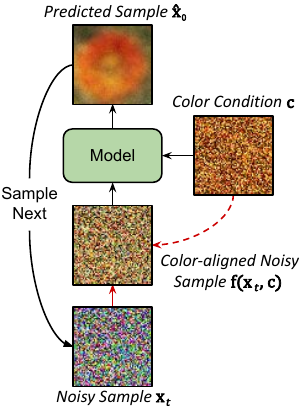}\label{fig:pipeline:c}}
  \hfill
  \subfloat[Our zero-shot approximation of the color-aligned diffusion]{\includegraphics[width=0.49\columnwidth,trim={1mm 72mm 202mm -3mm},clip]{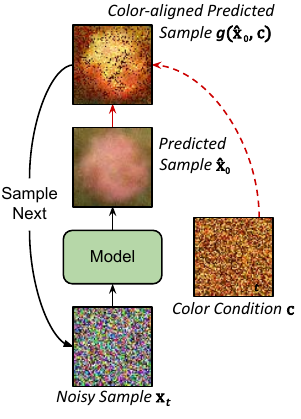}\label{fig:pipeline:d}}
\caption{\textbf{Pipelines} of the regular diffusion (a) and its color-conditioned version (b), our color-aligned diffusion (c), where noisy samples $\mathbf{x}_t$ are aligned with conditioned colors, and our zero-shot version (d), where color alignment is directly applied to predicted sample $\hat{\mathbf{x}}_0$.}
\label{fig:pipeline}
\end{figure}

We visualize the diffusion process in~\cref{fig:qualitative_initial}. Our color alignment technique benefits the sampling in several aspects. First, it enforces the generated pixel colors within the target space, guaranteeing the \textit{accuracy}. Second, its non-spatial manner allows pixels distribute more freely, fulfilling the color \textit{disentanglement}. For the sake of \textit{completeness}, we further concatenate $\mathbf{c}$ to the model's input as an extra hint. However, to completely break down the spatial information, we opt to randomly permutate the pixel locations in $\mathbf{c}$, resulting in $\psi(\mathbf{c})$ with $\psi$ being an image permutation operator. Such a permutation allows us to directly take $\mathbf{x}_0$ as color condition in training time, avoiding extra data requirement and data leakage. Finally, training and sampling of our color-aligned diffusion process (\cref{fig:pipeline:c}) are formulated as:
\begin{align}
\begin{split}
    \mathcal{L} & = \nabla_{\theta} \|~ \bm{\epsilon}' - \bm{\epsilon}_{\theta}(f(\mathbf{x}_t, \mathbf{c}), t, \mathbf{c}) ~\|^2_2 \\
    \bm{\epsilon}' & = \frac{f(\mathbf{x}_t, \mathbf{c}) - \sqrt{\Bar{\alpha}_t} \mathbf{x}_0}{\sqrt{1 - \Bar{\alpha}_t}},~~\mathbf{c} = \psi(\mathbf{x}_0) \label{eq:ours_train}
\end{split}
\\
\begin{split}
    \hat{\mathbf{x}}_0 & = \frac{1}{\sqrt{\Bar{\alpha}_{t}}}(f(\mathbf{x}_t, \mathbf{c}) - \sqrt{1-\Bar{\alpha}_{t}} \bm{\epsilon}_{\theta}(f(\mathbf{x}_t, \mathbf{c}), t, \mathbf{c})) \label{eq:ours_pred}
\end{split}
\end{align}
where $\bm{\epsilon}'$ is the adapted noise complemented with the effect of $f$. Notably, we alter only what $\bm{\epsilon}_{\theta}$ sees and predicts. The parameters $\theta$ thus can be learned more easily with color condition compared with the regular one. All other non-learnable steps, e.g., the regular forward process and~\cref{eq:regular_step}, remain unchanged for the ease of sampling. 
For more technical visualizations, please refer to the supplementary material.

\subsection{Color-aligned diffusion in latent space}
\label{sec:method:align_latent_space}
Training of diffusion models in their latent space is a common practice. In this section, we extend our color alignment technique to latent diffusion models via fine-tuning of a pre-trained latent model to achieve color alignment.

Following the Stable Diffusion in~\cite{rombach2022high}, we encode $\mathbf{x}_{0}$ into its latent representation $\mathbf{z}_{0}$. 
We then train our model with alignment in the latent space, i.e., replacing $\mathbf{x}$ by $\mathbf{z}$ in~\cref{eq:ours_train,eq:ours_pred}.
However, when high-frequency structural information is present in $\mathbf{x}_{0}$, unintended colors and local structural information can be encoded in $\mathbf{z}_{0}$.
We discovered that blurring $\mathbf{x}_{0}$ prior to encoding it enhances the color \textit{disentanglement} and \textit{accuracy}, with minimal loss of the \textit{completeness} in generated results. 
Also, we found that suspending color alignment at late time steps (e.g., $t < 0.2T$ where diffusion model is fixing details) allows a proper refinement of the latent, making latent decoding towards more photo-realistic details.

\subsection{Zero-shot color-aligned diffusion}
\label{sec:method:zeroshot_align}
To mitigate the computational burden of training, we introduce a zero-shot approximation of our color alignment method.
As shown in~\cref{eq:ours_train,eq:ours_pred}, the training aims to reconstruct $\hat{\mathbf{x}}_0$ whose color distribution follows the color condition $\mathbf{c} = \psi(\mathbf{x}_0)$. 
Under the zero-shot setting, to achieve this ability without learning the parameters $\theta$, we directly map an unconditionally generated $\hat{\mathbf{x}}_0$ (from~\cref{eq:regular_pred}) to $\mathbf{c}$ during the sampling process using a color alignment function $g(\hat{\mathbf{x}}_0,\mathbf{c})$:
\begin{align}
    g(\hat{\mathbf{x}}_0,\mathbf{c}) = \argmin_{\psi(\mathbf{c})}\| \hat{\mathbf{x}}_0 - \psi(\mathbf{c}) \|^2_2
    \label{eq:color_alignment_one_shot}
\end{align}
where we map the unconditional $\hat{\mathbf{x}}_0$ to a closest reorganization of the conditional colors $\psi(\mathbf{c})$ for color alignment.

Unlike $f$ defined in~\cref{eq:color_alignment}, $g$ in~\cref{eq:color_alignment_one_shot} performs one-to-one mapping, i.e., each location in $\mathbf{c}$ is used for color alignment exactly once. Note that, like $f$, our $g$ also applies non-spatial mapping. This would ensure desired properties (color \textit{accuracy}, \textit{completeness}, and \textit{disentanglement}) from the zero-shot color-aligned diffusion. However, since $g$ is not involved in the learning process, the color-aligned image $g(\hat{\mathbf{x}}_0, \mathbf{c})$ could be messy, showing invalid structures and outliers (see~\cref{fig:pipeline:d}). Fortunately, we observed that pre-trained diffusion models have capability of clustering semantics and eliminating outliers at early sampling steps (e.g., $t > 0.2T$). We also suspend the color alignment at late time steps to prevent excessive color changes and allow the model to refine details. Additionally, such zero-shot approximation is extendable to latent diffusion models. We found the same conclusion for using the blurring strategy as in~\cref{sec:method:align_latent_space} applies.  

Another interest could be replacing $f$ by $g$ in learning. However, we find this not applicable because $g(\mathbf{x}_{t}, \mathbf{c})$ corrupts the distribution of $\mathbf{x}_{t}$ too heavily. Its subsequent adapted noise $\bm{\epsilon}'$ (the learning target in~\cref{eq:ours_train}) deviates too far from the standard Gaussian $\mathcal{N}(\mathbf{0}, \mathbf{I})$, eliminating the effect of a proper diffusion process.

\section{Experiments}
\subsection{Datasets}
\label{sec:datasets}
We experimented our method in two scenarios: color-aligned diffusion in image space and color-aligned diffusion in latent space. For diffusion in image space, we ran experiments on Oxford-flower~\cite{nilsback2008automated} (7.17k training and 1.02k test images) and Microsoft-emoji~\cite{microsoft2021emoji} datasets (6.80k training and 0.76k test images) at resolution 64 $\times$ 64. For latent diffusion, we experimented on a sub-set of a high-quality generic dataset Text-to-image-2M~\cite{huggingface2021t2m} (300k training and 20k test images) at resolution 512 $\times$ 512. To validate the color \textit{disentanglement} ability of our method, we replaced original prompts in Text-to-image-2M~\cite{huggingface2021t2m} with 50 daily object prompts randomly generated by ChatGPT~\cite{achiam2023gpt} (e.g., those in~\cref{fig:qualitative_baseline}).

\subsection{Baselines}
\label{sec:baselines}
Our color-aligned image diffusion version is adapted from the DDPM~\cite{ho2020denoising}. Therefore, we compare our method with the DDPM to demonstrate its extra color-conditioned ability without requiring additional data creation or labeling. For the same reason, we also compare our latent diffusion version with the pre-trained Stable Diffusion in~\cite{rombach2022high}.

Given that our color reference can be an in-the-wild image or a manual drawing, we compare our method with SOTA reference-based generation methods. IP-Adapter~\cite{ye2023ip} and T2I-Adapter~\cite{mou2024t2i} convert an image reference into features, which are subsequently combined with the model features for image generation. Style-Aligned~\cite{hertz2024style} transfers source image information from its attention layer to the target generation attention layer, thereby preserving spatial and style features. ControlNet-Color is a variant of ControlNet~\cite{zhang2023adding} that accepts a predefined pixelized color image as input and generates a refined version in which the colors are approximately aligned with the original image spatially. IP-Adapter~\cite{ye2023ip} and ControlNet-Color~\cite{zhang2023adding} require fine-tuning, while Style-Aligned~\cite{hertz2024style} operates as a zero-shot approach.

\begin{figure}[t]
    \subfloat[Our color-aligned image diffusion process. The top row shows our training inputs, created from early to late steps in the forward process. The bottom row is a test case from our reverse process, sampled from pure noise.]{\begin{minipage}[c]{\columnwidth}
    \centering
        \includegraphics[width=\columnwidth,trim={4mm 104mm 144mm 3mm},clip]{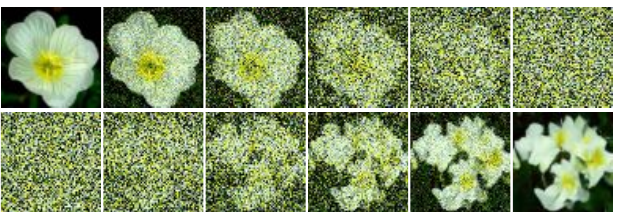}\label{fig:qualitative_initial:a}
    \end{minipage}}
    
    \subfloat[Regular image diffusion process in DDPM~\cite{ho2020denoising} in same order as (a).]{\begin{minipage}[c]{\columnwidth}
    \centering
        \includegraphics[width=\columnwidth,trim={4mm 104mm 144mm 3mm},clip]{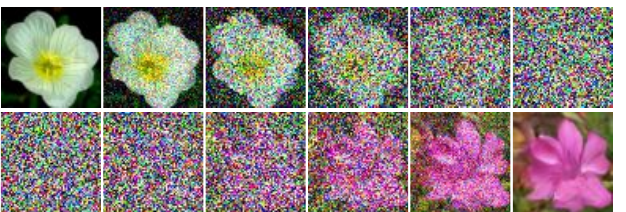}\label{fig:qualitative_initial:b}
    \end{minipage}}
    
    \subfloat[Several color-aligned image synthesis results. From top to bottom rows are condition, our generated results, and regular diffusion's results.]{\begin{minipage}[c]{\columnwidth}
    \centering
        \includegraphics[width=\columnwidth,trim={4mm 85mm 144mm 3mm},clip]{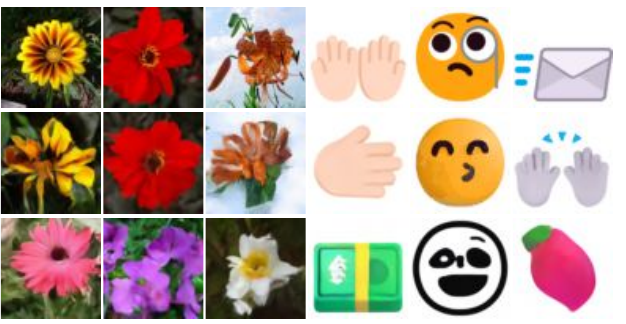}\label{fig:qualitative_initial:c}
    \end{minipage}}
    
    \caption{\textbf{Qualitative results of color-aligned image diffusion.} (a)-(b) Visualization of the diffusion process by our method and DDPM~\cite{ho2020denoising}. (c) Generation results of our method and DDPM~\cite{ho2020denoising}.}
    \label{fig:qualitative_initial}
\end{figure}

\subsection{Implementation details}
\label{sec:implementation_details}
We follow default settings in the DDPM~\cite{ho2020denoising} and the Stable Diffusion~\cite{rombach2022high} for network architectures and diffusion scheduling, specifically, from the huggingface-diffusers~\cite{von_Platen_Diffusers_State-of-the-art_diffusion}. Further details, e.g., DDPM re-training, Stable Diffusion fine-tuning are presented in our supplementary material.

We trained our image diffusion version for 100 epochs and latent diffusion version for 160k iterations. We performed inference at 50 time steps. We used classifier-free guidance~\cite{ho2022classifier} with guidance scale 5. All experiments were executed on two RTX 3090 GPUs with batch size 4.

\subsection{Qualitative study}
\label{sec:qualitative}
\noindent\textbf{Color-aligned image diffusion.}
We visually compare our method with the regular diffusion in DDPM~\cite{ho2020denoising} in~\cref{fig:qualitative_initial}. As shown, our method (\cref{fig:qualitative_initial:a}) well restricts the diffusion process to the conditional color space, wherein only whites, yellows, and background blacks are allowed. In contrast, the regular diffusion in DDPM (\cref{fig:qualitative_initial:b}) generates out-of-bounds colors during the diffusion process, leading to an unintended reconstruction result.

\cref{fig:qualitative_initial:c} illustrates an immediate application of our color-aligned image diffusion in image rephrasing, i.e., colors in an input image are reorganized by our model to construct new structures. More importantly, this conditional process does not require any additional data (e.g., labeled data or paired data) or pre-defined spatial information (e.g., semantic maps or masks) in both training or inference. This makes our technique scalable to larger datasets.

\begin{figure*}[t]
    \subfloat[Input Condition]{\label{fig:qualitative_baseline:a}\begin{minipage}[c]{\textwidth}
        \includegraphics[width=0.120\textwidth]{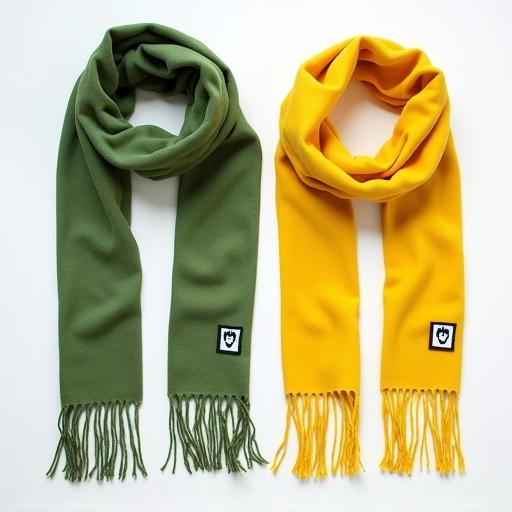}
        \hfill
        \includegraphics[width=0.120\textwidth]{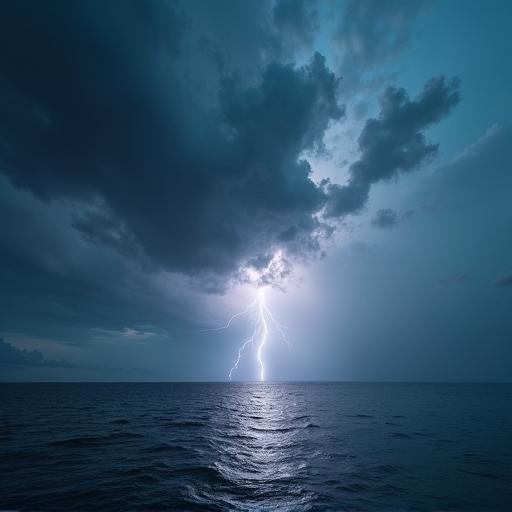}
        \hfill
        \includegraphics[width=0.120\textwidth]{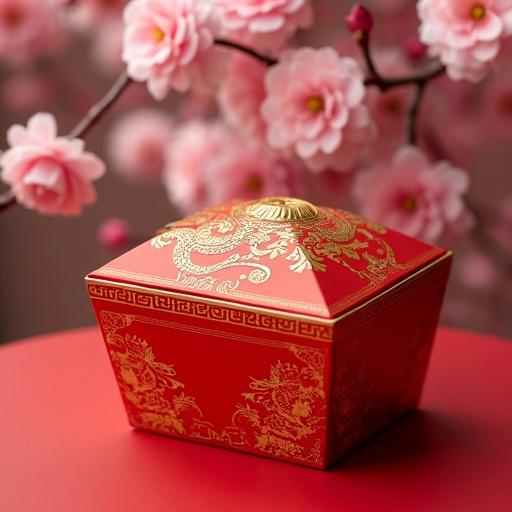}
        \hfill
        \includegraphics[width=0.120\textwidth]{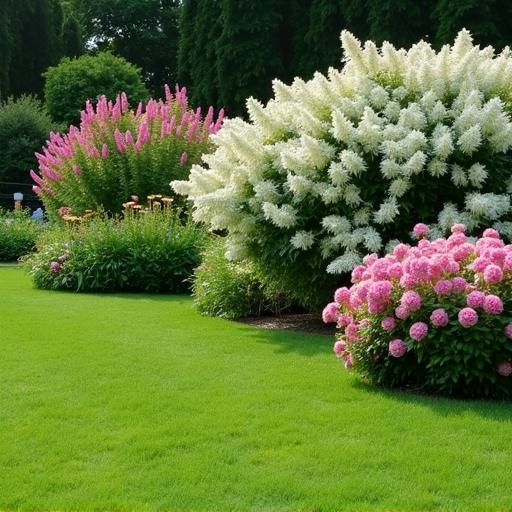}
        \hfill
        \includegraphics[width=0.120\textwidth]{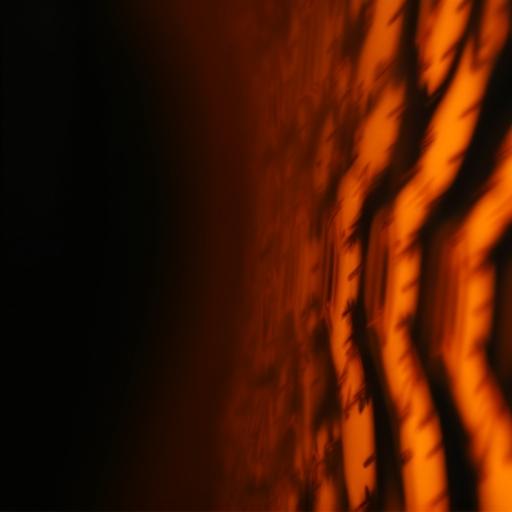}
        \hfill
        \includegraphics[width=0.120\textwidth]{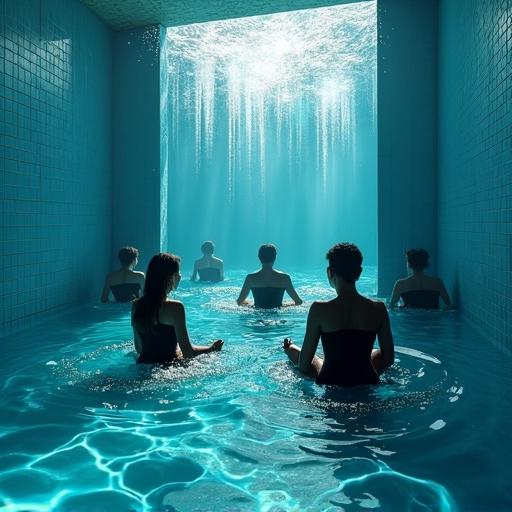}
        \hfill
        \includegraphics[width=0.120\textwidth]{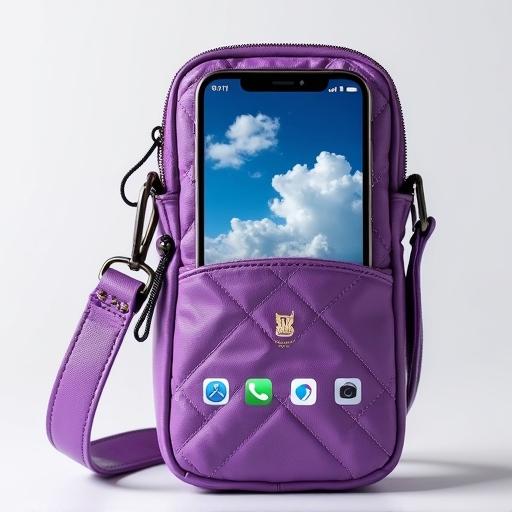}
        \hfill
        \includegraphics[width=0.120\textwidth]{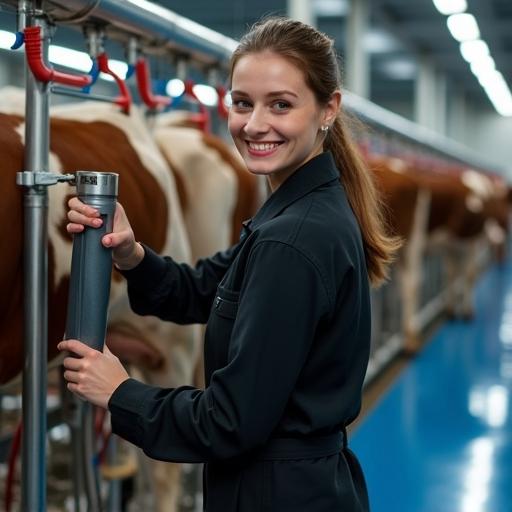}
        \smallskip
        \hspace{4.4mm}``wallet''\hspace{5.0mm}
        \hspace{6.2mm}``ring''\hspace{5.0mm}
        \hspace{5.7mm}``flower''\hspace{5.0mm}
        \hspace{6.3mm}``car''\hspace{5.0mm}
        \hspace{8.6mm}``pen''\hspace{5.0mm}
        \hspace{5.7mm}``basket''\hspace{5.0mm}
        \hspace{6.1mm}``map''\hspace{5.0mm}
        \hspace{7.2mm}``bus''\hspace{5.0mm}
    \end{minipage}}
    \smallskip

    \subfloat[Ours (Fine-tune)]{\label{fig:qualitative_baseline:b}\begin{minipage}[c]{\textwidth}
        \includegraphics[width=0.120\textwidth]{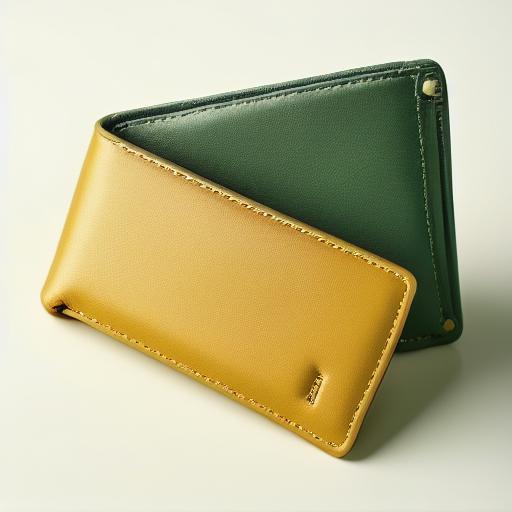}
        \hfill
        \includegraphics[width=0.120\textwidth]{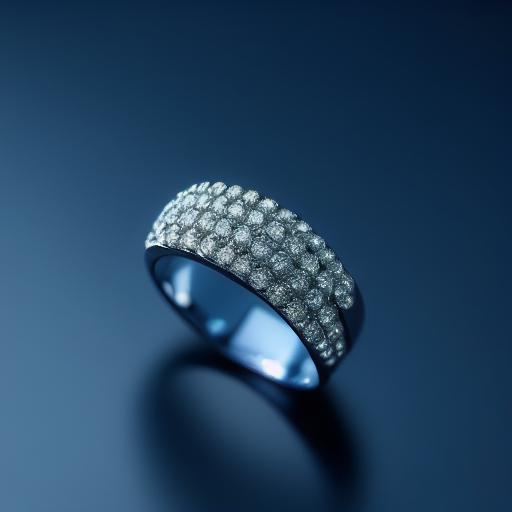}
        \hfill
        \includegraphics[width=0.120\textwidth]{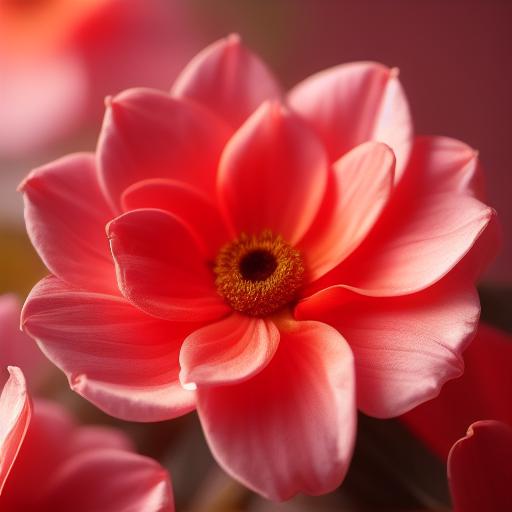}
        \hfill
        \includegraphics[width=0.120\textwidth]{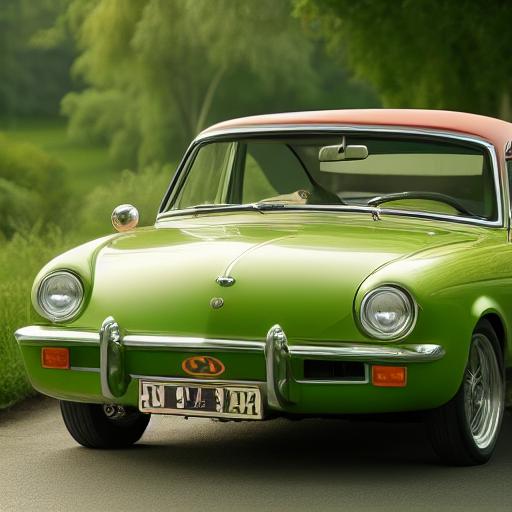}
        \hfill
        \includegraphics[width=0.120\textwidth]{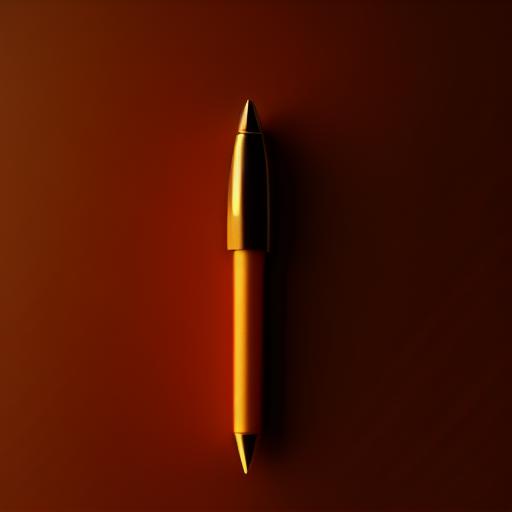}
        \hfill
        \includegraphics[width=0.120\textwidth]{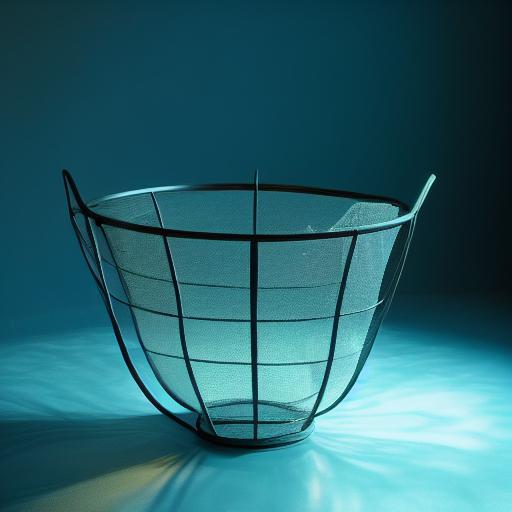}
        \hfill
        \includegraphics[width=0.120\textwidth]{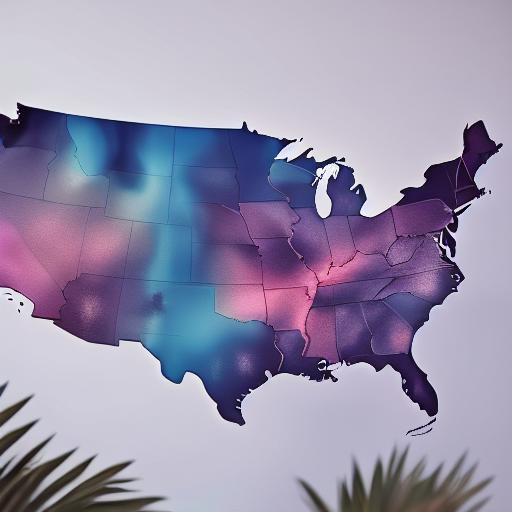}
        \hfill
        \includegraphics[width=0.120\textwidth]{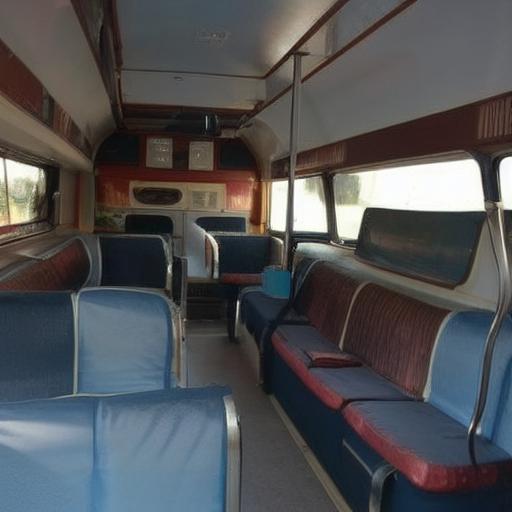}
    \end{minipage}}
    \smallskip

    \subfloat[Ours (Zero-shot)]{\label{fig:qualitative_baseline:c}\begin{minipage}[c]{\textwidth}
        \includegraphics[width=0.120\textwidth]{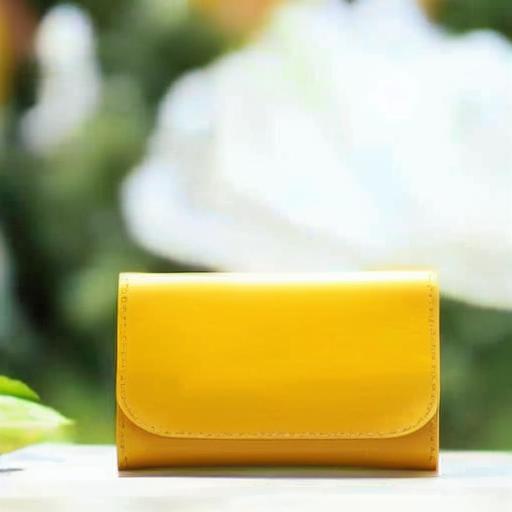}
        \hfill
        \includegraphics[width=0.120\textwidth]{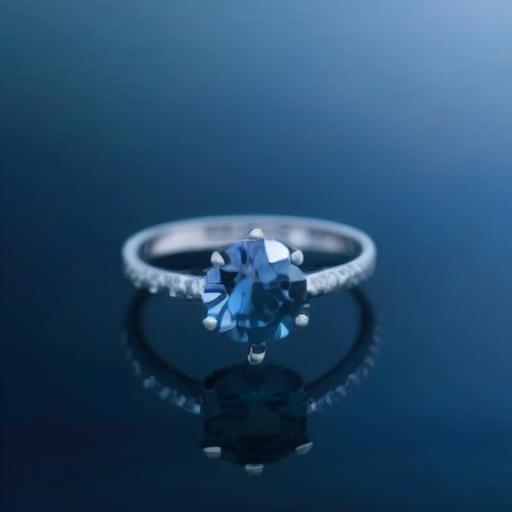}
        \hfill
        \includegraphics[width=0.120\textwidth]{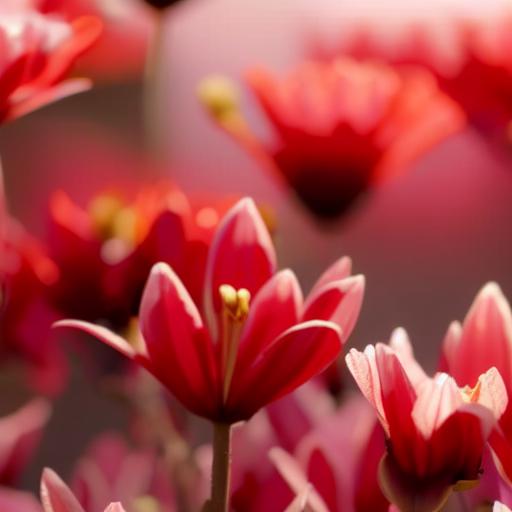}
        \hfill
        \includegraphics[width=0.120\textwidth]{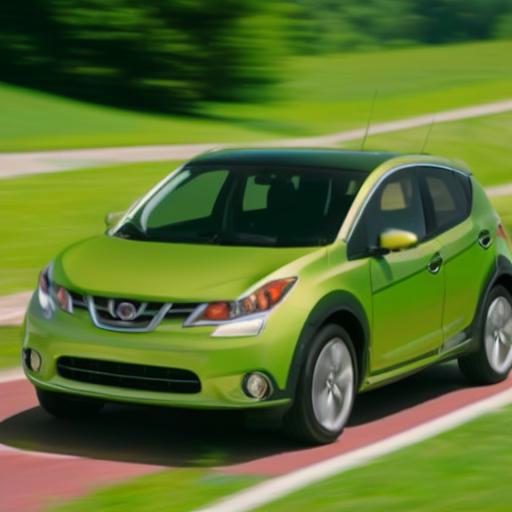}
        \hfill
        \includegraphics[width=0.120\textwidth]{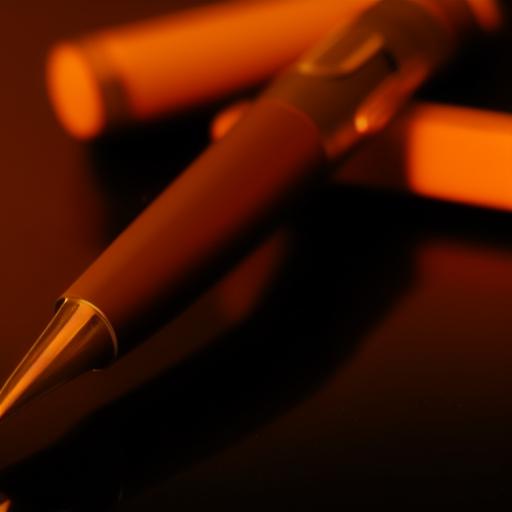}
        \hfill
        \includegraphics[width=0.120\textwidth]{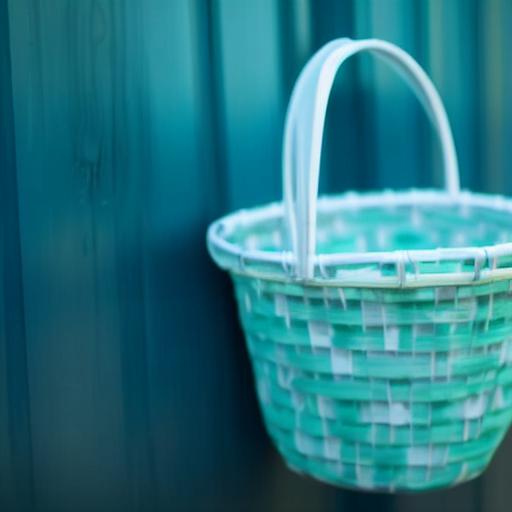}
        \hfill
        \includegraphics[width=0.120\textwidth]{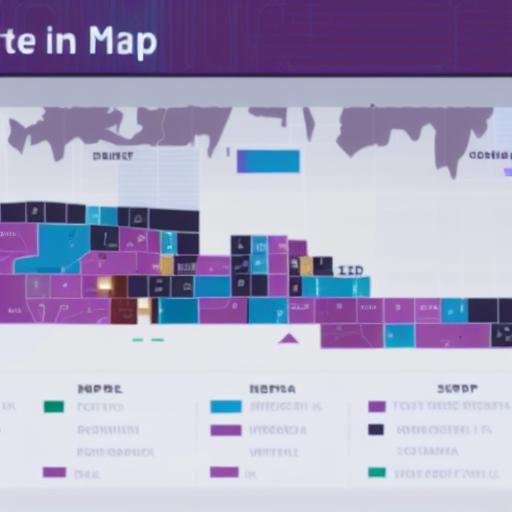}
        \hfill
        \includegraphics[width=0.120\textwidth]{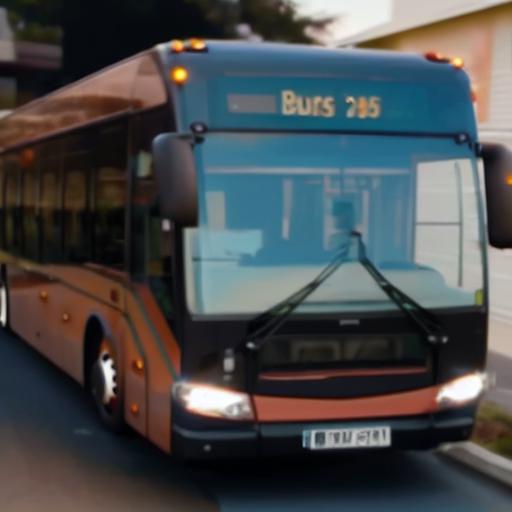}
    \end{minipage}}

    \subfloat[IP-Adapter~\cite{ye2023ip}]{\label{fig:qualitative_baseline:d}\begin{minipage}[c]{\textwidth}
        \includegraphics[width=0.120\textwidth]{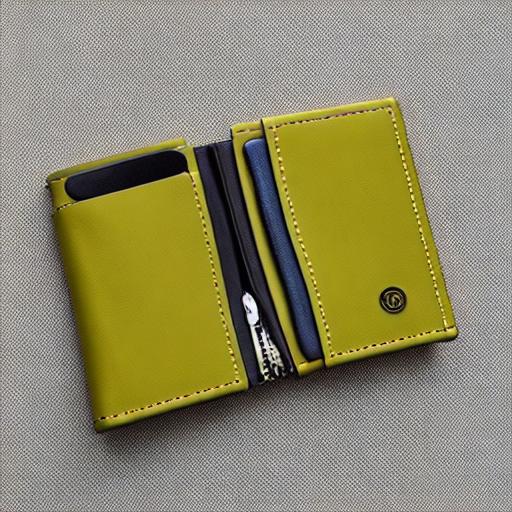}
        \hfill
        \includegraphics[width=0.120\textwidth]{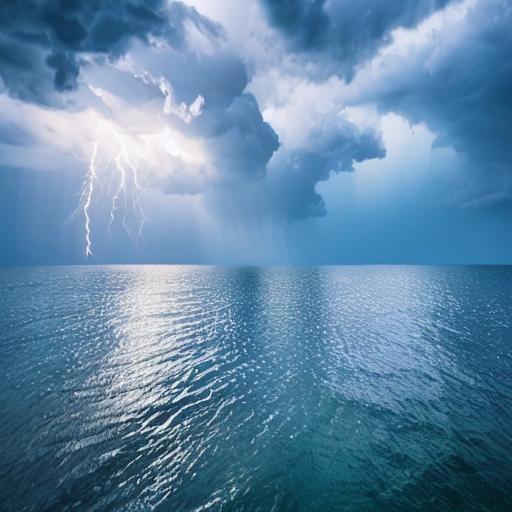}
        \hfill
        \includegraphics[width=0.120\textwidth]{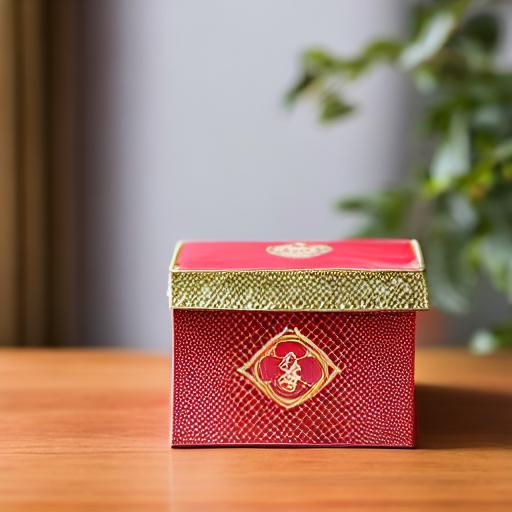}
        \hfill
        \includegraphics[width=0.120\textwidth]{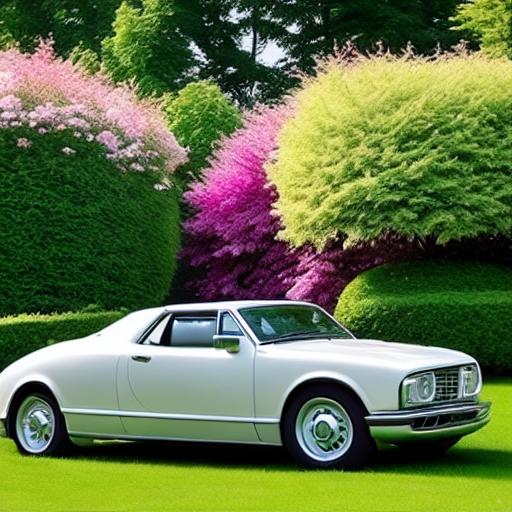}
        \hfill
        \includegraphics[width=0.120\textwidth]{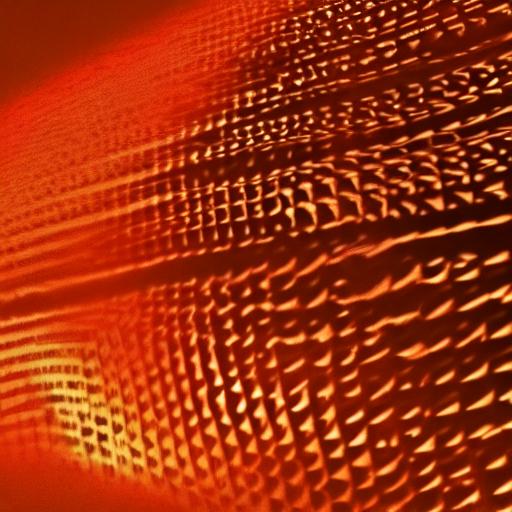}
        \hfill
        \includegraphics[width=0.120\textwidth]{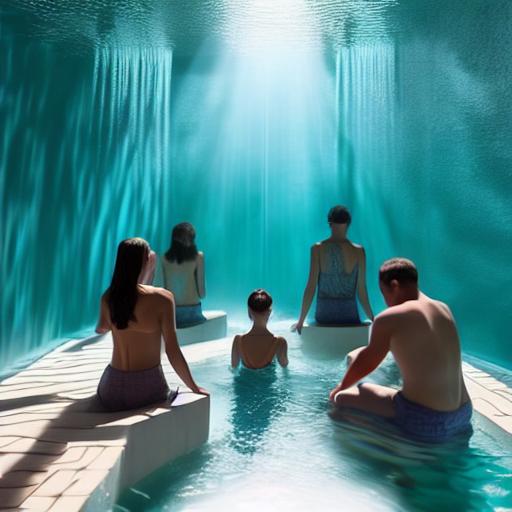}
        \hfill
        \includegraphics[width=0.120\textwidth]{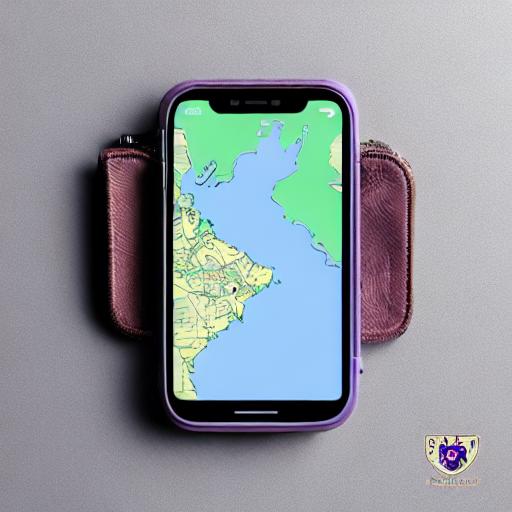}
        \hfill
        \includegraphics[width=0.120\textwidth]{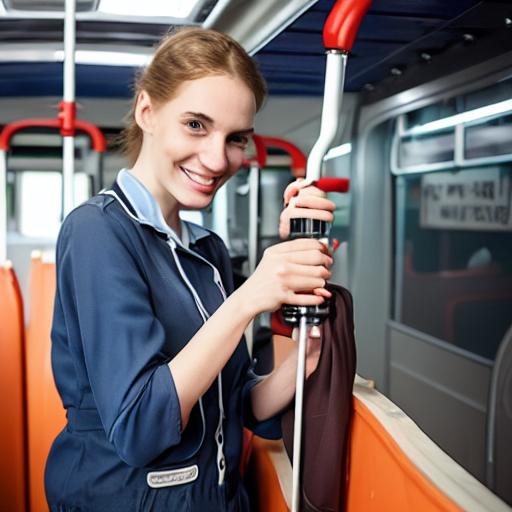}
    \end{minipage}}
    \smallskip

    \subfloat[T2I-Adapter~\cite{mou2024t2i}]{\label{fig:qualitative_baseline:e}\begin{minipage}[c]{\textwidth}
        \includegraphics[width=0.120\textwidth]{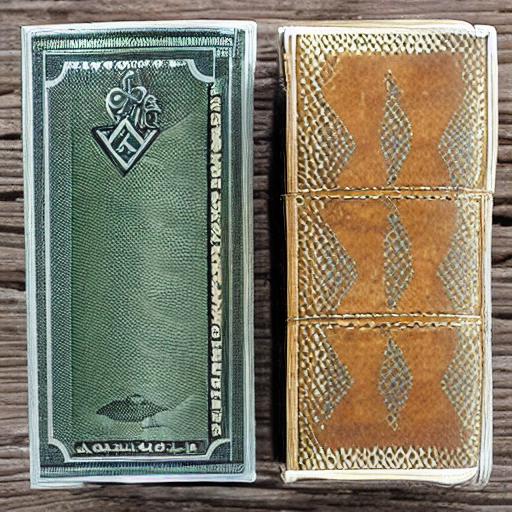}
        \hfill
        \includegraphics[width=0.120\textwidth]{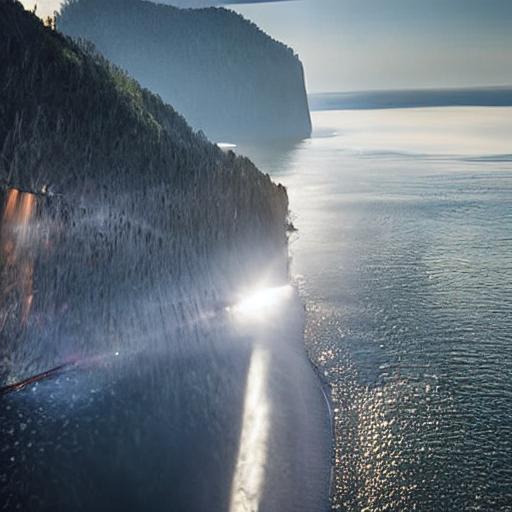}
        \hfill
        \includegraphics[width=0.120\textwidth]{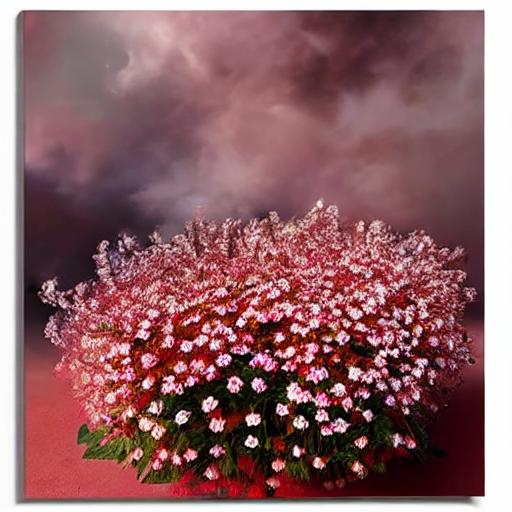}
        \hfill
        \includegraphics[width=0.120\textwidth]{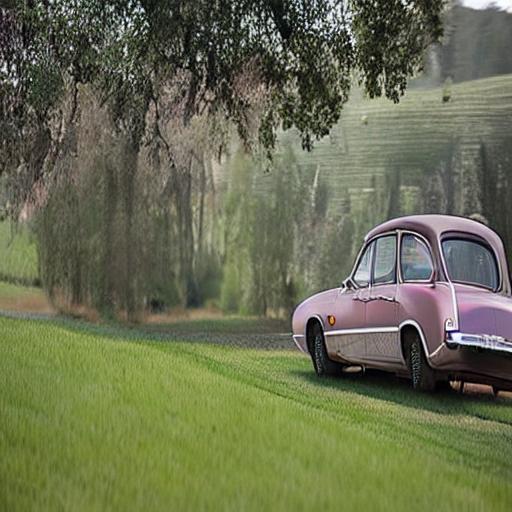}
        \hfill
        \includegraphics[width=0.120\textwidth]{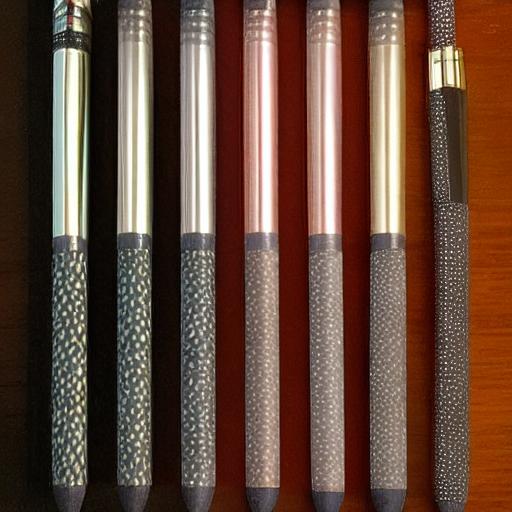}
        \hfill
        \includegraphics[width=0.120\textwidth]{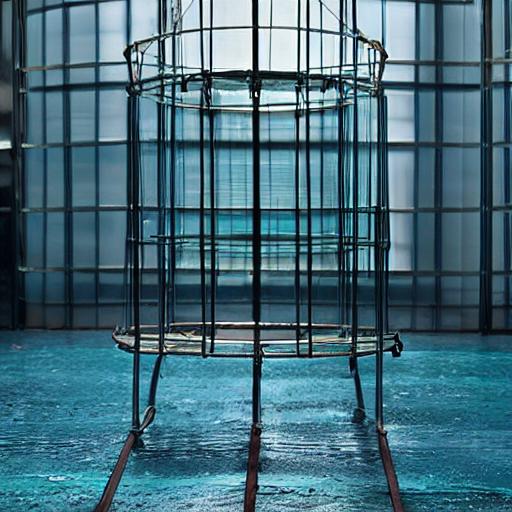}
        \hfill
        \includegraphics[width=0.120\textwidth]{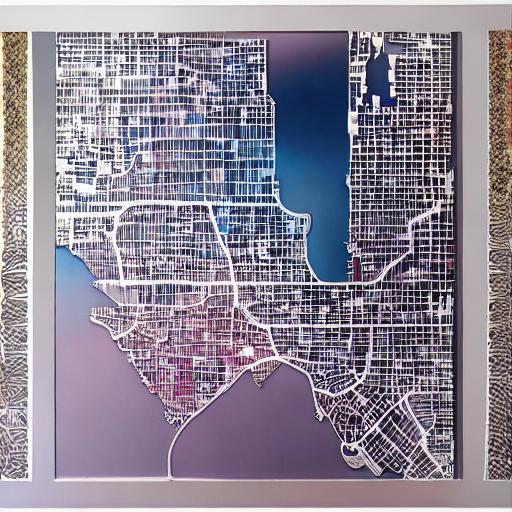}
        \hfill
        \includegraphics[width=0.120\textwidth]{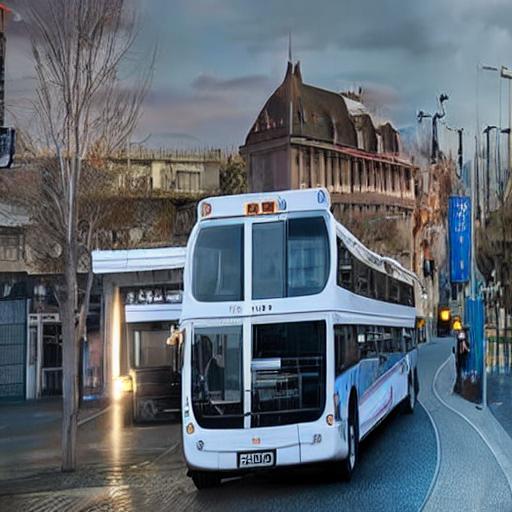}
    \end{minipage}}
    \smallskip

    \subfloat[Style-Aligned~\cite{hertz2024style}]{\label{fig:qualitative_baseline:f}\begin{minipage}[c]{\textwidth}
        \includegraphics[width=0.120\textwidth]{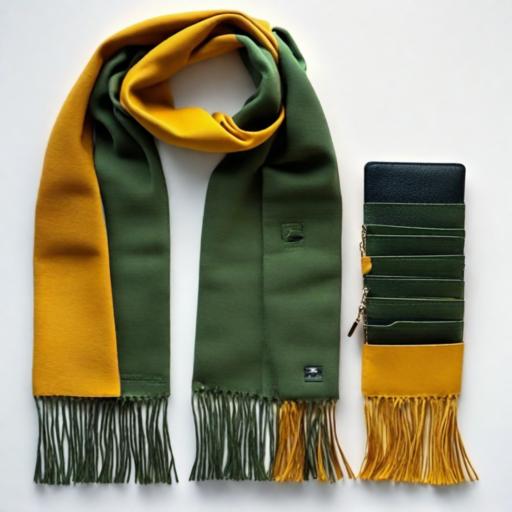}
        \hfill
        \includegraphics[width=0.120\textwidth]{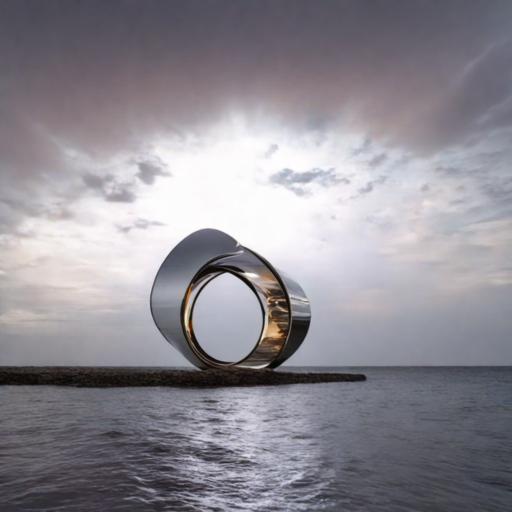}
        \hfill
        \includegraphics[width=0.120\textwidth]{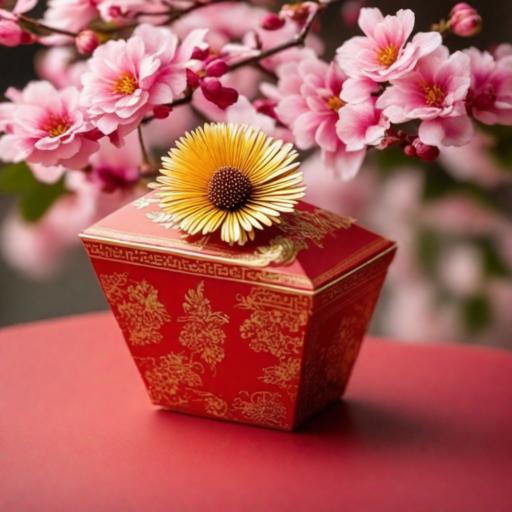}
        \hfill
        \includegraphics[width=0.120\textwidth]{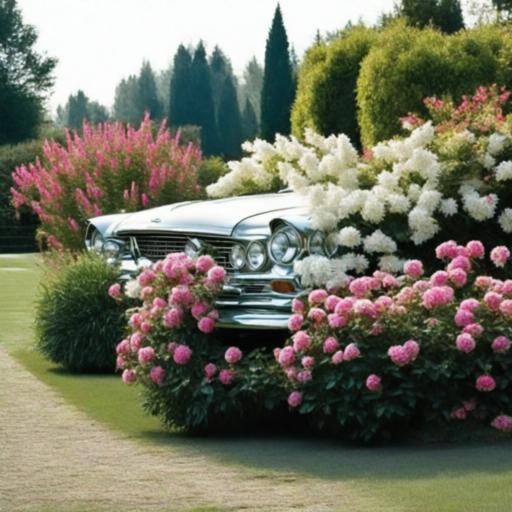}
        \hfill
        \includegraphics[width=0.120\textwidth]{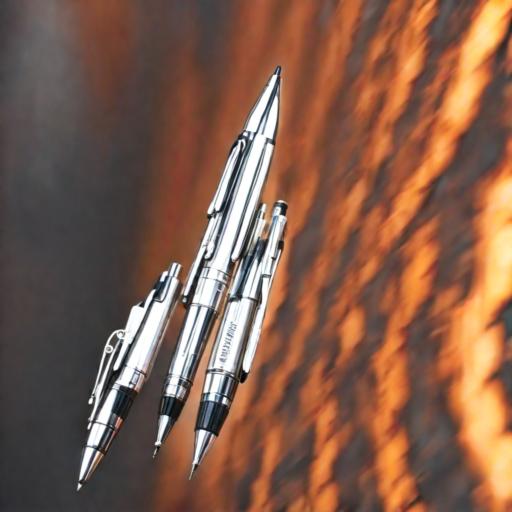}
        \hfill
        \includegraphics[width=0.120\textwidth]{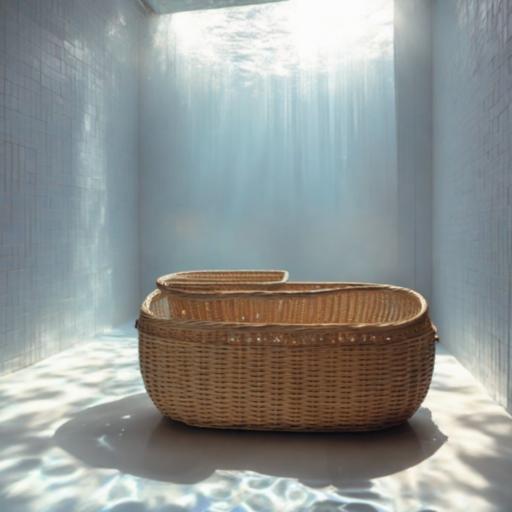}
        \hfill
        \includegraphics[width=0.120\textwidth]{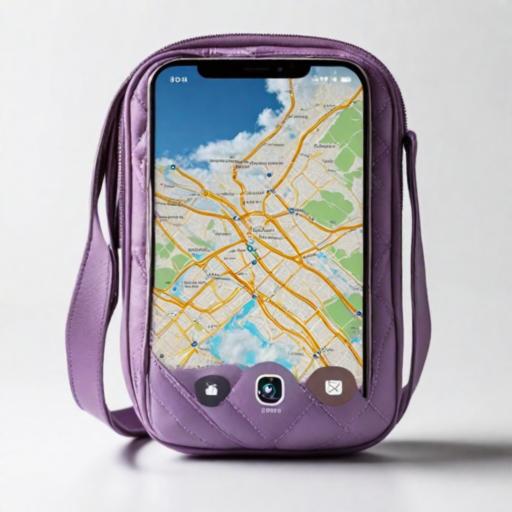}
        \hfill
        \includegraphics[width=0.120\textwidth]{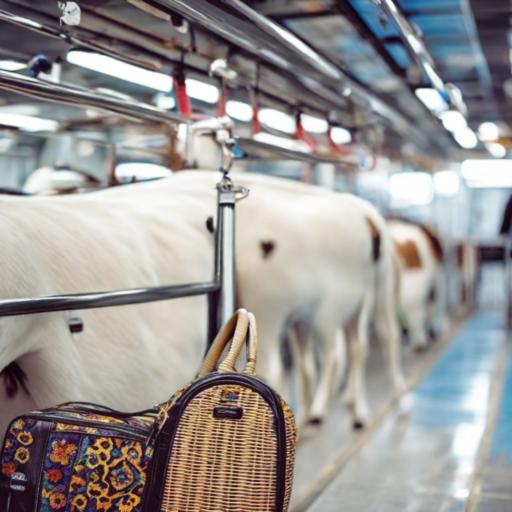}
    \end{minipage}}
    \smallskip

    \subfloat[ControlNet-Color~\cite{zhang2023adding}]{\label{fig:qualitative_baseline:g}\begin{minipage}[c]{\textwidth}
        \includegraphics[width=0.120\textwidth]{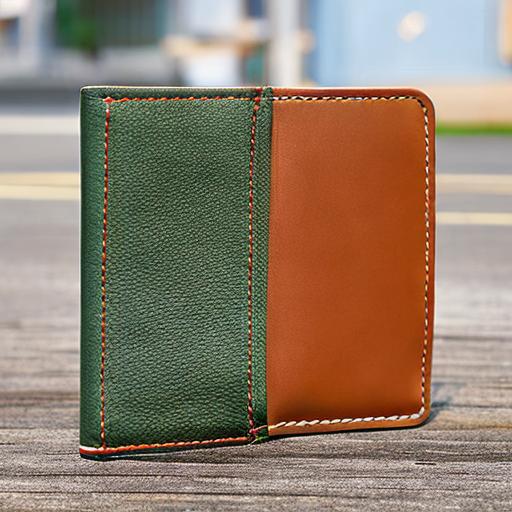}
        \hfill
        \includegraphics[width=0.120\textwidth]{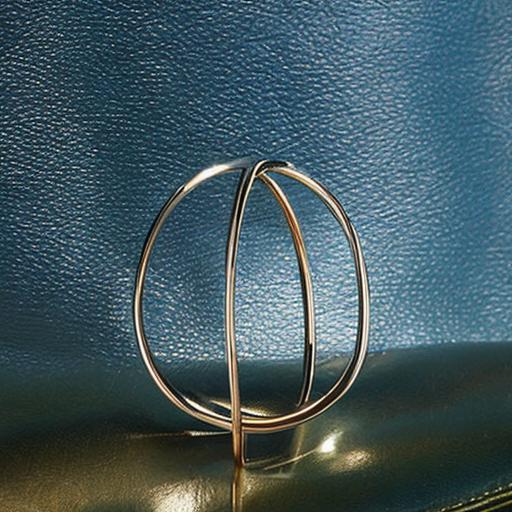}
        \hfill
        \includegraphics[width=0.120\textwidth]{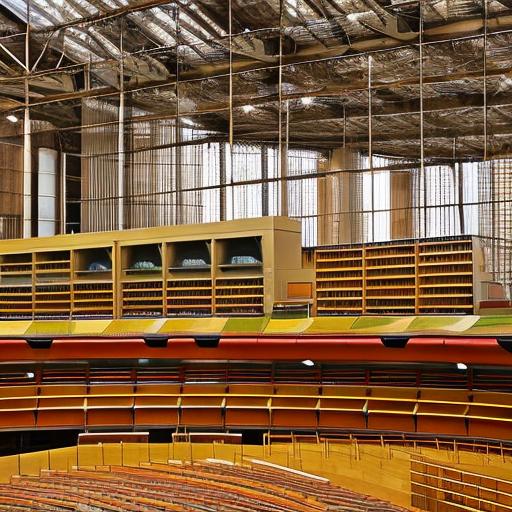}
        \hfill
        \includegraphics[width=0.120\textwidth]{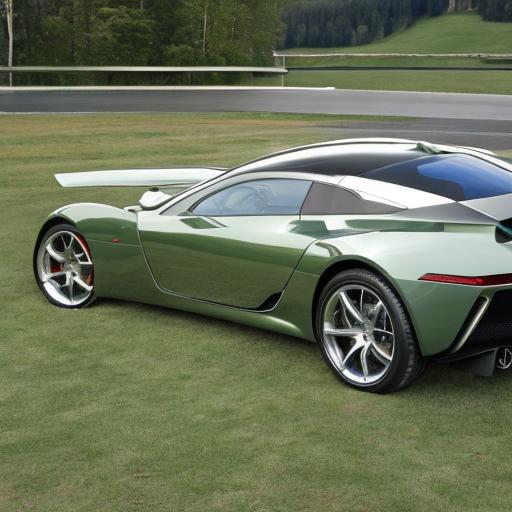}
        \hfill
        \includegraphics[width=0.120\textwidth]{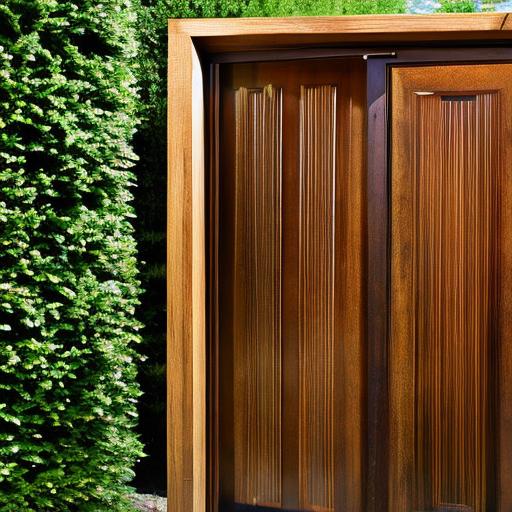}
        \hfill
        \includegraphics[width=0.120\textwidth]{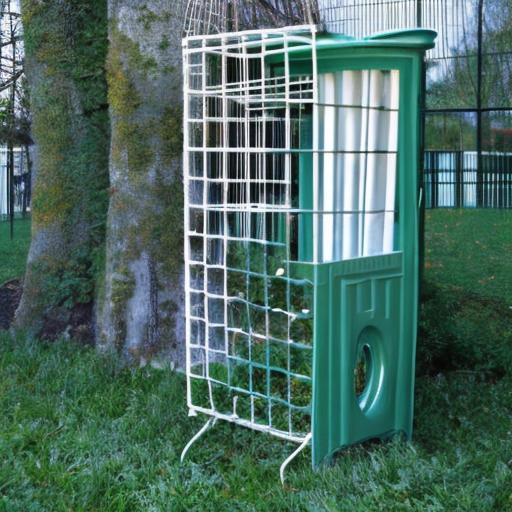}
        \hfill
        \includegraphics[width=0.120\textwidth]{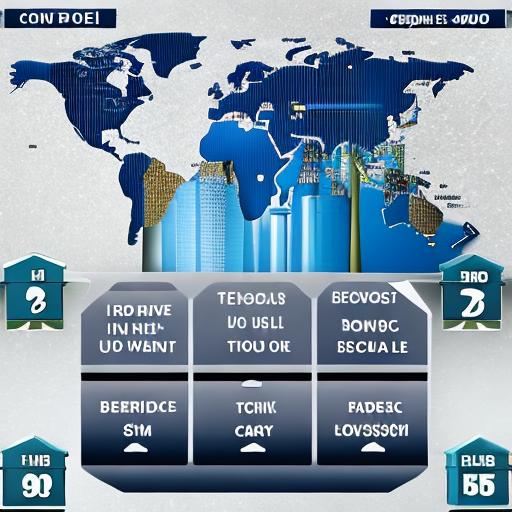}
        \hfill
        \includegraphics[width=0.120\textwidth]{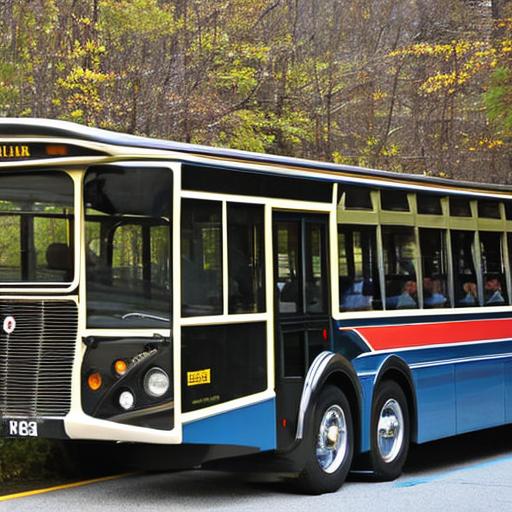}
    \end{minipage}}
    
    \caption{\textbf{Qualitative results of color-aligned latent diffusion}. Each input (first row) includes an in-the-wild image as color condition and a target text prompt. Each column presents results of experimented methods using the same input.} 
    \label{fig:qualitative_baseline}
\end{figure*}

\noindent\textbf{Color-aligned latent diffusion.}
We present our color-aligned latent diffusion results in comparison with existing baselines in~\cref{fig:qualitative_baseline}. As demonstrated, our method (\cref{fig:qualitative_baseline:b,fig:qualitative_baseline:c}) effectively conditions the synthesis on target color inputs. Color values and proportions are aligned, i.e., color \textit{accuracy} and \textit{completeness} are maintained. Additionally, our method generates contents that align well with target text prompts, showcasing the ability for color \textit{disentanglement} where colors are rearranged into new structures. Furthermore, with these color-conditioned generation criteria preserved, our method achieves image quality and diversity on par with the original diffusion model. 

In contrast, the baselines produce suboptimal results in various aspects.
IP-Adapter~\cite{ye2023ip} (\cref{fig:qualitative_baseline:d}), T2I-Adapter~\cite{mou2024t2i} (\cref{fig:qualitative_baseline:e}), and Style-Aligned~\cite{hertz2024style} (\cref{fig:qualitative_baseline:f}) construct target contents on top of an input image reference, thus often struggle to disentangle colors. Since the spatial information of colors is taken into account, these methods likely to fail completely if input reference and target text prompt belong to significantly different domains. ControlNet-Color~\cite{zhang2023adding} (\cref{fig:qualitative_baseline:g}) treats input reference as a weak guidance, leading to failures to accurately preserve color values and proportions.

\noindent\textbf{Generalizability and controllability.}
Our method also works with manual drawings, for example, a rough color pattern. As shown in~\cref{fig:qualitative_editing}, our method (the fine-tuned version) can generate images aligned with the color values and proportions specified in input color patterns. Moreover, even if only a few color values presented, our method can adjust these values to achieve appropriate smoothness, brightness, and shadow in generated contents. 

These manual color conditions can be easily modified to influence the generation process. For example, the amount of each color value can be scaled (\cref{fig:qualitative_editing:a}). The color values can be tuned intuitively without the need for careful prompt engineering (\cref{fig:qualitative_editing:b}). Additional colors can be incorporated into conditions along with user ideation (\cref{fig:qualitative_editing:c}). By controlling target text prompt, we can specify the placement of colors (\cref{fig:qualitative_editing:d}). Note that, like the regular diffusion, our method can generate diverse results from the same input condition by sampling different noises. 

\begin{figure}[h]
  \includegraphics[width=\columnwidth,trim={4mm 120mm 184mm 3mm},clip]{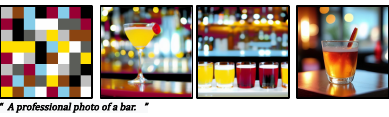}
  \\[-1.3mm]
  \includegraphics[width=\columnwidth,trim={4mm 120mm 184mm 3mm},clip]{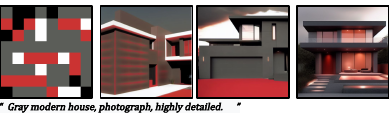}
  \\[-1.3mm]
  \includegraphics[width=\columnwidth,trim={4mm 120mm 184mm 3mm},clip]{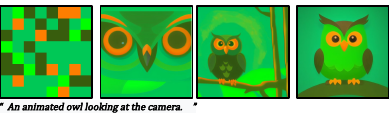}
  \\[-1.3mm]
  \includegraphics[width=\columnwidth,trim={4mm 120mm 184mm 3mm},clip]{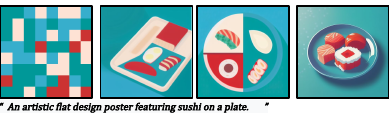}
  \\[-1.3mm]
  \includegraphics[width=\columnwidth,trim={4mm 120mm 184mm 3mm},clip]{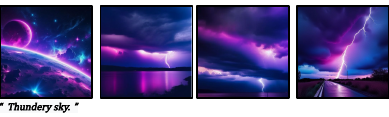}
  \\[-5.6mm]
  \subfloat[Condition]{\label{fig:zeroshot:a}~~~~~~~~~~~~~~~~~~~~~~}
  \subfloat[Zero-shot]{\label{fig:zeroshot:b}~~~~~~~~~~~~~~~~~~~~~~~~~~~~~~~~~~~~~~~~~~~~~~~~~~}
  \subfloat[Fine-tune]{\label{fig:zeroshot:c}~~~~~~~~~~~~~~~~~~~~~~~}
  
    \caption{\textbf{Qualitative results of our zero-shot approach}. The first four rows show results from manual color conditions, while the fifth row presents results from in-the-wild imagery condition.}
\label{fig:zeroshot}
\end{figure}

\noindent\textbf{Zero-shot diffusion.}
We showcase the ability of our method in zero-shot diffusion in~\cref{fig:zeroshot}. Our experiments indicate that, compared with our fine-tuned version, the zero-shot version tends to generate images with less fine-grained details, resulting in issues such as blurry lighting, uniform texture, and shadow missing. Since our zero-shot approach mechanically maps colors without model training, it performs more comparably to the fine-tuned version when applied to well-sampled color conditions, e.g., in-the-wild images in the fifth row in~\cref{fig:zeroshot} and those in~\cref{fig:qualitative_baseline}.

\subsection{Quantitative study}
\label{sec:quantitative}
\noindent\textbf{Metrics.}
We evaluate the quality of generated images using the widely adopted FID score~\cite{heusel2017gans}. To assess the color \textit{disentanglement}, we use the CLIP-Score~\cite{hessel2021clipscore} measuring the similarity between a generated image and a target text prompt within the CLIP~\cite{radford2021learning} embedding space. Since the color \textit{accuracy} and \textit{completeness} can be evaluated independently of spatial relations, we measure the proximity between the color  distributions of a generated image and input color condition. We employ the Chamfer Distance (CD)~\cite{athitsos2003estimating}, a commonly used metric for measuring the distance between two point sets, to quantify the distance between generated RGB points $\hat{\mathbf{x}}_0$ and input condition RGB points $\mathbf{c}$:

\begin{figure*}[t]
    \subfloat[Scaling color proportion.]{\label{fig:qualitative_editing:a}\begin{minipage}[c]{\textwidth}
    \centering
        \includegraphics[width=\textwidth,trim={1mm 119mm 94mm 0mm},clip]{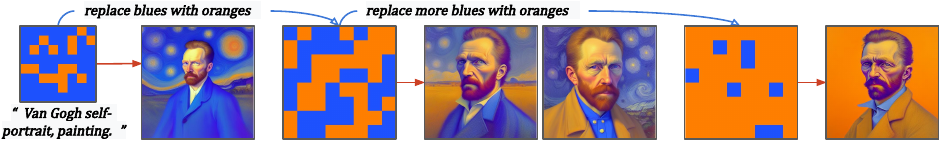}
    \end{minipage}}
    \smallskip
    \smallskip

    \subfloat[Adjusting color values.]{\label{fig:qualitative_editing:b}\begin{minipage}[c]{\textwidth}
    \centering
        \includegraphics[width=\textwidth,trim={1mm 119mm 94mm 0mm},clip]{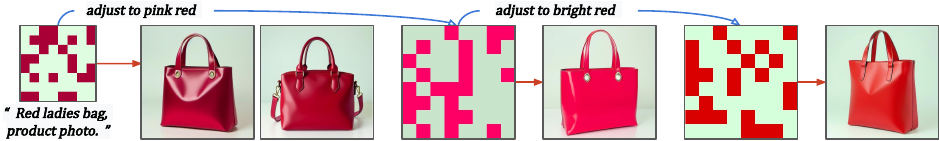}
    \end{minipage}}
    \smallskip
    \smallskip

    \subfloat[Editing color palette.]{\label{fig:qualitative_editing:c}\begin{minipage}[c]{\textwidth}
    \centering
        \includegraphics[width=\textwidth,trim={1mm 119mm 94mm 0mm},clip]{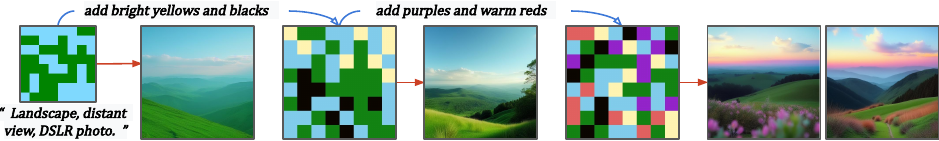}
    \end{minipage}}
    \smallskip
    \smallskip

    \subfloat[Controlling color prompts.]{\label{fig:qualitative_editing:d}\begin{minipage}[c]{\textwidth}
    \centering
        \includegraphics[width=\textwidth,trim={1mm 119mm 94mm 0mm},clip]{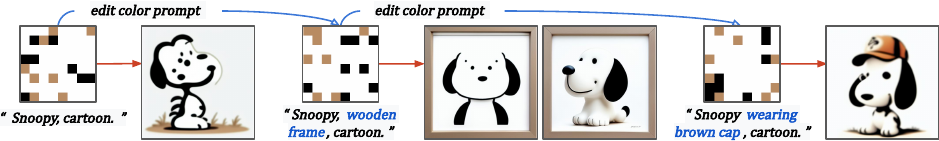}
    \end{minipage}}
    
    \caption{\textbf{Qualitative results of sampling and editing of input color conditions}. All results are generated by our fine-tuned color-aligned latent model. Red arrows represent the generation process. Blue arrows represent the editing of color conditions.} 
    \label{fig:qualitative_editing}
\end{figure*}

\begin{table*}
    \small
    \centering
    \begin{tabular}{l|c|cc|cc}
        \toprule
        Method & FID $\downarrow$ & CD-A $\downarrow$ & CD-C $\downarrow$ & Re-training Cost $\downarrow$ & Inference Cost $\downarrow$ \\
        \midrule
        Image Diffusion (Re-train)~\cite{ho2020denoising} & 63.9; 149 & 40.6; 42.4 & 42.6; 39.0 & \underline{1.00} & \textbf{1.00} \\ 
        \midrule
        Ours (Image; Re-train) & \underline{57.5}; \textbf{45.7} & \textbf{0.00}; \textbf{0.00} & \textbf{4.12}; \textbf{3.65} & 1.06 & \underline{1.13} \\ 
        Ours (Image; Zero-shot) & \textbf{50.0}; \underline{84.5} & \underline{5.59}; \underline{5.46} & \underline{8.28}; \underline{10.4} & \textbf{0.00} & 6.76 \\
        \bottomrule
    \end{tabular}

    \vspace{-0.1cm}
    
    \caption{\textbf{Quantitative results of image diffusion}. In each metric column, the left and right numbers correspond to the scores from the Oxford-flower dataset~\cite{nilsback2008automated} and Microsoft-emoji~\cite{microsoft2021emoji} dataset, respectively.}
    \label{tab:quantitative_image_space}
\end{table*}


\begin{table*}
    \small
    \centering
    \begin{tabular}{l|cc|cc|cc}
        \toprule
        Method & FID $\downarrow$ & CLIPScore $\uparrow$ & CD-A $\downarrow$ & CD-C $\downarrow$ & Inference Cost $\downarrow$ \\
        \midrule
        Stable Diffusion (Pre-train)~\cite{rombach2022high} & \textbf{72.6}; 72.6 & 27.3; 27.3 & 166; 27.8 & 65.3; 16.0 & \textbf{1.00} \\ 
        \midrule
        Ours (Latent; Fine-tune) & \underline{86.7}; \underline{69.4} & \textbf{29.0}; \textbf{27.4} & 73.9; \underline{4.98} & \underline{15.8}; \underline{4.87} & \underline{1.03} \\
        Ours (Latent; Zero-shot) & 104; 77.9 & \underline{28.3}; \underline{27.4} & \textbf{35.2}; \textbf{2.68} & \textbf{3.19}; \textbf{3.93} & 3.52 \\
        \midrule
        Baseline (IP-Adapter~\cite{ye2023ip}) & 202; \textbf{50.9} & 25.5; 22.0 & 145; 10.1 & 106; 13.9 & 1.44 \\
        Baseline (T2I-Adapter~\cite{mou2024t2i}) & 134; 77.1 & 24.3; 24.9 & 143; 8.34 & 24.7; 7.01 & 1.56 \\
        Baseline (Style-Aligned~\cite{hertz2024style}) & 177; 73.1 & 23.5; 22.9 & \underline{45.1}; 6.18 & 32.5; 8.50 & 5.69 \\
        Baseline (ControlNet-Color~\cite{zhang2023adding}) & 105; 81.0 & 22.9; 23.4 & 129; 13.1 & 49.8; 6.54 & 1.78 \\
        \bottomrule
    \end{tabular}
 
    \caption{\textbf{Quantitative results of latent diffusion}. In each metric column, the left and right numbers correspond to the scores from the settings of manually sampled color condition and in-the-wild image color condition, respectively.}
    \label{tab:quantitative_latent_space}
\end{table*}

\newcommand\cdacc{\operatorname{CD-A}}
\newcommand\cdcomp{\operatorname{CD-C}}
\begin{align}
    \cdacc(\hat{\mathbf{x}}_0, \mathbf{c}) &= \frac{\sum_{\hat{\mathbf{x}}_0[p]}\min_{\mathbf{c}[q]}\| \hat{\mathbf{x}}_{0}[p] - \mathbf{c}[q] \|^2_2}{|\hat{\mathbf{x}}_0|}
    \\
    \cdcomp(\hat{\mathbf{x}}_0, \mathbf{c}) &= \frac{\sum_{\mathbf{c}[q]}\min_{\hat{\mathbf{x}}_0[p]}\| \hat{\mathbf{x}}_{0}[p] - \mathbf{c}[q] \|^2_2}{|\mathbf{c}|}
    \label{eq:chamfer_distance}
\end{align}
where $\mathrm{CD} = \cdacc + \cdcomp$ consists of an \textit{accuracy} term and a \textit{completeness} term that measure the set distance from one to another; $|\cdot|$ denotes set cardinality. Lastly, we evaluate the inference cost of different methods, using the regular diffusion as the standard reference.

\noindent\textbf{Baseline comparison.}
We first present the performance scores for the image diffusion case study in~\cref{tab:quantitative_image_space}. The scores clearly indicate that our technique effectively preserves conditioned-colors. Note that, for inference, the color alignment is performed on RGB pixels until the final diffusion step, making the CD-A approach zero. We found that our fine-tuned version requires nearly the same training and inference costs as the regular diffusion model.

We report the performance scores for the latent diffusion case study in~\cref{tab:quantitative_latent_space}. As shown, our method generally outperforms the baselines in most of the metrics. The baselines exhibit significant failures in metrics reflecting the \textit{disentanglement} (CLIP-score), \textit{accuracy} and \textit{completeness} (CD-A and CD-C scores) of generated colors in synthesized images. When provided with manual color conditions, our zero-shot version achieves worse FID score than the fine-tuned version does. This aligns with our observations in~\cref{sec:qualitative} indicating imagery details lacking caused by the zero-shot version. The zero-shot setting's very low CD scores also indicate color overfitting caused by mechanically mapping the manual color conditions.

\noindent\textbf{Ablation study.}
We validate the effectiveness of important components in our method in~\cref{tab:ablation_main_module}. Specifically, we investigate the role of the color mappings $f$, and the inputting of the color condition $\mathbf{c}$ (those in~\cref{fig:pipeline:c}). When mappings are not used, noisy samples are not altered and only color conditions are fed to the diffusion model. As shown in~\cref{tab:ablation_main_module}, the mappings and color conditions significantly boost up the performance of our method.

\begin{table}
    \small
    \centering
    \begin{tabular}{cc|c|cc}
        \toprule
        Mapping & Condition & FID $\downarrow$ & CD-A $\downarrow$ & CD-C $\downarrow$ \\
        \midrule
        \ding{55} & \ding{55} & 63.9; 149 & 40.6; 42.4 & 42.6; 39.0 \\
        \ding{55} & \ding{51} & 57.9; 88.1 & 15.4; 10.6 & 18.4; 15.2 \\
        \ding{51} & \ding{55} & 59.8; 45.7 & 0.00; 0.00 & 5.10; 4.62 \\ 
        \ding{51} & \ding{51} & 57.5; 45.6 & 0.00; 0.00 & 4.12; 3.65 \\
        \bottomrule
    \end{tabular}
    \caption{\textbf{Ablation study} on color mapping and color condition. For each metric, the left and right numbers are from the Oxford-flower~\cite{nilsback2008automated} and Microsoft-emoji~\cite{microsoft2021emoji} datasets, respectively.}
    \label{tab:ablation_main_module}
\end{table}

\vspace{-0.1cm}

\section{Conclusion and Discussion}
We propose a color alignment method that operates on the diffusion process of diffusion models. Diffused colors are mapped to conditional colors across diffusion steps, enabling a continuous pathway for synthesizing target color patterns while maintaining the creativity of diffusion models.

There are issues that would be addressed in our future work. First, while our method can condition on color values and proportions, it is worthwhile to understand the influence of the spatial information in the color conditions. Second, it is of great interest to extend our color alignment to video and 3D data, potentially with integration of multi-view attributes.


{\small
\bibliographystyle{ieee_fullname}
\bibliography{egbib}
}

\clearpage
\part*{Appendix}
\section{Additional comparison results}
\label{sec:additional_results}

\begin{figure*}[t]
    \subfloat[Input Condition]{\label{fig:supp_qualitative_baseline:a}\begin{minipage}[c]{\textwidth}
        \includegraphics[width=0.120\textwidth]{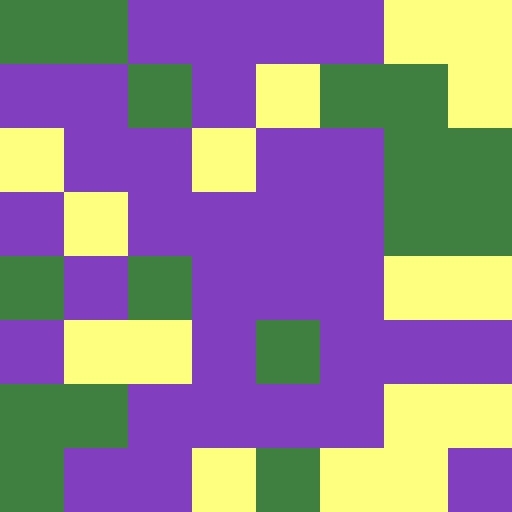}
        \hfill
        \includegraphics[width=0.120\textwidth]{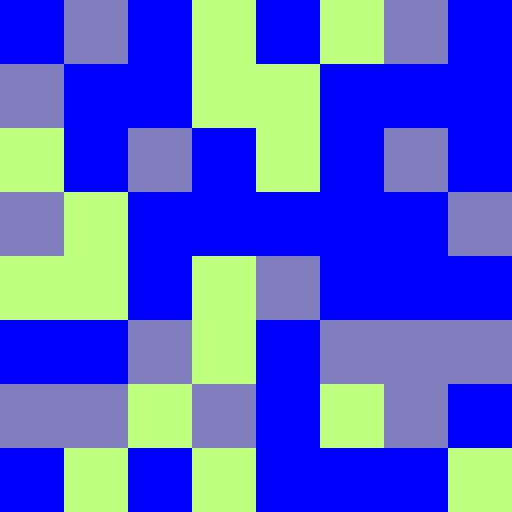}
        \hfill
        \includegraphics[width=0.120\textwidth]{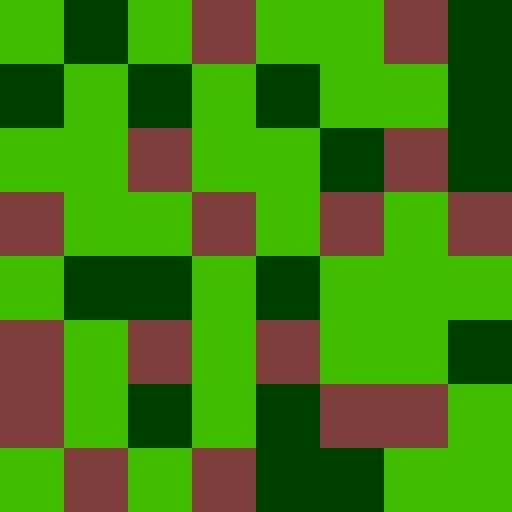}
        \hfill
        \includegraphics[width=0.120\textwidth]{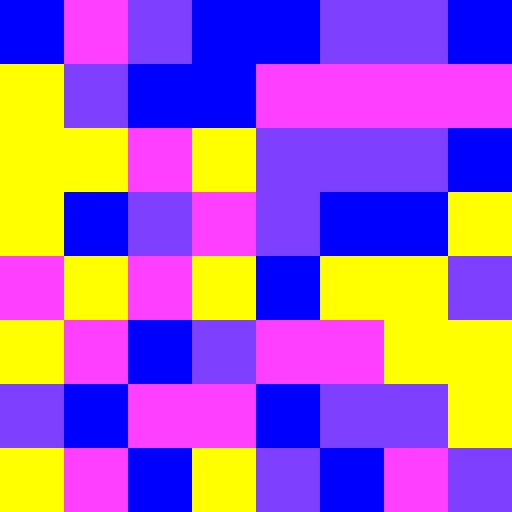}
        \hfill
        \includegraphics[width=0.120\textwidth]{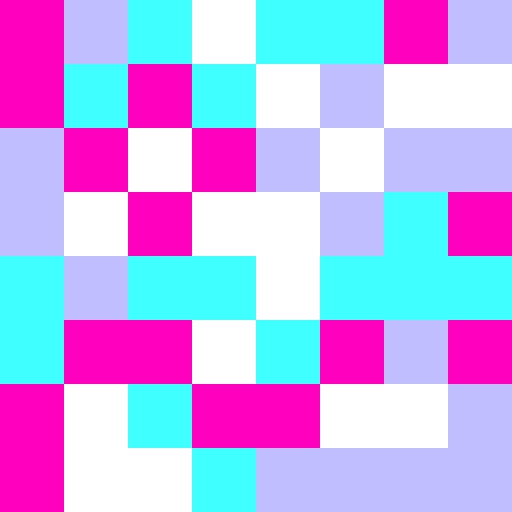}
        \hfill
        \includegraphics[width=0.120\textwidth]{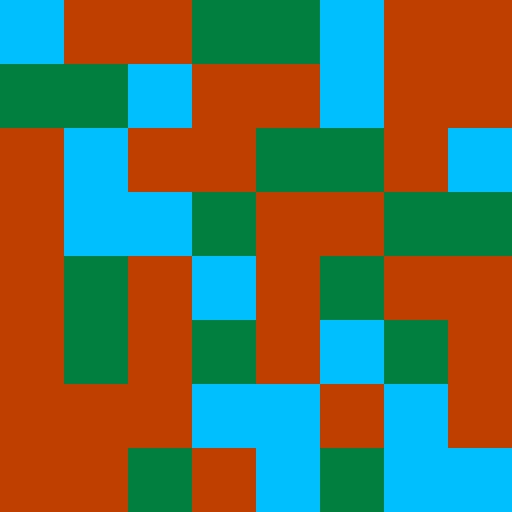}
        \hfill
        \includegraphics[width=0.120\textwidth]{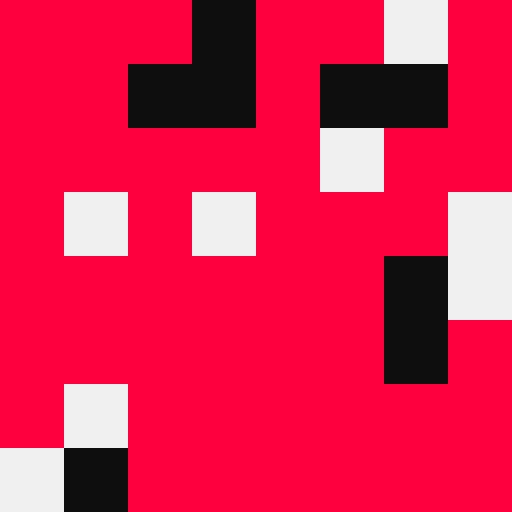}
        \hfill
        \includegraphics[width=0.120\textwidth]{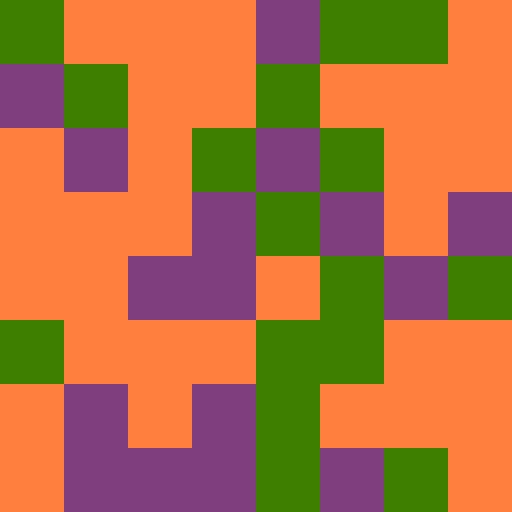}
        \smallskip
        \hspace{5.4mm}``chair''\hspace{5.0mm}
        \hspace{6.9mm}``dog''\hspace{5.0mm}
        \hspace{8.1mm}``cup''\hspace{5.0mm}
        \hspace{5.3mm}``airplane''\hspace{5.0mm}
        \hspace{4.0mm}``house''\hspace{5.0mm}
        \hspace{4.1mm}``statue''\hspace{5.0mm}
        \hspace{5.5mm}``bottle''\hspace{5.0mm}
        \hspace{4.5mm}``ladder''\hspace{5.0mm}
    \end{minipage}}
    \smallskip

    \subfloat[Ours (Fine-tune)]{\label{fig:supp_qualitative_baseline:b}\begin{minipage}[c]{\textwidth}
        \includegraphics[width=0.120\textwidth]{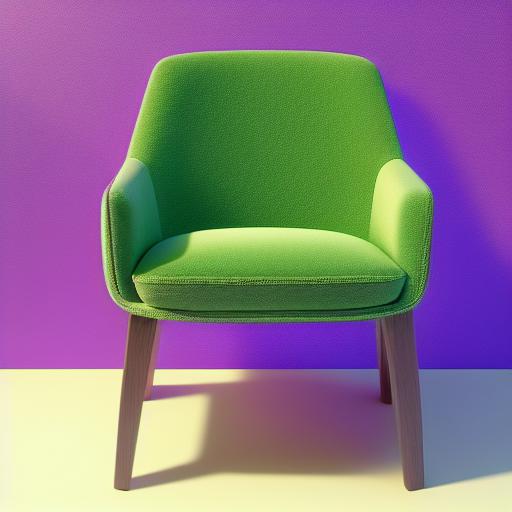}
        \hfill
        \includegraphics[width=0.120\textwidth]{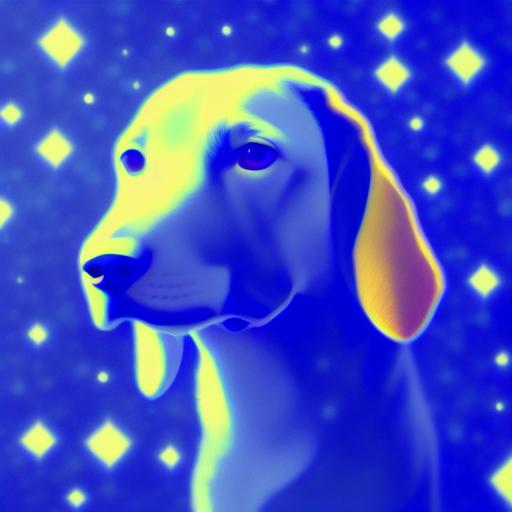}
        \hfill
        \includegraphics[width=0.120\textwidth]{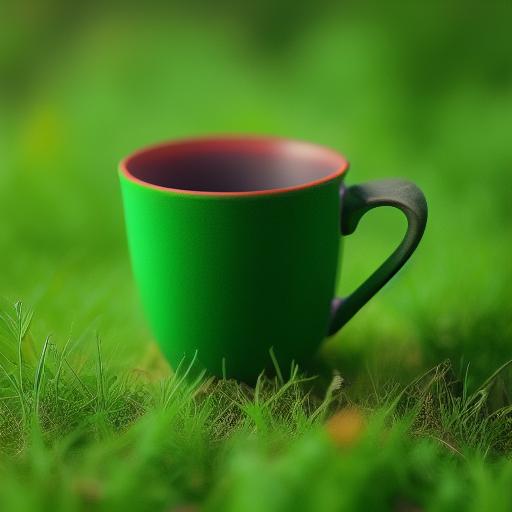}
        \hfill
        \includegraphics[width=0.120\textwidth]{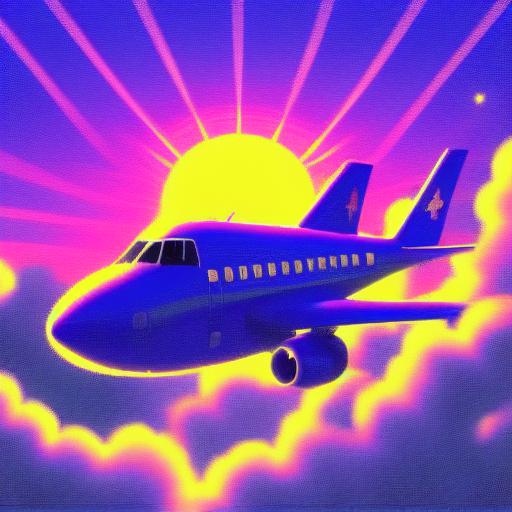}
        \hfill
        \includegraphics[width=0.120\textwidth]{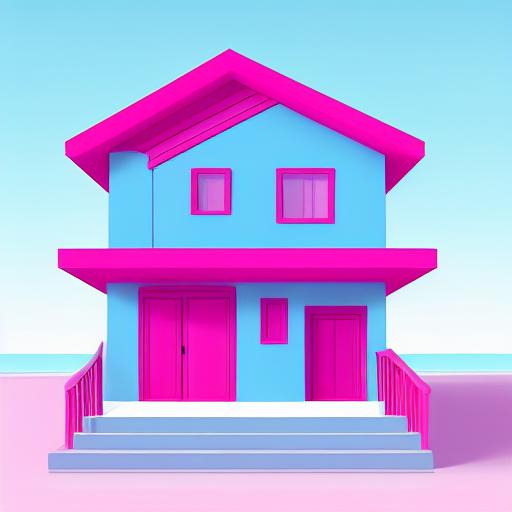}
        \hfill
        \includegraphics[width=0.120\textwidth]{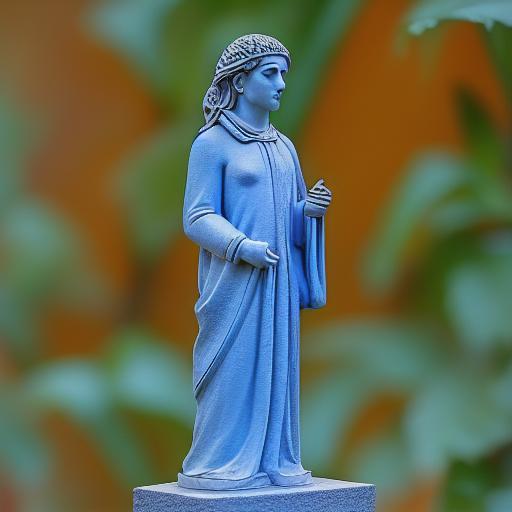}
        \hfill
        \includegraphics[width=0.120\textwidth]{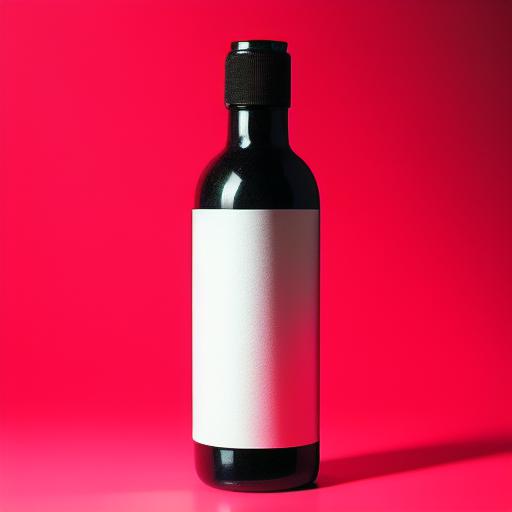}
        \hfill
        \includegraphics[width=0.120\textwidth]{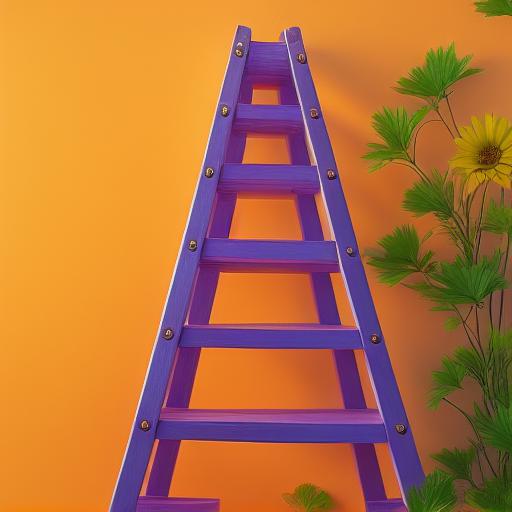}
    \end{minipage}}
    \smallskip

    \subfloat[Ours (Zero-shot)]{\label{fig:supp_qualitative_baseline:c}\begin{minipage}[c]{\textwidth}
        \includegraphics[width=0.120\textwidth]{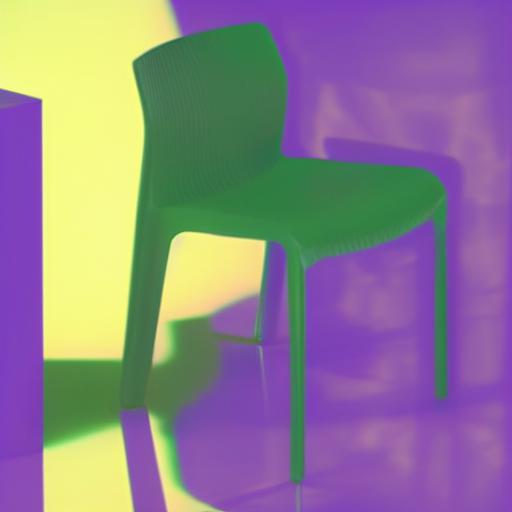}
        \hfill
        \includegraphics[width=0.120\textwidth]{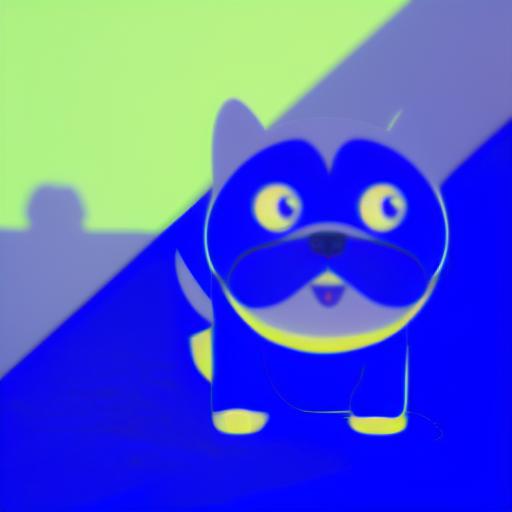}
        \hfill
        \includegraphics[width=0.120\textwidth]{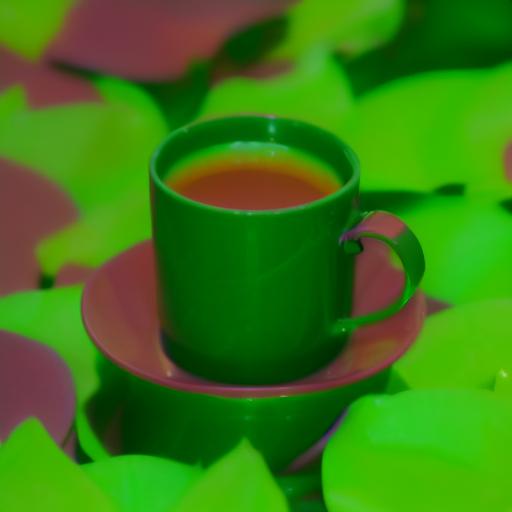}
        \hfill
        \includegraphics[width=0.120\textwidth]{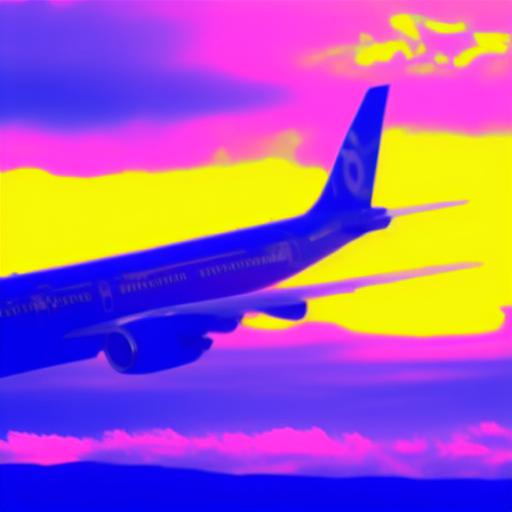}
        \hfill
        \includegraphics[width=0.120\textwidth]{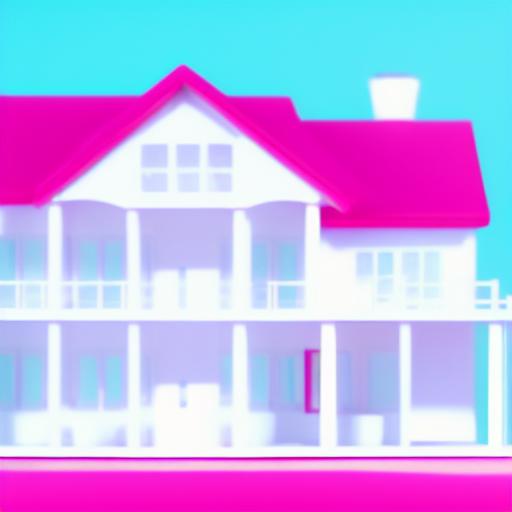}
        \hfill
        \includegraphics[width=0.120\textwidth]{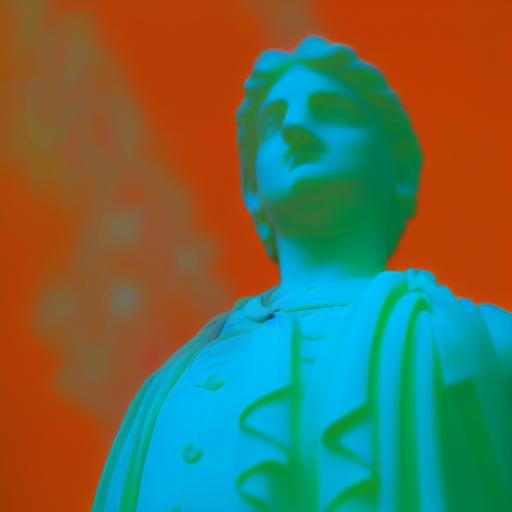}
        \hfill
        \includegraphics[width=0.120\textwidth]{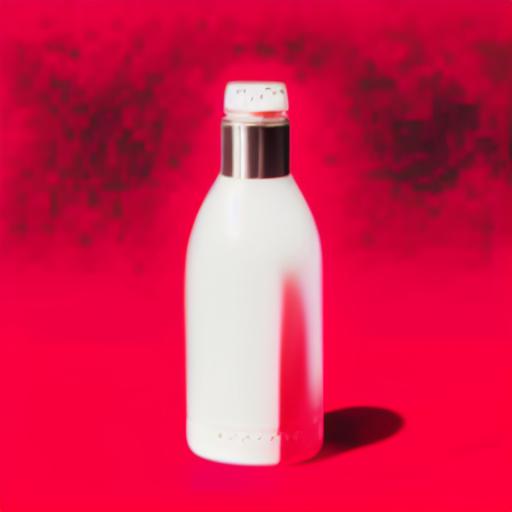}
        \hfill
        \includegraphics[width=0.120\textwidth]{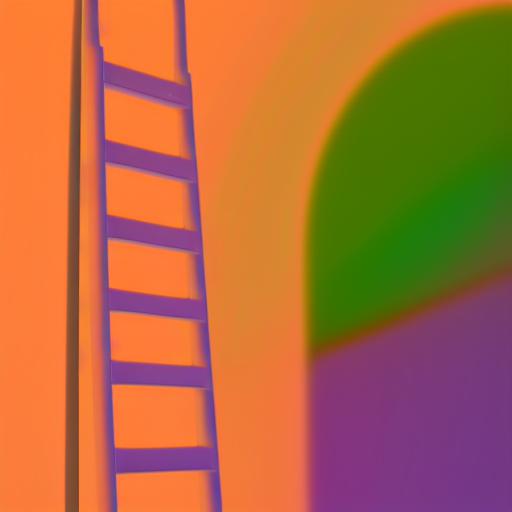}
    \end{minipage}}
    \smallskip

    \subfloat[IP-Adapter~\cite{ye2023ip}]{\label{fig:supp_qualitative_baseline:d}\begin{minipage}[c]{\textwidth}
        \includegraphics[width=0.120\textwidth]{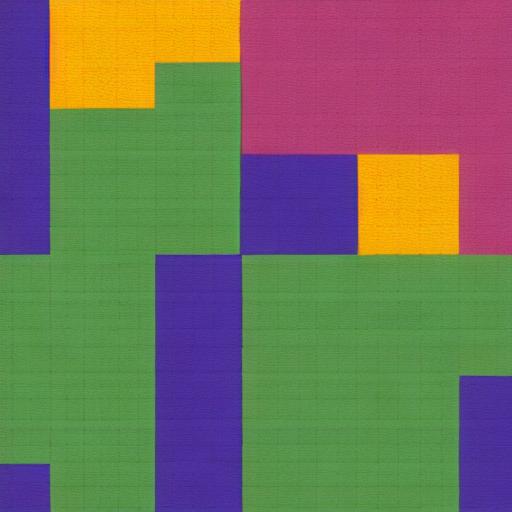}
        \hfill
        \includegraphics[width=0.120\textwidth]{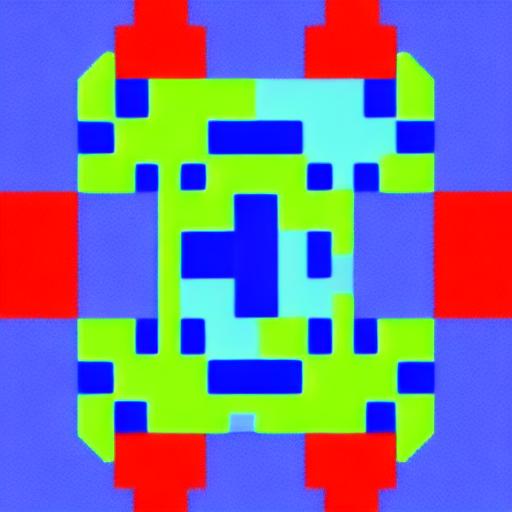}
        \hfill
        \includegraphics[width=0.120\textwidth]{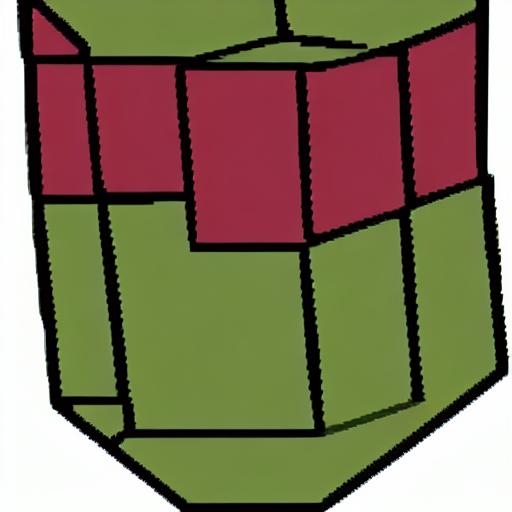}
        \hfill
        \includegraphics[width=0.120\textwidth]{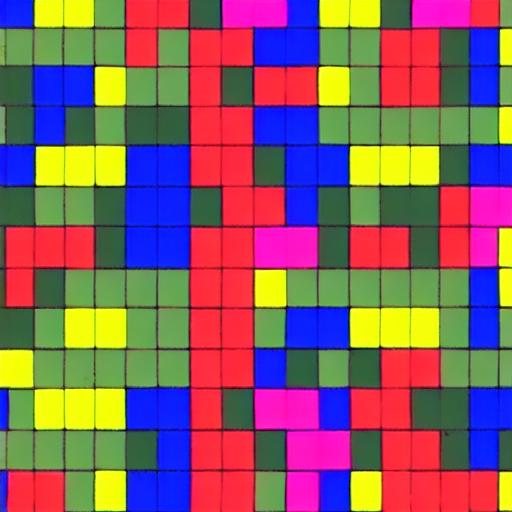}
        \hfill
        \includegraphics[width=0.120\textwidth]{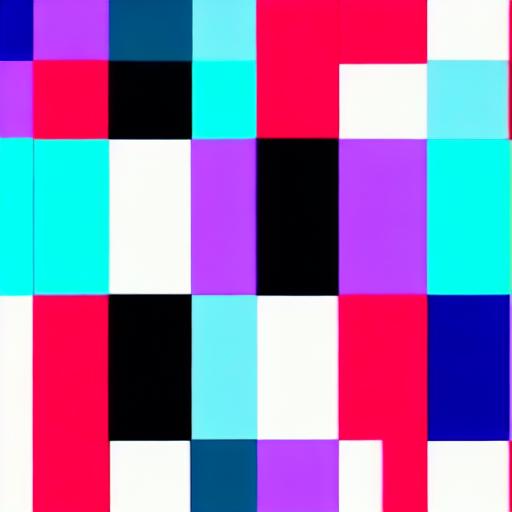}
        \hfill
        \includegraphics[width=0.120\textwidth]{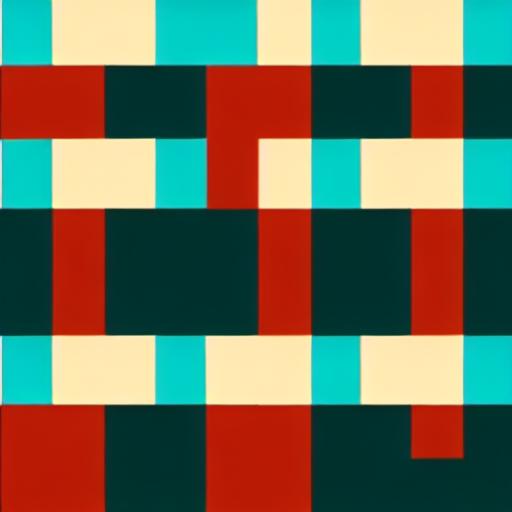}
        \hfill
        \includegraphics[width=0.120\textwidth]{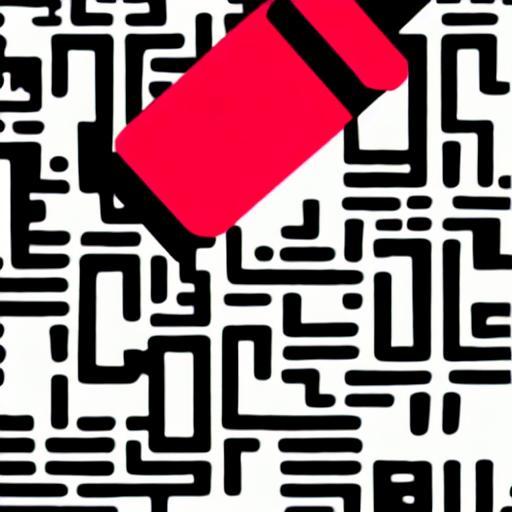}
        \hfill
        \includegraphics[width=0.120\textwidth]{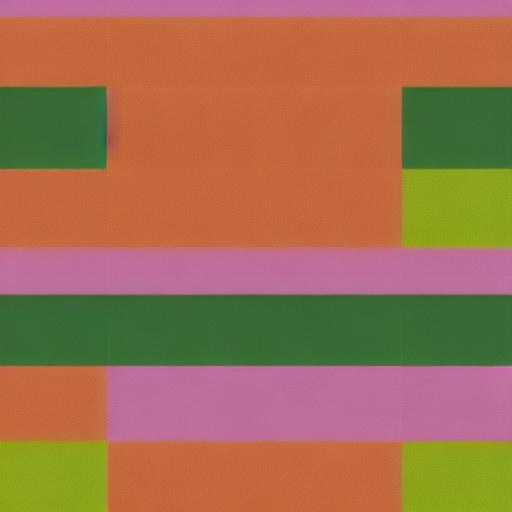}
    \end{minipage}}
    \smallskip

    \subfloat[T2I-Adapter~\cite{mou2024t2i}]{\label{fig:supp_qualitative_baseline:e}\begin{minipage}[c]{\textwidth}
        \includegraphics[width=0.120\textwidth]{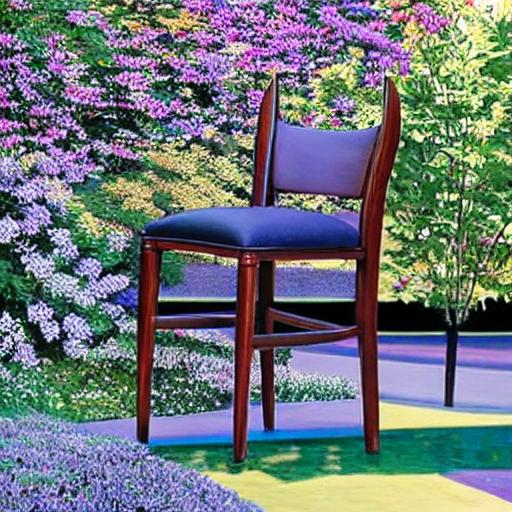}
        \hfill
        \includegraphics[width=0.120\textwidth]{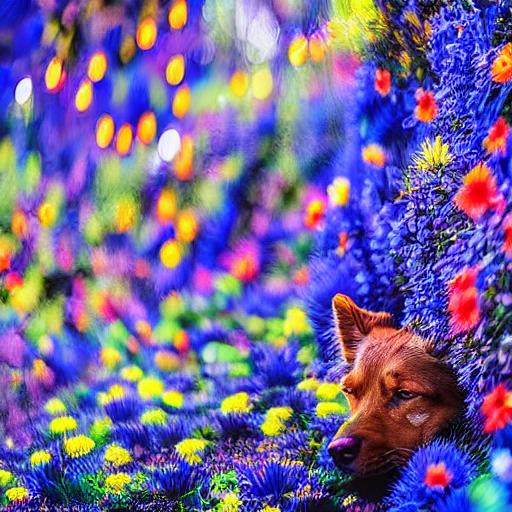}
        \hfill
        \includegraphics[width=0.120\textwidth]{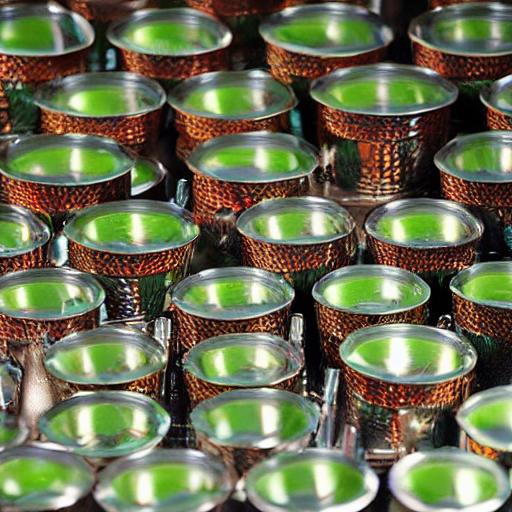}
        \hfill
        \includegraphics[width=0.120\textwidth]{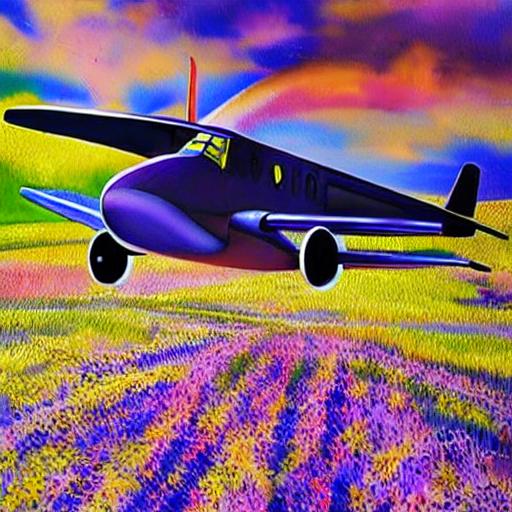}
        \hfill
        \includegraphics[width=0.120\textwidth]{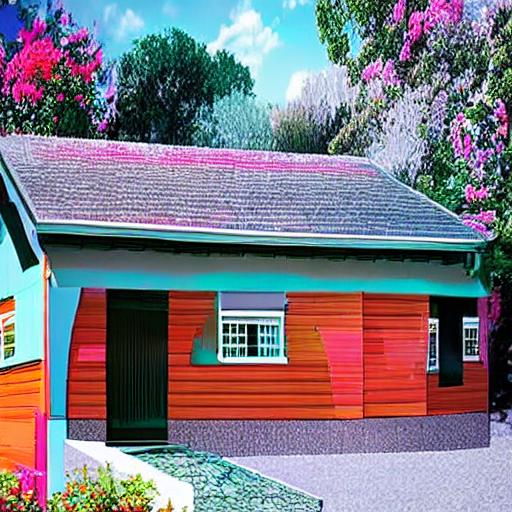}
        \hfill
        \includegraphics[width=0.120\textwidth]{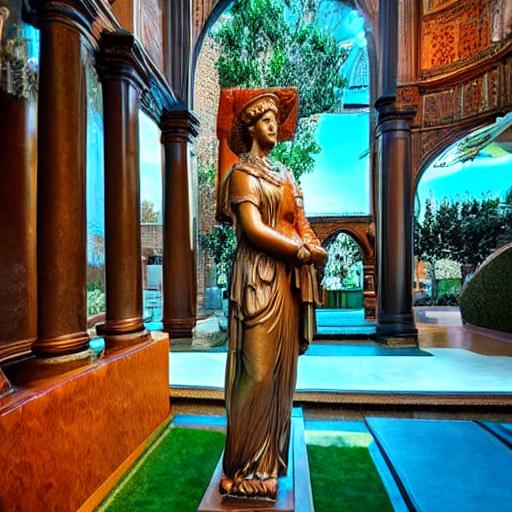}
        \hfill
        \includegraphics[width=0.120\textwidth]{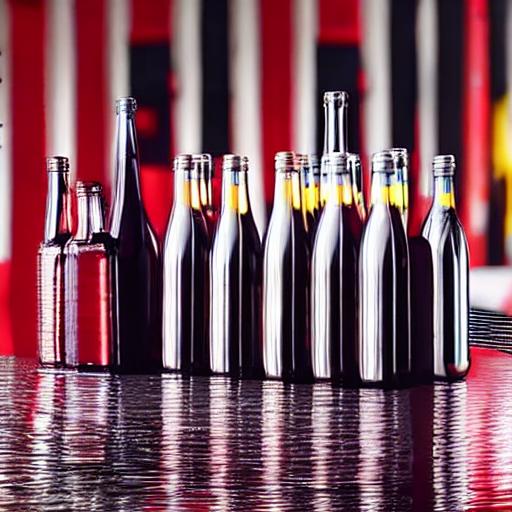}
        \hfill
        \includegraphics[width=0.120\textwidth]{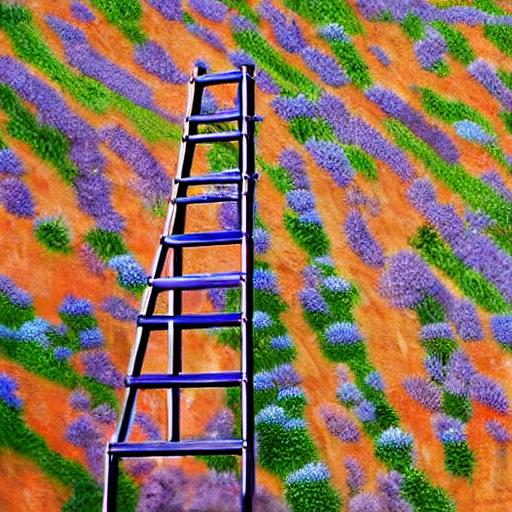}
    \end{minipage}}
    \smallskip

    \subfloat[Style-Aligned~\cite{hertz2024style}]{\label{fig:supp_qualitative_baseline:f}\begin{minipage}[c]{\textwidth}
        \includegraphics[width=0.120\textwidth]{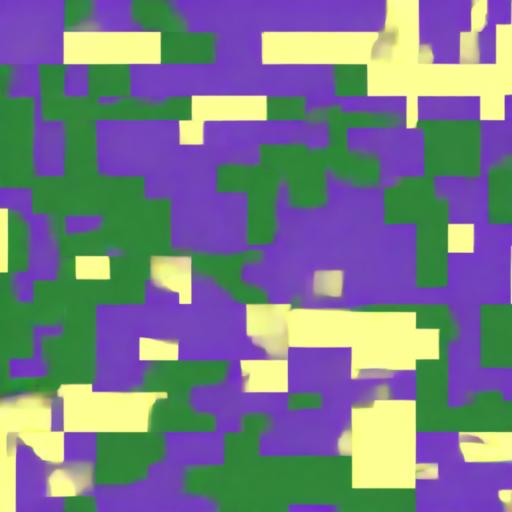}
        \hfill
        \includegraphics[width=0.120\textwidth]{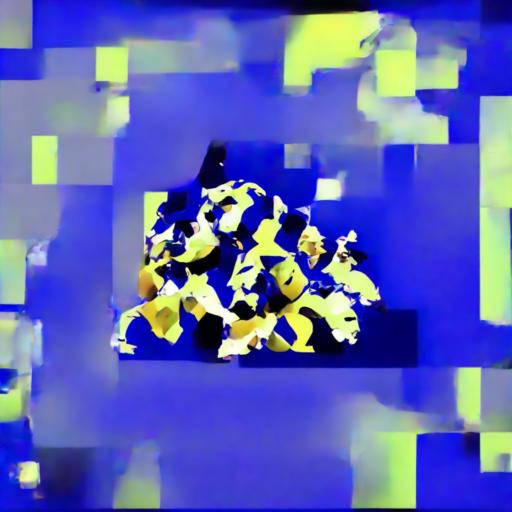}
        \hfill
        \includegraphics[width=0.120\textwidth]{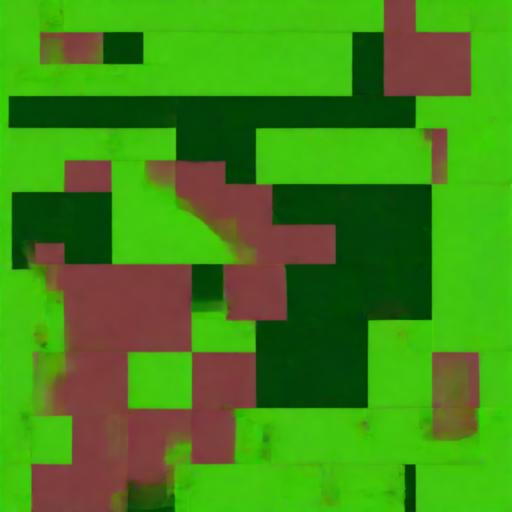}
        \hfill
        \includegraphics[width=0.120\textwidth]{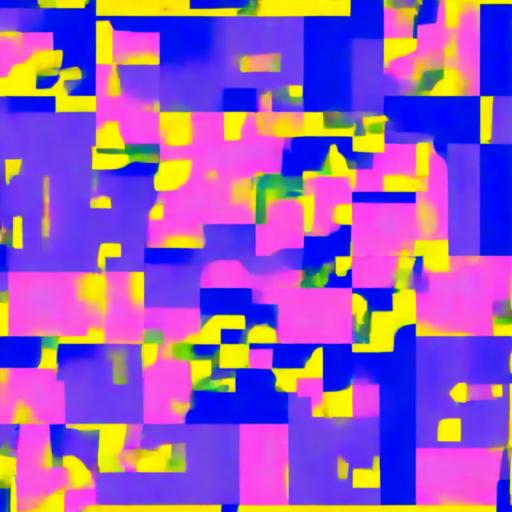}
        \hfill
        \includegraphics[width=0.120\textwidth]{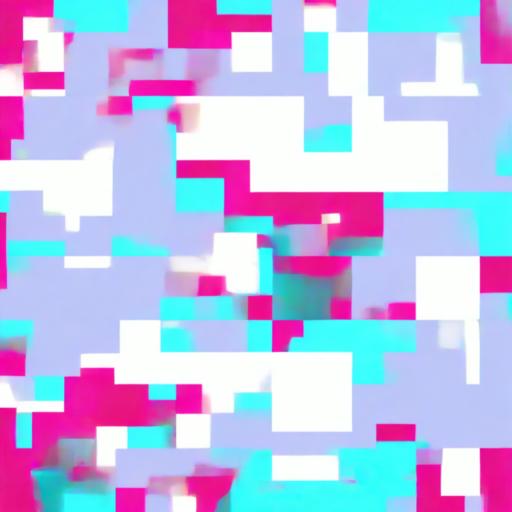}
        \hfill
        \includegraphics[width=0.120\textwidth]{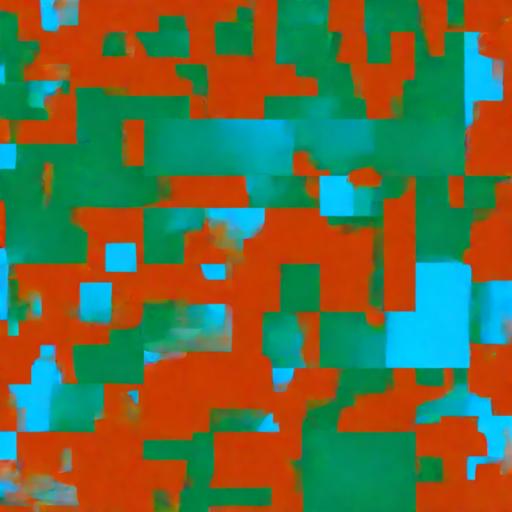}
        \hfill
        \includegraphics[width=0.120\textwidth]{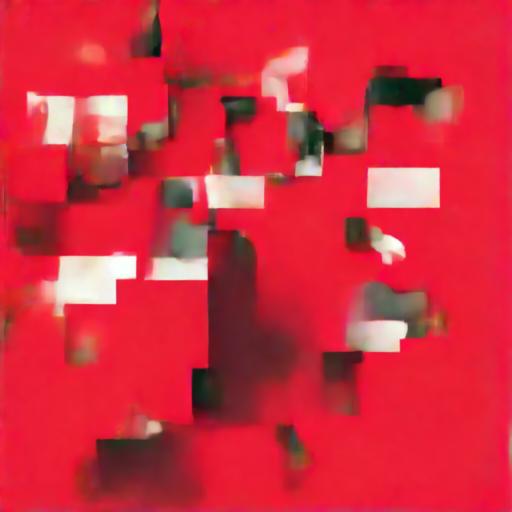}
        \hfill
        \includegraphics[width=0.120\textwidth]{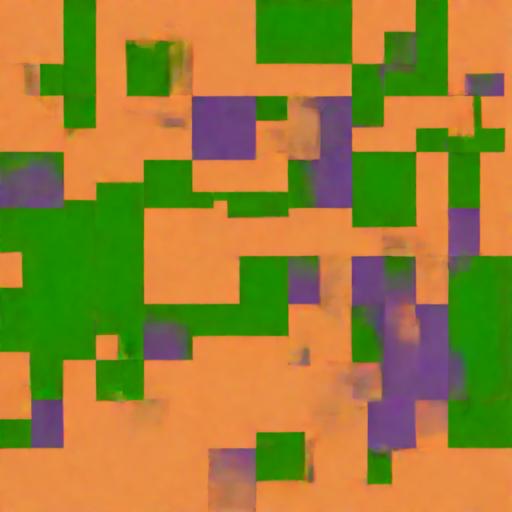}
    \end{minipage}}
    \smallskip

    \subfloat[ControlNet-Color~\cite{zhang2023adding}]{\label{fig:supp_qualitative_baseline:g}\begin{minipage}[c]{\textwidth}
        \includegraphics[width=0.120\textwidth]{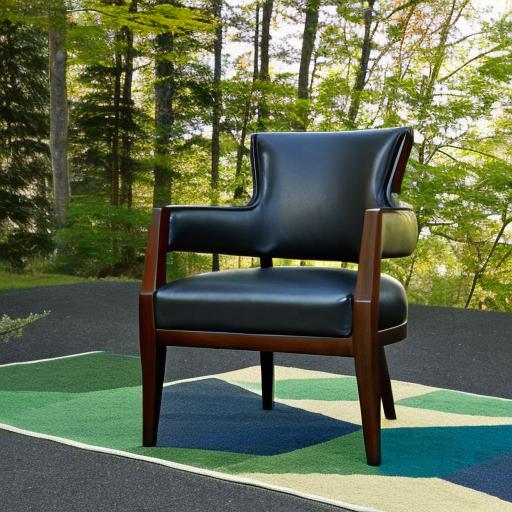}
        \hfill
        \includegraphics[width=0.120\textwidth]{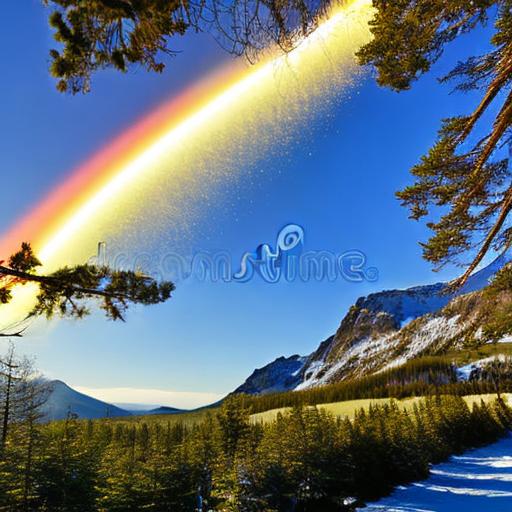}
        \hfill
        \includegraphics[width=0.120\textwidth]{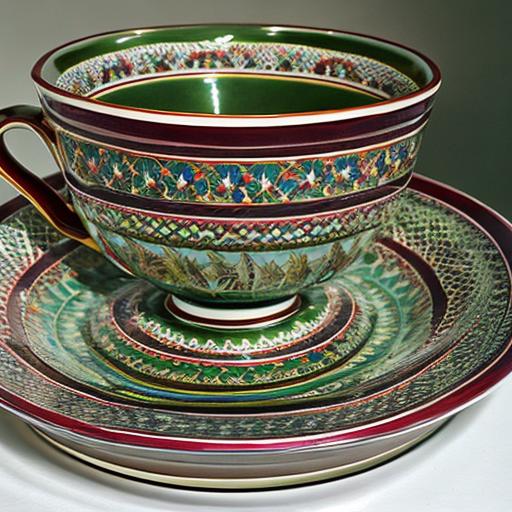}
        \hfill
        \includegraphics[width=0.120\textwidth]{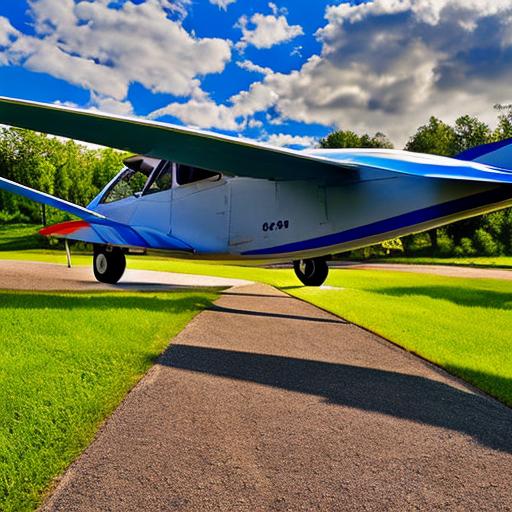}
        \hfill
        \includegraphics[width=0.120\textwidth]{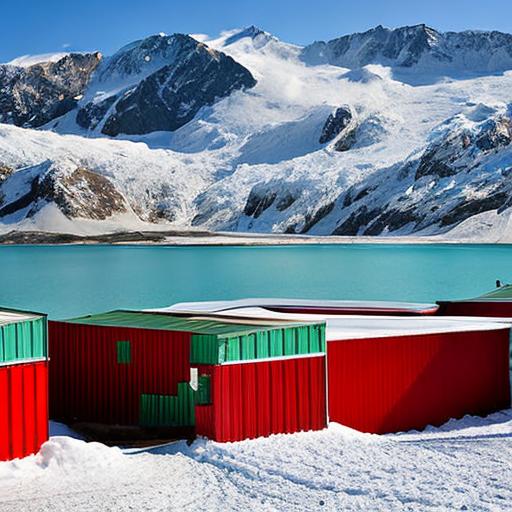}
        \hfill
        \includegraphics[width=0.120\textwidth]{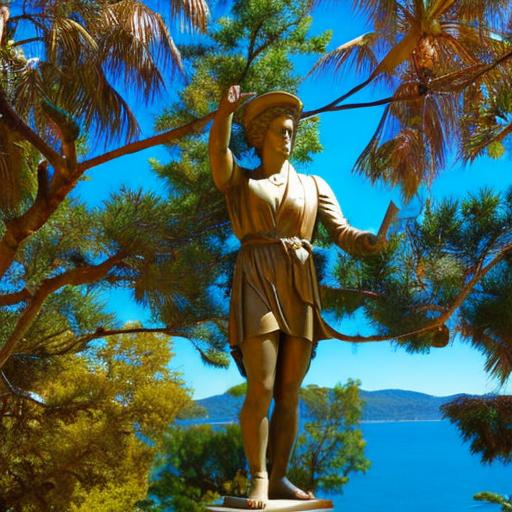}
        \hfill
        \includegraphics[width=0.120\textwidth]{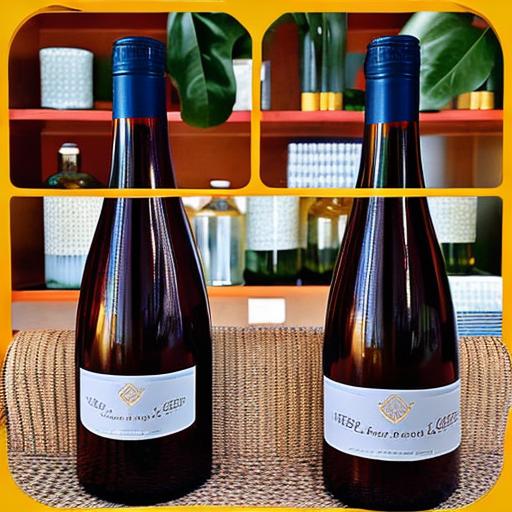}
        \hfill
        \includegraphics[width=0.120\textwidth]{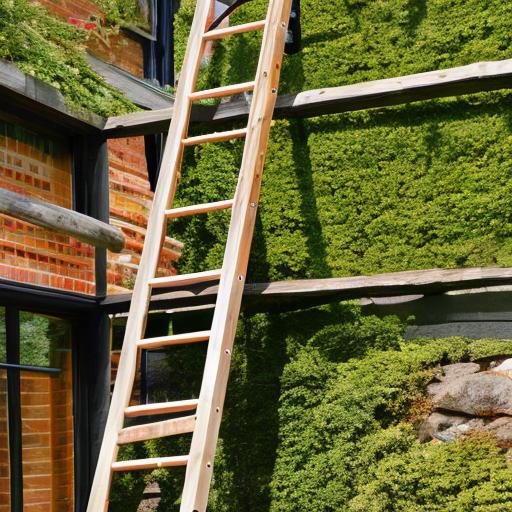}
    \end{minipage}}
    
    \caption{\textbf{Additional qualitative results of color-aligned latent diffusion on manual drawing conditions}. Each input (first row) includes a manual drawing as color condition. Targeting a text prompt, each column presents results of experimented methods.} 
    \label{fig:supp_qualitative_baseline}
\end{figure*}

In~\cref{sec:qualitative}, we present qualitative comparisons of our color-aligned diffusion method and existing baselines, highlighting the effectiveness of our method in color-conditioned image synthesis on in-the-wild color conditions. In this section, we provide additional comparison results on manual drawing conditions in~\cref{fig:supp_qualitative_baseline}, which also confirms the superior effectiveness of our method over existing baselines.

\begin{figure}[th]
  \begin{subfigure}{0.191\columnwidth}
    \includegraphics[width=\textwidth]{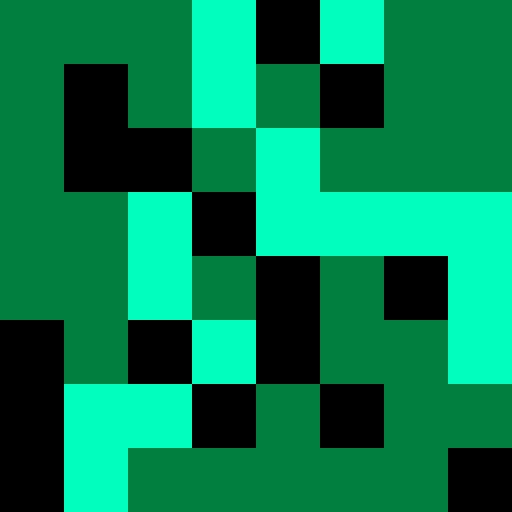}
  \end{subfigure}
  \hfill 
  \begin{subfigure}{0.191\columnwidth}
    \includegraphics[width=\textwidth]{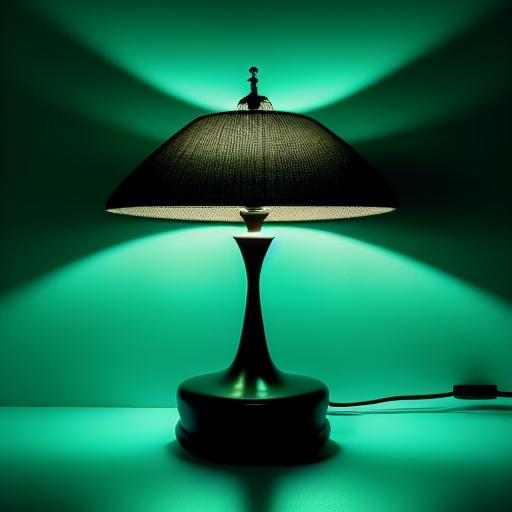}
  \end{subfigure}
  \hfill 
  \begin{subfigure}{0.191\columnwidth}
    \includegraphics[width=\textwidth]{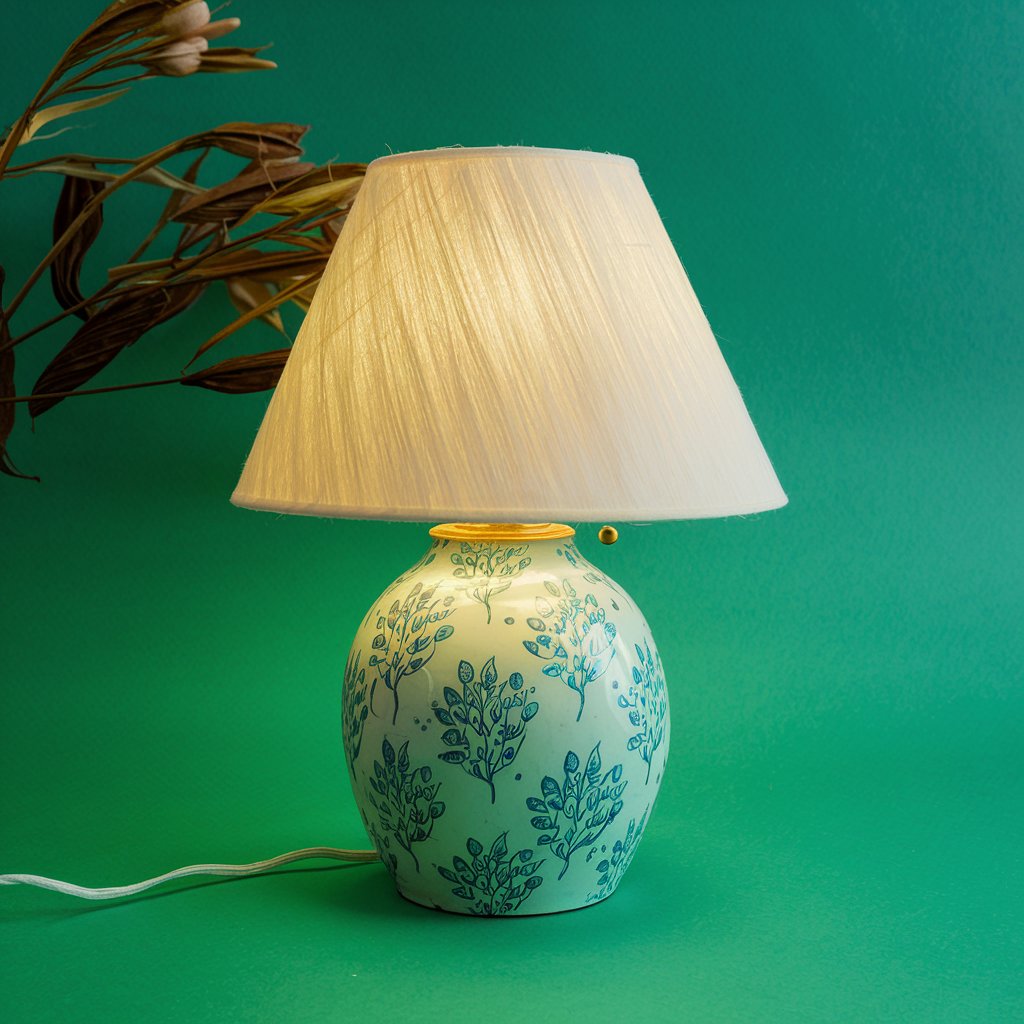}
  \end{subfigure}
  \hfill 
  \begin{subfigure}{0.191\columnwidth}
    \includegraphics[width=\textwidth]{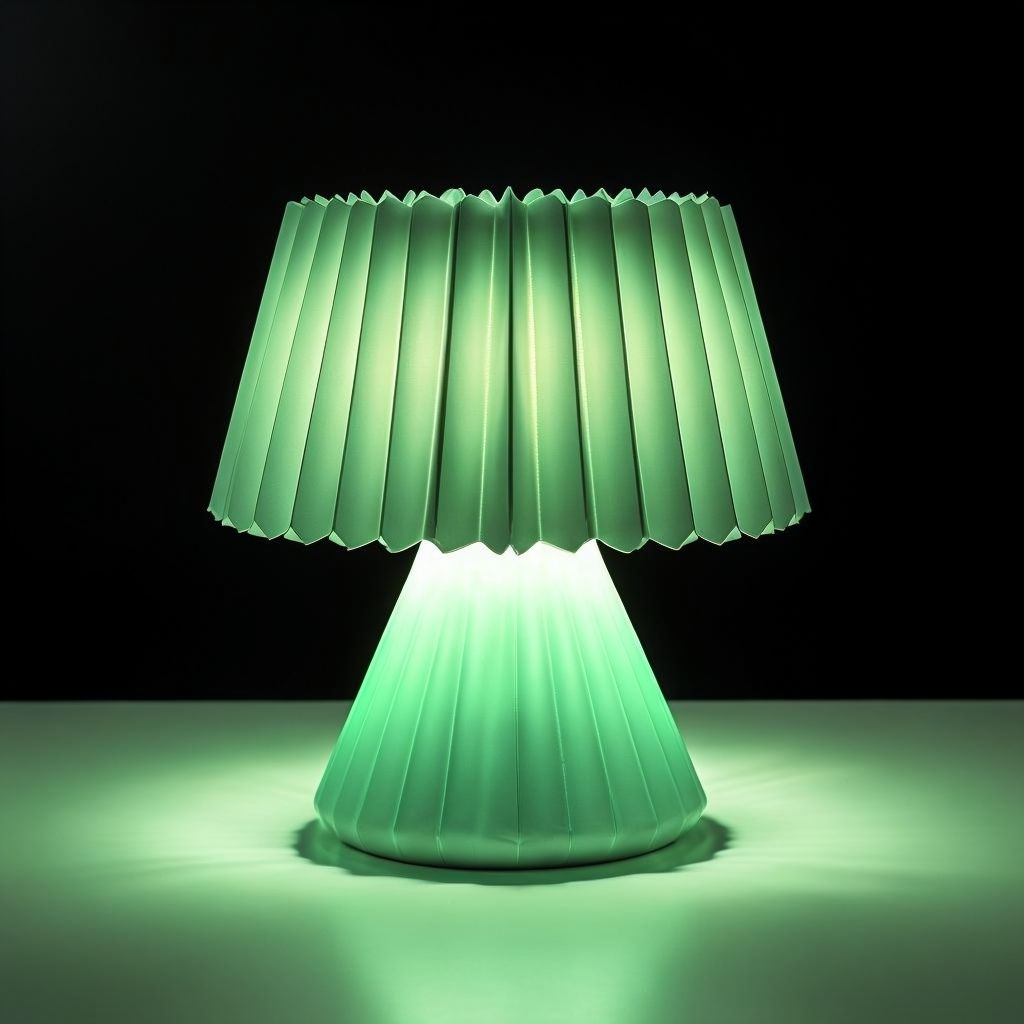}
  \end{subfigure}
  \hfill 
  \begin{subfigure}{0.191\columnwidth}
    \includegraphics[width=\textwidth]{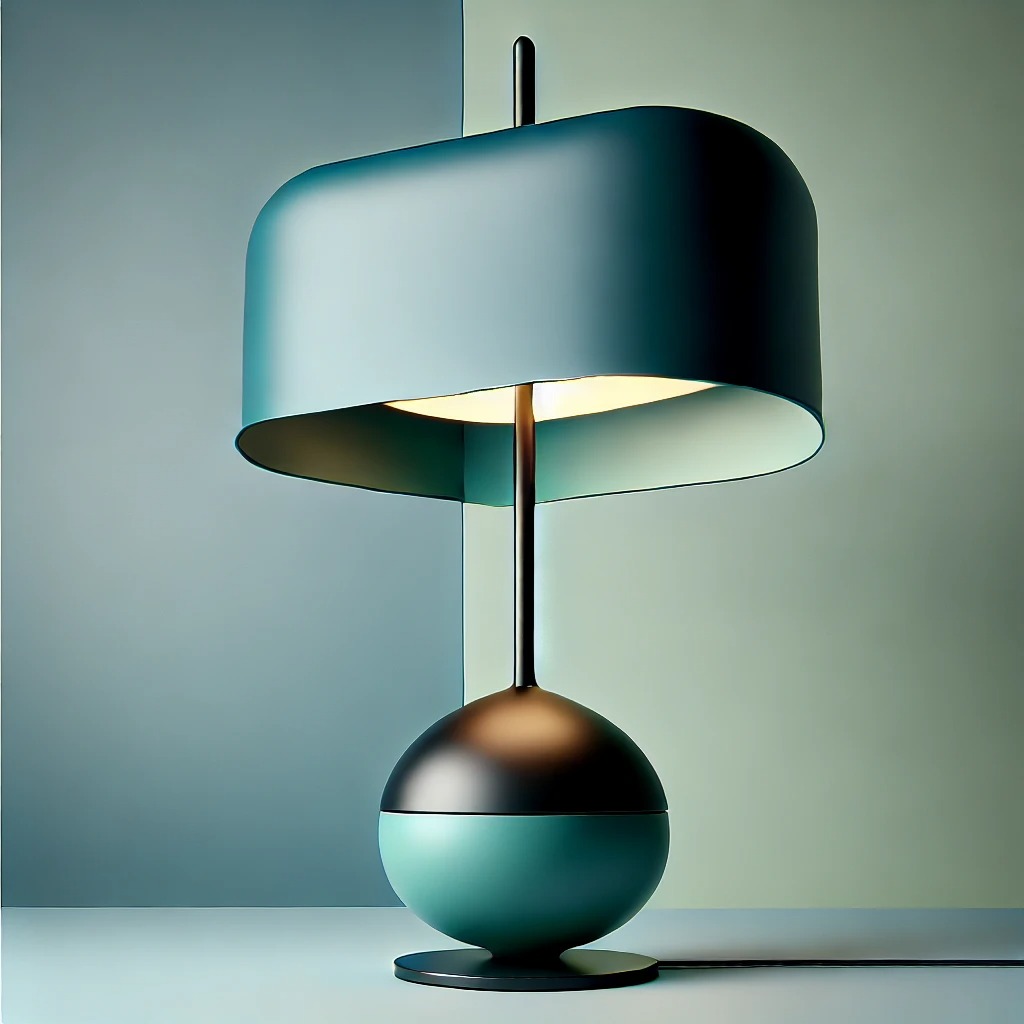}
  \end{subfigure}
  \\[0.6mm]
  \begin{subfigure}{0.191\columnwidth}
    \includegraphics[width=\textwidth]{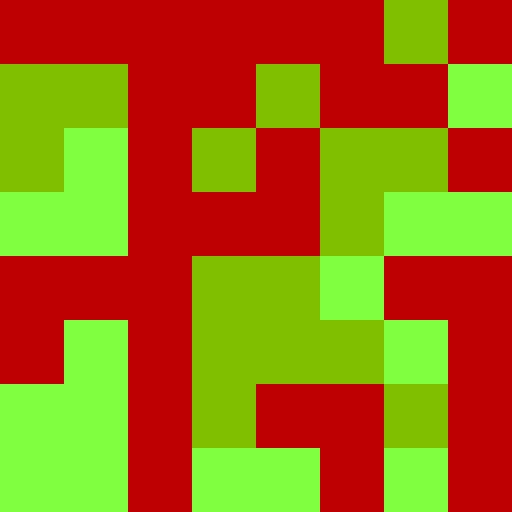}
  \end{subfigure}
  \hfill 
  \begin{subfigure}{0.191\columnwidth}
    \includegraphics[width=\textwidth]{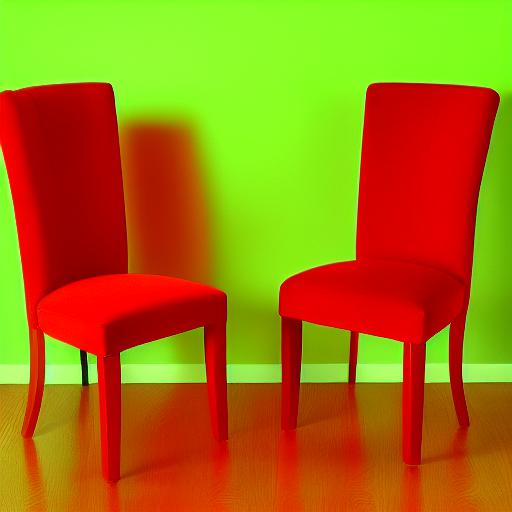}
  \end{subfigure}
  \hfill 
  \begin{subfigure}{0.191\columnwidth}
    \includegraphics[width=\textwidth]{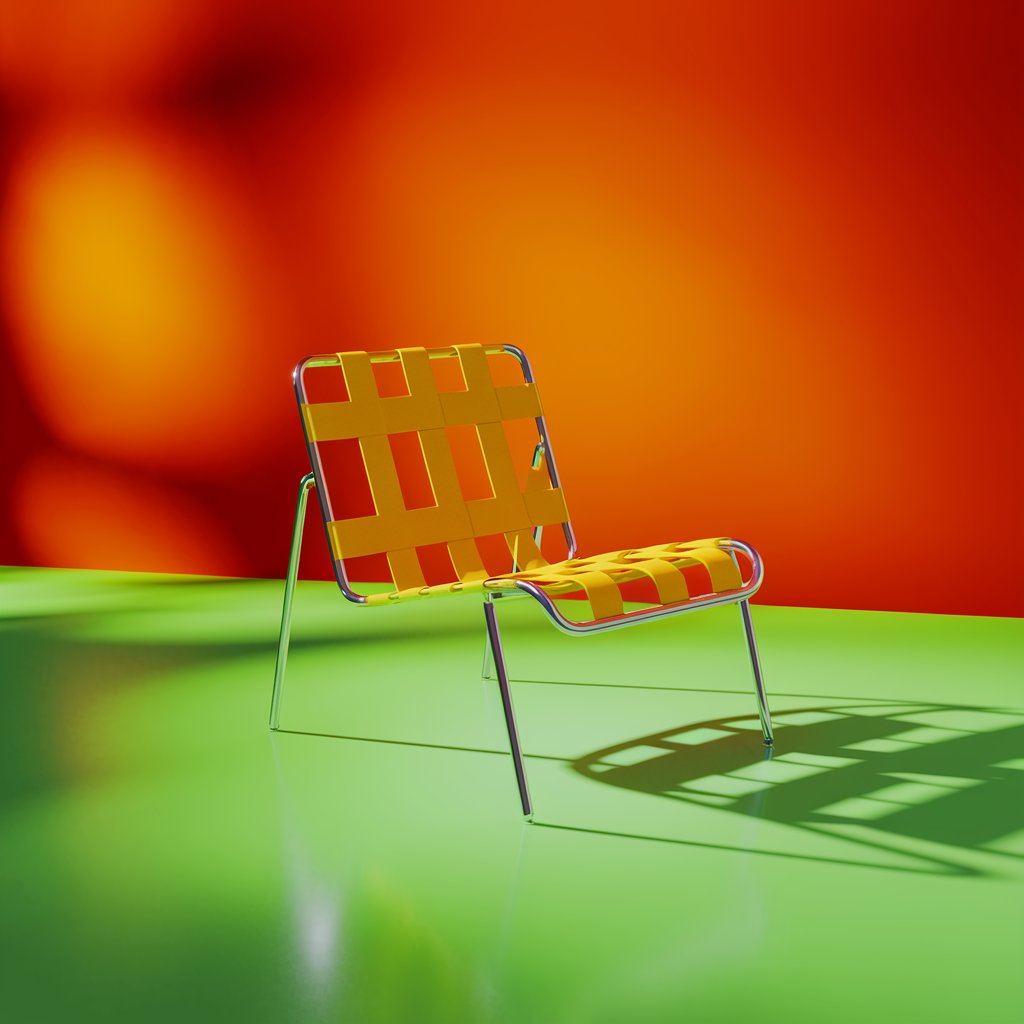}
  \end{subfigure}
  \hfill 
  \begin{subfigure}{0.191\columnwidth}
    \includegraphics[width=\textwidth]{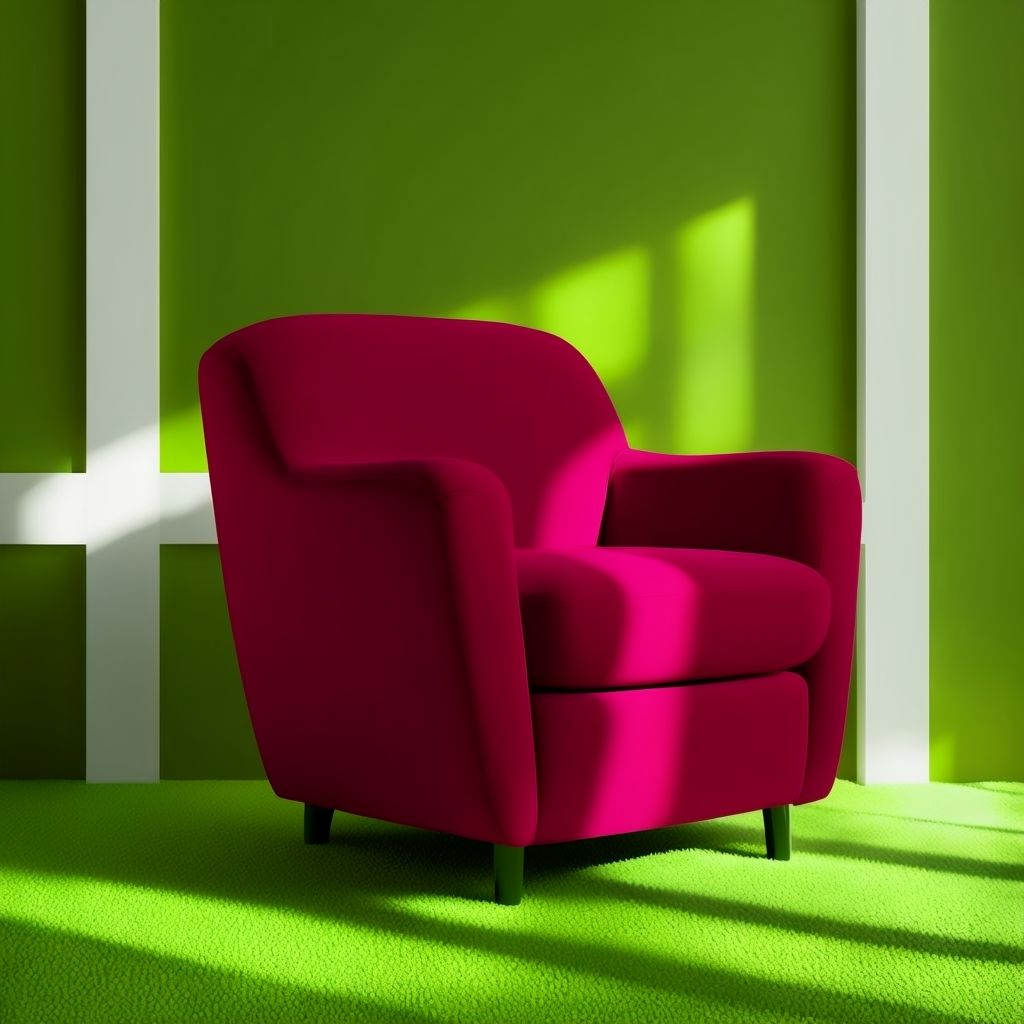}
  \end{subfigure}
  \hfill 
  \begin{subfigure}{0.191\columnwidth}
    \includegraphics[width=\textwidth]{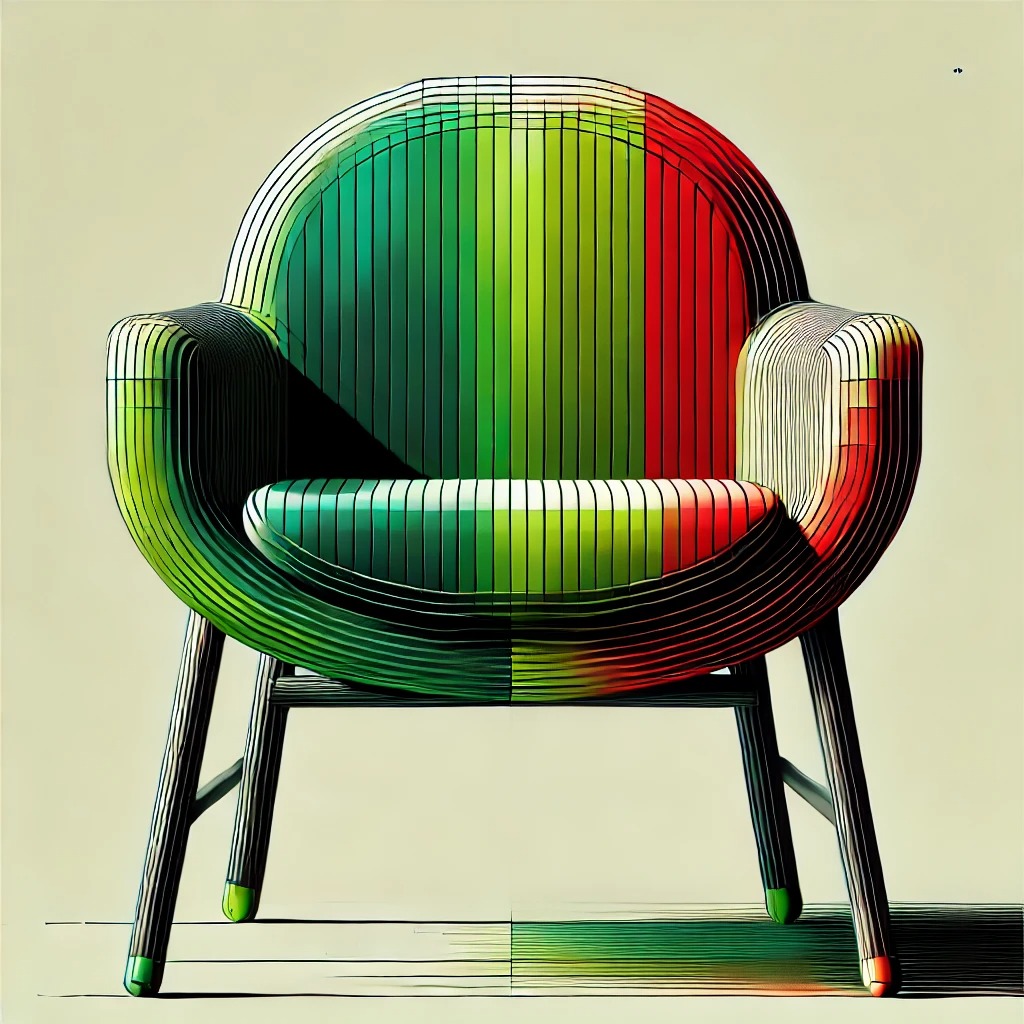}
  \end{subfigure}
  \\[0.6mm]
  \begin{subfigure}{0.191\columnwidth}
    \includegraphics[width=\textwidth]{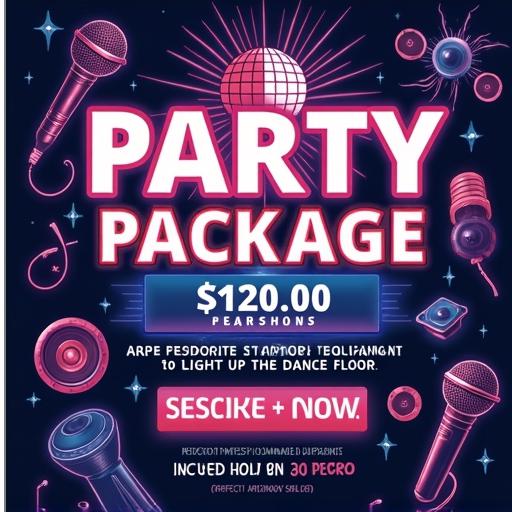}
  \end{subfigure}
  \hfill 
  \begin{subfigure}{0.191\columnwidth}
    \includegraphics[width=\textwidth]{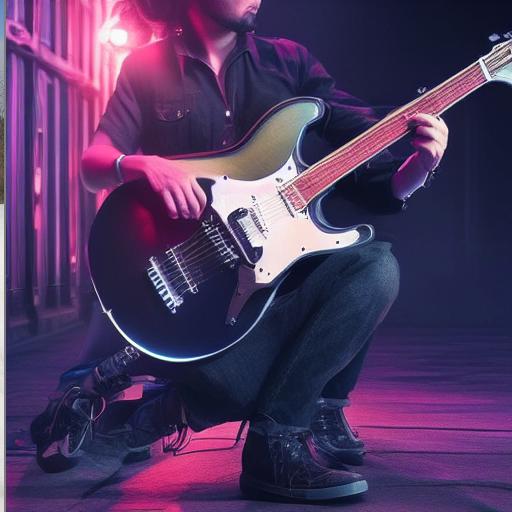}
  \end{subfigure}
  \hfill 
  \begin{subfigure}{0.191\columnwidth}
    \includegraphics[width=\textwidth]{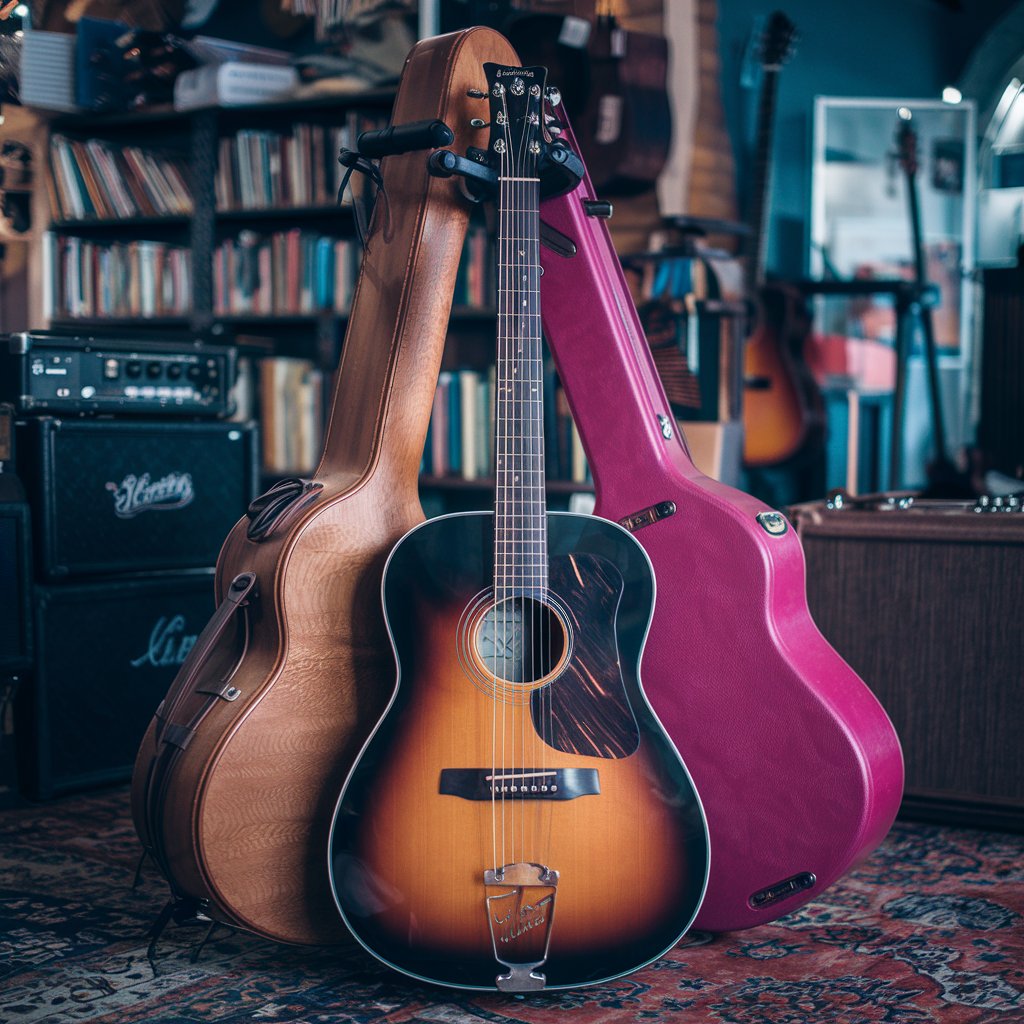}
  \end{subfigure}
  \hfill 
  \begin{subfigure}{0.191\columnwidth}
    \includegraphics[width=\textwidth]{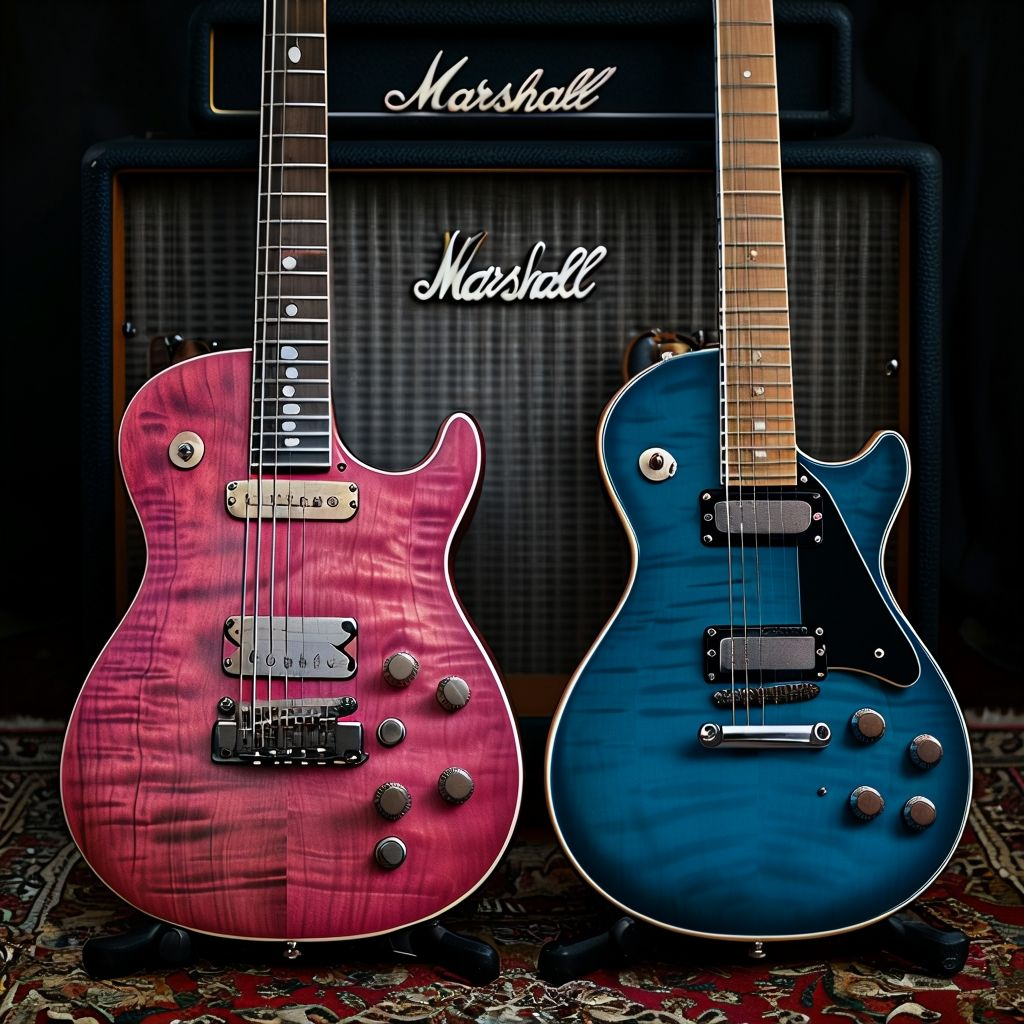}
  \end{subfigure}
  \hfill 
  \begin{subfigure}{0.191\columnwidth}
    \includegraphics[width=\textwidth]{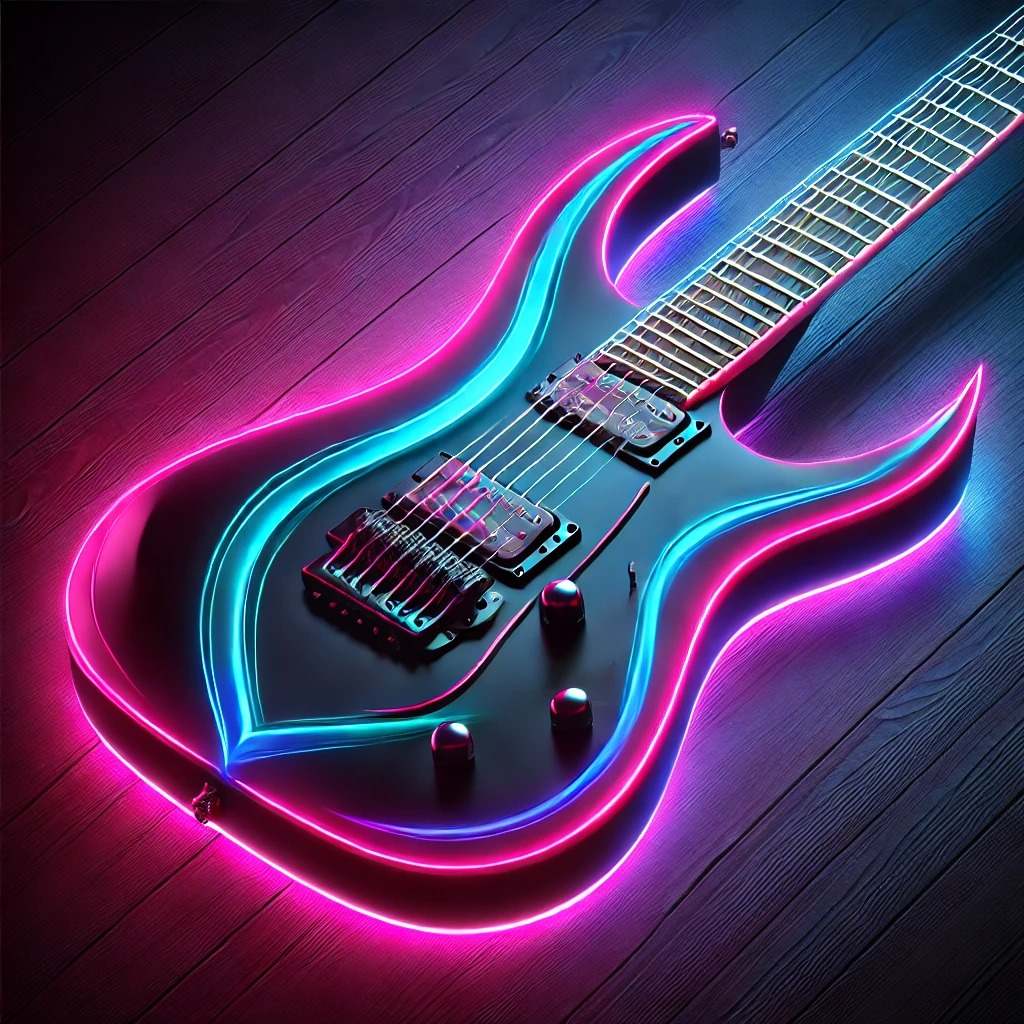}
  \end{subfigure}
  \\[0.6mm]
  \subfloat[Input Condition]{\includegraphics[width=0.191\columnwidth]{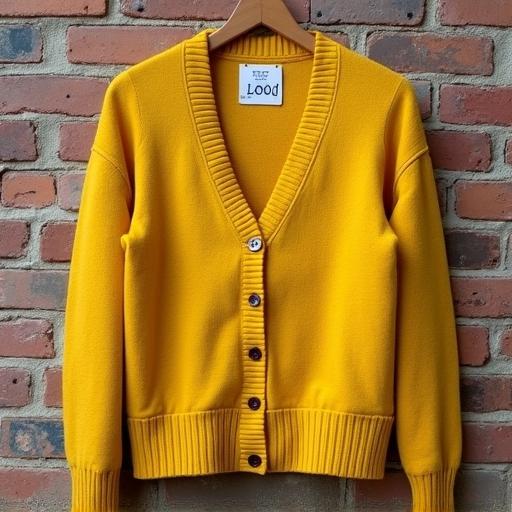}}
  \hfill 
  \subfloat[Ours (Fine\\-tune)]{\includegraphics[width=0.191\columnwidth]{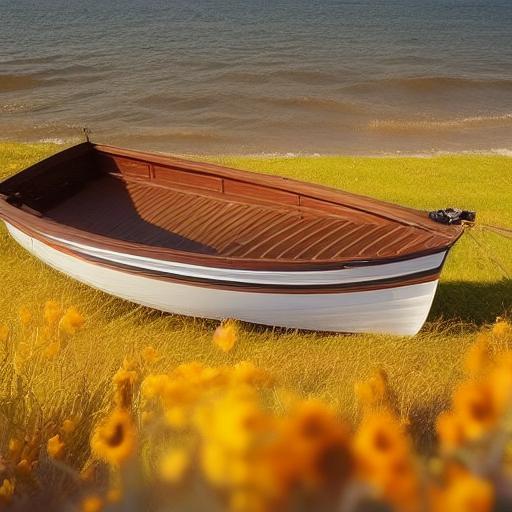}}
  \hfill 
  \subfloat[Ideogram-2.0~\cite{Ideogram20}]{\includegraphics[width=0.191\columnwidth]{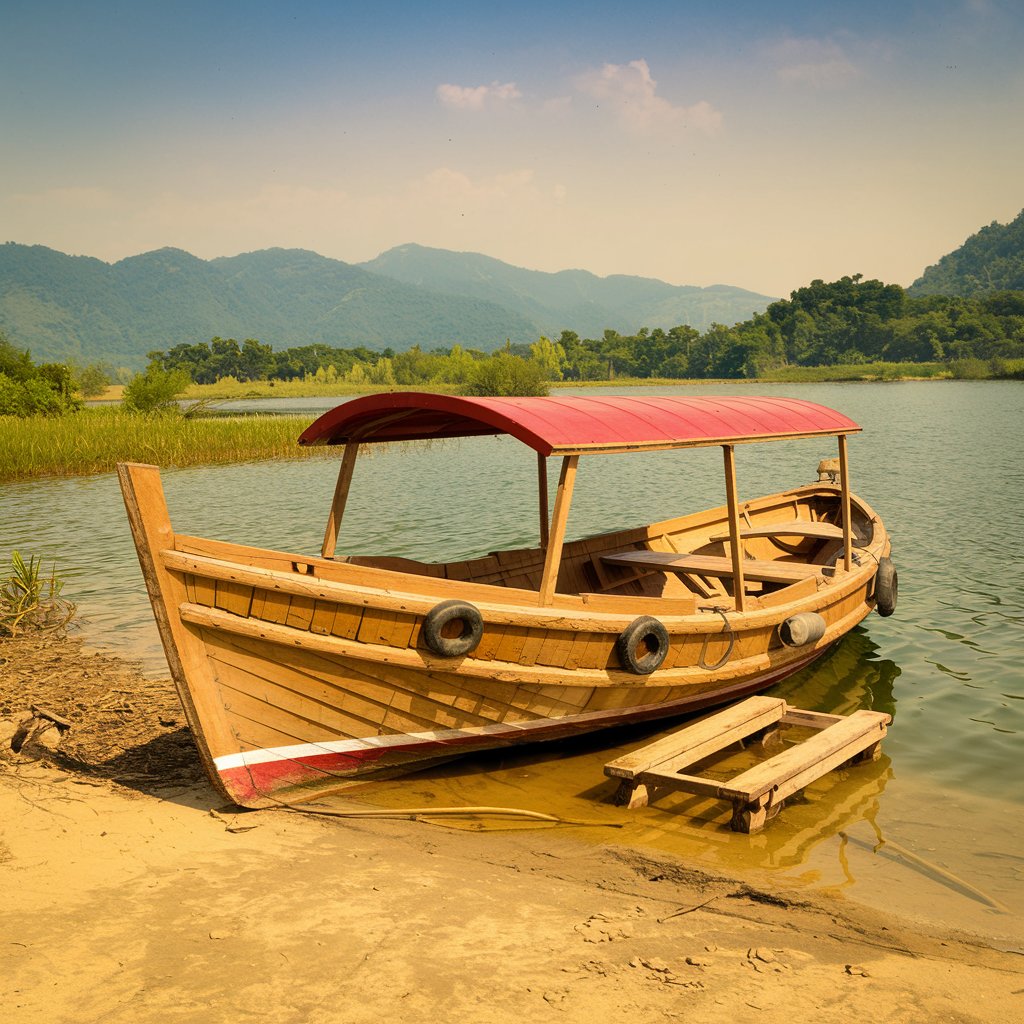}}
  \hfill 
  \subfloat[Playgrou-\\nd-v3~\cite{liu2024playground}]{\includegraphics[width=0.191\columnwidth]{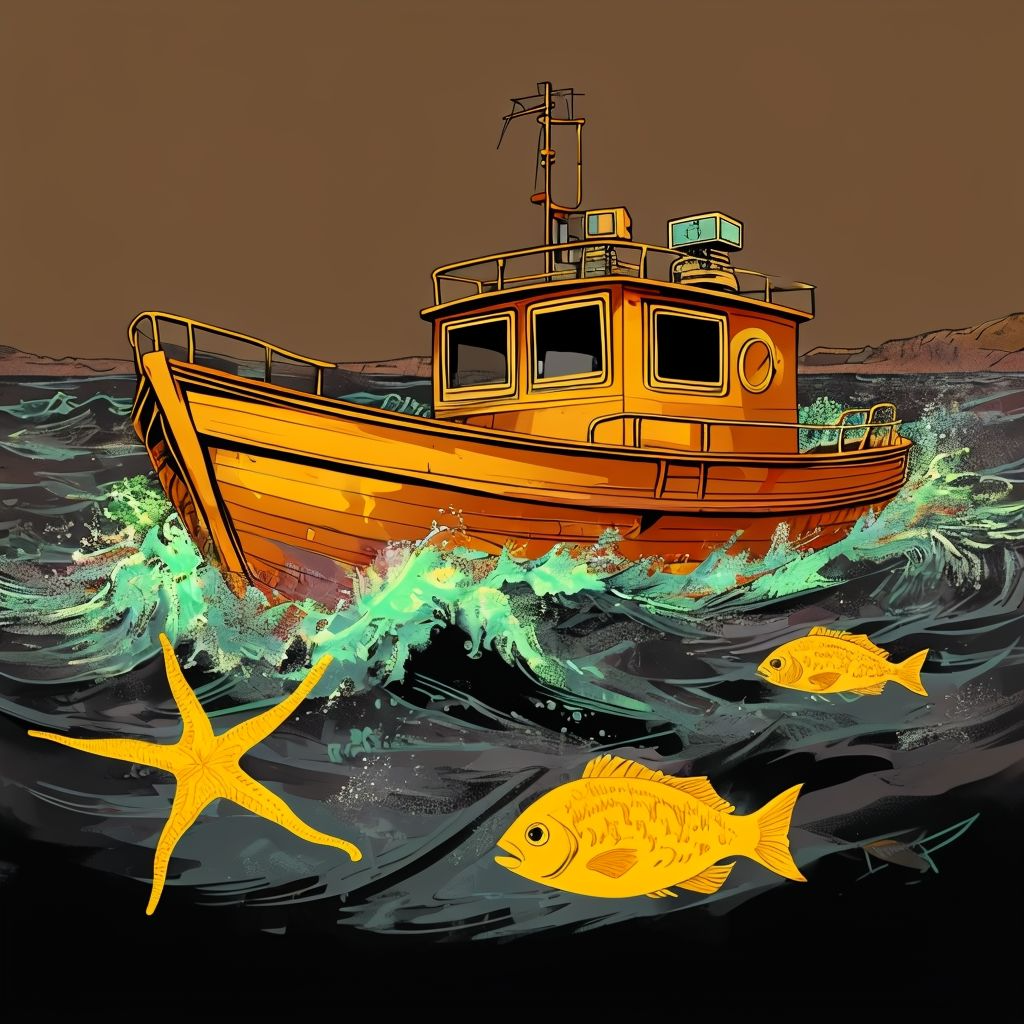}}
  \hfill 
  \subfloat[GPT-4o (VLM)~\cite{achiam2023gpt}]{\includegraphics[width=0.191\columnwidth]{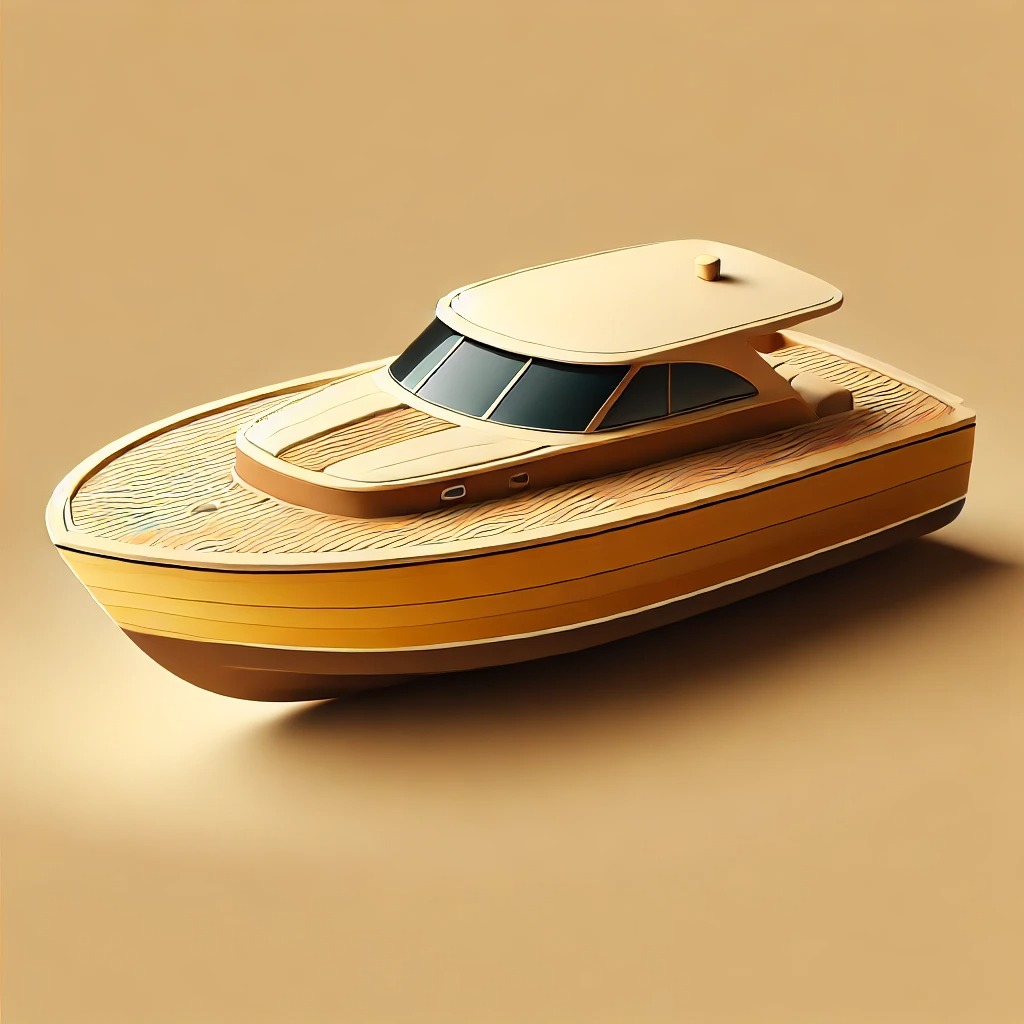}}

    \caption{\textbf{Additional qualitative comparisons with non-spatial palette conditioning baselines}. The first column shows the input color condition, and the remaining columns present the results of the experimented methods.} 
    \label{fig:supp_qualitative_nonspatial_baseline}
\end{figure}

We also conduct comparisons of our work with recent commercial products (Ideogram-2.0~\cite{Ideogram20} and Playground-v3~\cite{liu2024playground}) which offer non-spatial palette conditioning, and a recent VLM GPT-4o~\cite{achiam2023gpt}. We perform qualitative comparisons due to limited access to these methods. As shown in~\cref{fig:supp_qualitative_nonspatial_baseline}, these approaches condition the image synthesis roughly on input palettes, leading to less accurate results with unwanted or missing colors.

\section{Additional ablation studies}
\label{sec:additional_ablation_studies}
We conduct additional ablation studies on auxiliary technical components of our proposed color alignment method.

\subsection{Blurring before latent encoding}
Recall that for color-aligned latent diffusion, we blur the image color condition $\mathbf{x}_0$ before encoding it into the latent representation $\mathbf{z}_0$ for subsequent processes (\cref{sec:method:align_latent_space,sec:method:zeroshot_align}). We found it beneficial for reducing local high-frequency information in $\mathbf{x}_0$, leading to more accurate color encoding in $\mathbf{z}_0$.
Particularly, we implement the blurring operation as bilinear down-sampling and then bilinear up-sampling of $\mathbf{x}_0$, with the down- and up-sampling sizes defined as the strength of the blur (e.g., a strength of 3 means down-sampling to one-third of its size and then up-sampling back to its original size). 

\begin{figure*}[t]
  \includegraphics[width=\textwidth,trim={5mm 85mm 120mm 3mm},clip]{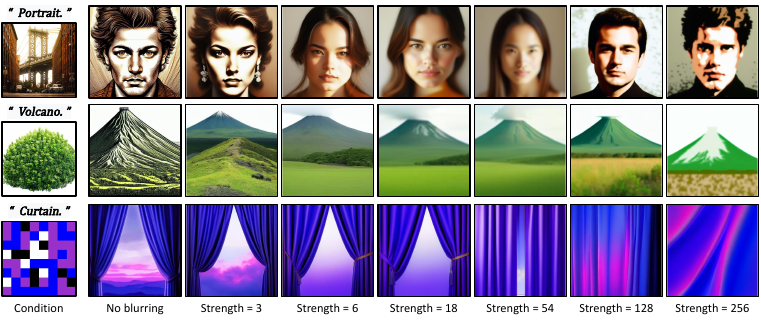}
  \\[-5.6mm]
    \caption{\textbf{Qualitative ablation study of blurring color condition before latent encoding}. The first column is the input condition. The rest columns, presenting from left to right, are the results from increasing blurring strength.}
\label{fig:ablation_blur}
\end{figure*}

We demonstrate the effect of the blurring operation in~\cref{fig:ablation_blur}. As shown, without blurring $\mathbf{x}_0$, the results tend to have dotted and fragmented texture (see the second column, better to be zoomed in). We found that a strength of 3 effectively balances texture smoothing and color conditioning, avoiding excessive strength that can make the output  overly smooth, blurry, and texture lacking. A quantitative evaluation of the blurring operation is presented in~\cref{tab:quantitative_ablation_blur}, which clearly shows that our current setting (i.e., strength=3) well balances all the color-conditioned image synthesis criteria evident by respective performance metrics.

\begin{table*}
    \centering
    \begin{tabular}{l|cc|cc}
        & FID $\downarrow$ & CLIPScore $\uparrow$ & CD-A $\downarrow$ & CD-C $\downarrow$ \\
        \midrule
        No blurring & 73.0 & 29.2 & 8.67 & 3.32\\
        \midrule
        Strength = 3 (Our current setting) & 70.5 & 29.4 & 4.63 & 5.12\\
        Strength = 6 & 75.6 & 29.5 & 3.60 & 6.41\\
        Strength = 18 & 78.9 & 29.3 & 3.36 & 8.67\\
        Strength = 54 & 82.2 & 28.9 & 3.15 & 14.2\\
        Strength = 128 & 79.9 & 29.0 & 3.88 & 13.6\\
        Strength = 256 & 86.5 & 28.2 & 4.70 & 33.9\\
        \bottomrule
    \end{tabular}
    \caption{\textbf{Quantitative ablation study of blurring color condition before latent encoding}. Note that all runs start with the same random seed for fair comparisons.}
    \label{tab:quantitative_ablation_blur}
\end{table*}

\subsection{Late-time stopping in latent color alignment}
\begin{figure*}[t]
  \includegraphics[width=\textwidth,trim={5mm 85mm 120mm 3mm},clip]{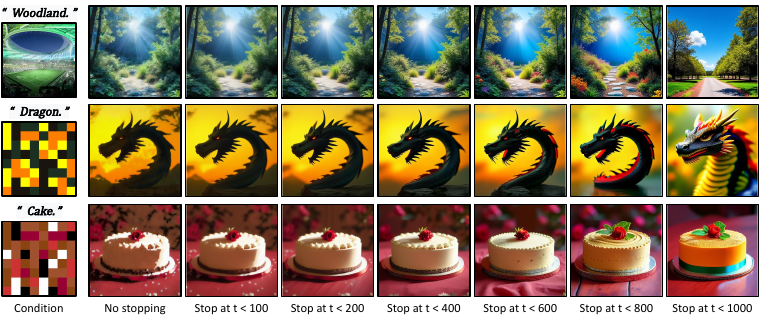}
  \\[-5.6mm]
    \caption{\textbf{Qualitative ablation study of late-time stopping of our latent color alignment}. The first column is the input condition. The rest columns, presenting from left to right, are the results from earlier late-time stopping of the alignment.}
\label{fig:ablation_late}
\end{figure*}

In our latent diffusion process, we propose to suspend the color alignment (\cref{eq:color_alignment,eq:color_alignment_one_shot}) at late time steps to allow for refinement of the final generated latent. This aligns with the goal of late time steps in the original diffusion, which focuses on refining local details of the final output. As shown in~\cref{fig:ablation_late}, without late-time stopping of the alignment, generated colors lack photo-realistic attributes such as natural lighting, vivid shadow, and clear semantics. We found stopping at time steps $t < 200$ enables the generation of these attributes. A too early stopping results in violations of the input color conditions, such as unintended generation of never-seen colors. As further validated quantitatively in~\cref{tab:quantitative_ablation_late}, stopping at $t < 200$ (or optionally $t < 400$) balances all the color-conditioned image synthesis criteria as indicated by performance metrics.

\begin{table*}
    \centering
    \begin{tabular}{l|cc|cc}
        & FID $\downarrow$ & CLIPScore $\uparrow$ & CD-A $\downarrow$ & CD-C $\downarrow$ \\
        \midrule
        No stopping & 76.8 & 29.3 & 3.32 & 4.58\\
        \midrule
        Stop at $t < 100$ & 70.6 & 29.3 & 3.77 & 5.78\\
        Stop at $t < 200$ (Our current setting) & 70.5 & 29.4 & 4.63 & 5.12\\
        Stop at $t < 400$ & 70.2 & 29.8 & 6.99 & 4.46\\
        Stop at $t < 600$ & 71.4 & 30.4 & 10.8 & 4.07\\
        Stop at $t < 800$ & 73.7 & 30.7 & 17.0 & 4.02\\
        Stop at $t < 1000$ & 78.4 & 30.6 & 26.8 & 4.42\\
        \bottomrule
    \end{tabular}
    \caption{\textbf{Quantitative ablation study of late-time stopping of our latent color alignment}. Note that all runs start with the same random seed (also used in~\cref{tab:quantitative_ablation_blur}) for fair comparisons.}
    \label{tab:quantitative_ablation_late}
\end{table*}

\section{More technical visualizations}
\label{sec:more_technical_details}
We provide additional visualizations to illustrate the differences between our color-aligned diffusion method and the regular diffusion method.

\begin{figure}
    \centering
    \includegraphics[width=\columnwidth,trim={1mm 120mm 180mm 3mm},clip]{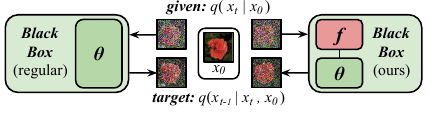}
    \caption{\textbf{Pipeline visualization} of our color-aligned diffusion compared to the regular pipeline.}
    \label{fig:pipeline_visualization}
\end{figure}

\begin{figure}[h]
    \centering
    \includegraphics[width=\columnwidth,trim={4mm 93mm 129mm 1mm},clip]{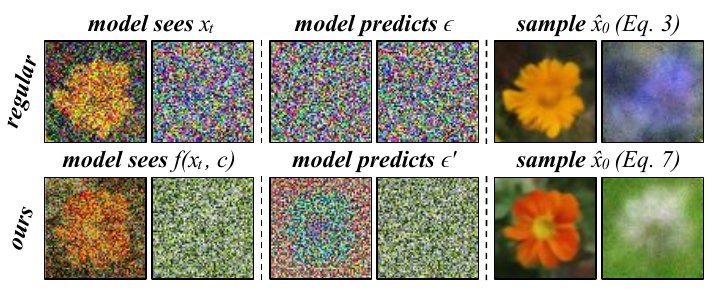}
    \caption{\textbf{Input and output visualization} of our color-aligned model compared to the regular model.}
    \label{fig:noise_visualization}
\end{figure}

Our method only modifies the intermediate pathway for reverse sampling, without affecting the overall objective image distribution across all diffusion steps.
Specifically, \cref{eq:color_alignment,eq:ours_train,eq:ours_pred} only influence the \textit{model query} (i.e., what the model $\theta$ sees and predicts) in the reverse sampling steps. 
The forward process (denoted by $q$) with no \textit{model query} involved, including $q(x_t|x_{t-1})$, $q(x_t|x_0)$, $q(x_{t-1}|x_t, x_0)$, and \cref{eq:regular_step}, remain identical to those in regular diffusion. As illustrated in \cref{fig:pipeline_visualization}, the original objective of the diffusion process (i.e., learning a mapping from Gaussian to image distribution) is preserved in our setting.
Additionally, in such a pathway, our model can adapt to inputs with non-Gaussian noise (distributed as in \cref{eq:ours_train}). We visualize this capability in \cref{fig:noise_visualization}.

\section{Data description details}
\label{sec:more_dataset_details}
We provide additional details about the data used in our experiments, supplementing the information in \cref{sec:datasets}.

Recall that to quantitatively assess the color \textit{disentanglement} in generated contents, we randomly generated 50 daily object prompts using ChatGPT~\cite{achiam2023gpt}. The text prompts are: 
``chair'', ``dog'', ``car'', ``book'', ``table'', ``house'', ``cat'',
``pen'', ``shirt'', ``bicycle'', ``shoe'', ``cup'', ``bed'', ``clock'', ``door'',
``flower'', ``fish'', ``camera'', ``blanket'', ``guitar'',
``bag'', ``bottle'', ``lamp'', ``desk'', ``towel'',
``suitcase'', ``basket'', ``helmet'', ``skateboard'',
``umbrella'',
``soap'', ``shampoo'', ``ladder'', ``painting'', ``brush'', ``glove'', ``hat'',
``belt'', ``wallet'', ``ring'', ``vase'', ``statue'', ``map'',
``ticket'', ``kite'',
``bus'', ``airplane'', ``rocket'', ``boat'', and ``crystal''. During evaluation, we randomly selected the prompts for generation.

To quantitatively evaluate our method under manual color conditions, we simulate user inputs with randomly selected color values and proportions. Each condition image includes 1 to 4 distinct random colors, with random color proportions of 25\%, 50\%, 75\%, or 100\%. If only one color is used, we replace 25\% of pixels of that color with pure black and white to simulate lighting and shadow colors.

\section{More implementation details}
\label{sec:more_implementation_details}

In this section, we provide additional implementation details of our method and other experimented baselines. This information supplements the descriptions in \cref{sec:baselines,sec:implementation_details}.

We followed the huggingface-diffusers~\cite{von_Platen_Diffusers_State-of-the-art_diffusion} implementation of DDPM~\cite{ho2020denoising} and Stable Diffusion~\cite{rombach2022high}. Specifically, we adopted the Stable Diffusion v1.5\footnote{https://huggingface.co/stable-diffusion-v1-5/stable-diffusion-v1-5} as the backbone for our image synthesis. This backbone was also used by all the compared baselines.
We used 1,000 time steps for training and 50 time steps for inference. We applied classifier-free guidance~\cite{ho2022classifier} to all methods with guidance scale 5, using the negative prompt ``Low quality, low resolution, blurry, ugly.''. We employed Adam Optimizer~\cite{kingma2014adam} with learning rate $1e-5$ and betas $(0.95, 0.999)$. For other hyperparameters, we followed the default settings in the Stable Diffusion v1.5.

To implement the color alignment in \cref{eq:color_alignment}, we updated $\mathbf{x}_t$ by searching for its most similar colors in $\mathbf{c}$ to achieve $f(\mathbf{x}_{t},\mathbf{c})$. Specifically, we applied the GPU-parallelizable PyTorch3D~\cite{ravi2020pytorch3d} Chamfer-loss\footnote{https://pytorch3d.readthedocs.io/en/latest/modules/loss.html} on every pixel color $\mathbf{x}_t[p]$ in $\mathbf{x}_t$ to find its most similar color $\mathbf{c}[q]$ in $\mathbf{c}$. This process results in a set of pixel pairs ($\mathbf{x}_t[p]$, $\mathbf{c}[q]$). Then, the color values of $\mathbf{c}[q]$ were assigned to the spatial locations of $\mathbf{x}_t[p]$ to form $f(\mathbf{x}_{t},\mathbf{c})$.

We applied the same idea to implement \cref{eq:color_alignment_one_shot}. Specifically, we constructed $g(\hat{\mathbf{x}}_0,\mathbf{c})$ by applying the Chamfer-loss updates multiple times on $\hat{\mathbf{x}}_0$ using $\mathbf{c}$. For each update, the pixels $\hat{\mathbf{x}}_0[p]$ were paired with their most proximate pixels $\mathbf{c}[q]$ (similar to the implementation of $f$ described earlier). However, for all pixel pairs ($\hat{\mathbf{x}}_0[p]$, $\mathbf{c}[q]$), we only selected those satisfied the one-to-one relation, that is, for the pairs where $\mathbf{c}[q]$ was repeatedly used, we only randomly selected one pair to form $g(\hat{\mathbf{x}}_0,\mathbf{c})$. The remaining unused pixels in $\hat{\mathbf{x}}_0$ and $\mathbf{c}$ were deferred to the next round of Chamfer-loss update, until all pixels were eventually paired to form a complete $g(\hat{\mathbf{x}}_0,\mathbf{c})$. This approach approximates the optimal one-to-one mapping at an acceptable cost.

\vspace{1cm}

\end{document}